\pdfoutput=1

\documentclass[10pt,twocolumn,letterpaper]{article}

\usepackage[pagenumbers]{cvpr} %

\usepackage{microtype}

\renewcommand{\paragraph}[1]{\par\vspace{.5em}\noindent\textbf{#1}\hspace{.6em}}

\usepackage{amsmath,amsfonts,bm}

\def\eqref#1{equation~\ref{#1}}

\def\1{\bm{1}}

\DeclareMathAlphabet{\mathsfit}{\encodingdefault}{\sfdefault}{m}{sl}
\SetMathAlphabet{\mathsfit}{bold}{\encodingdefault}{\sfdefault}{bx}{n}

\usepackage{multirow}
\usepackage{colortbl}
\usepackage{array}
\usepackage{algorithm}
\usepackage{algpseudocode}
\usepackage{fvextra}
\usepackage{xfp}
\usepackage{titletoc}
\usepackage{tikz}
\usetikzlibrary{arrows.meta}
\usepackage{pifont}%

\AddToHook{env/table/begin}{\setlength{\belowcaptionskip}{5pt}}
\AddToHook{env/table*/begin}{\setlength{\belowcaptionskip}{5pt}}

\AtEndPreamble{}

\newsavebox{\appTableBox}

\definecolor{cvprblue}{rgb}{0.21,0.49,0.74}
\usepackage[pagebackref,breaklinks,colorlinks,allcolors=cvprblue]{hyperref}

\def\confName{CVPR}
\def\confYear{2027}

\title{Mass Forcing: On-Policy Mass-Aware Distillation for Autoregressive Video Generation}

\author{
Yiyu Li\textsuperscript{1,2,\dag} \quad
Fei Wei\textsuperscript{2} \quad
Yong Wang\textsuperscript{2} \quad
Zaiquan Yang\textsuperscript{1}\\
Gerhard Petrus Hancke\textsuperscript{1} \quad
Xiangxiang Chu\textsuperscript{2} \quad
Rynson W.H. Lau\textsuperscript{3}\\[3pt]
\textsuperscript{1}City University of Hong Kong \quad
\textsuperscript{2}AMAP, Alibaba Group\\
\textsuperscript{3}City University of Hong Kong (Dongguan)
}

\makeatletter
\newcommand{\mfDeclareImage}[2]{\@namedef{mfimg@#1}{#2}}
\newcommand{\mfImage}[1]{\@ifundefined{mfimg@#1}{\errmessage{Undeclared image: #1}}{\@nameuse{mfimg@#1}}}
\makeatother
\mfDeclareImage{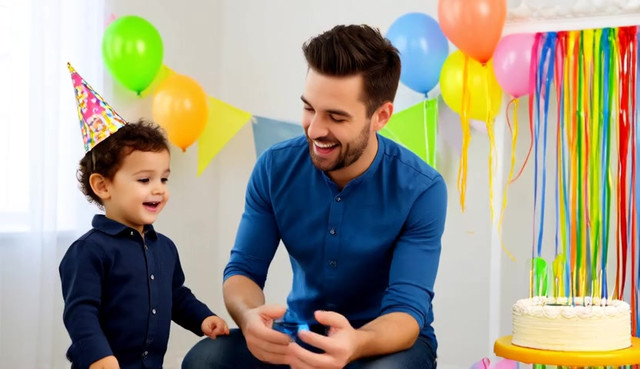}{\includegraphics[width=\teaserFrameW]{assets/teaser/causal-forcing-0.jpg}}
\mfDeclareImage{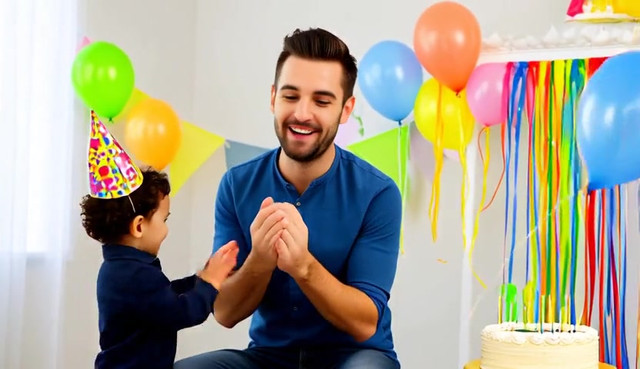}{\includegraphics[width=\teaserFrameW]{assets/teaser/causal-forcing-2.5.jpg}}
\mfDeclareImage{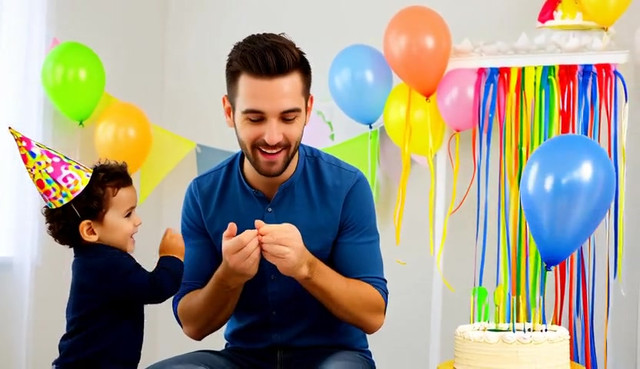}{\includegraphics[width=\teaserFrameW]{assets/teaser/causal-forcing-5.jpg}}
\mfDeclareImage{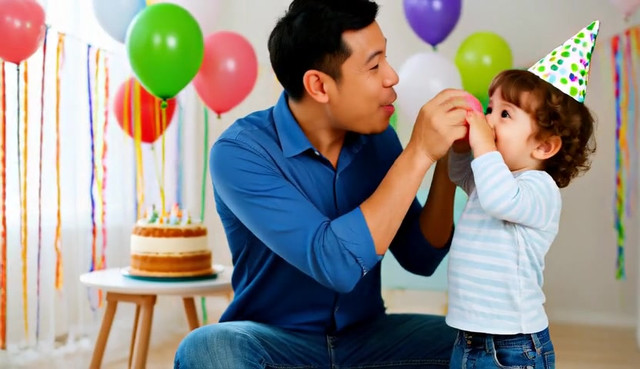}{\includegraphics[width=\teaserFrameW]{assets/teaser/ours-0.jpg}}
\mfDeclareImage{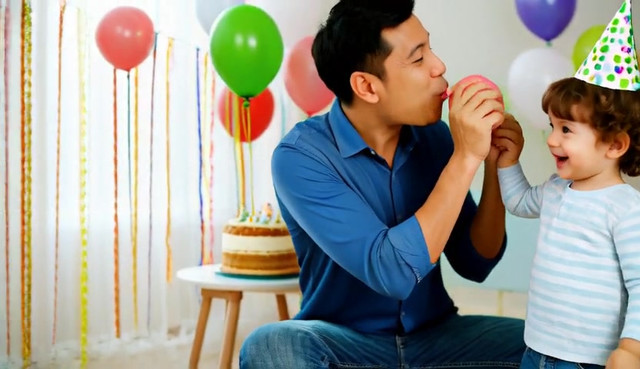}{\includegraphics[width=\teaserFrameW]{assets/teaser/ours-2.5.jpg}}
\mfDeclareImage{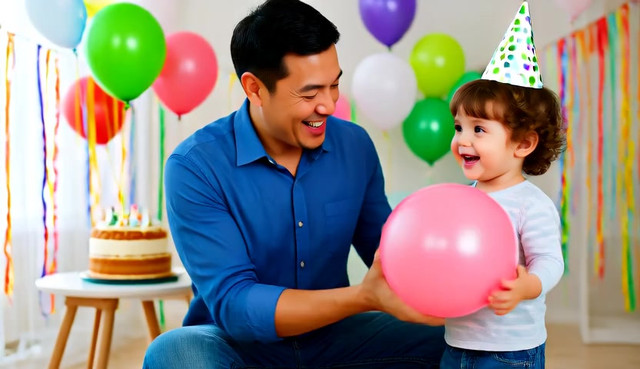}{\includegraphics[width=\teaserFrameW]{assets/teaser/ours-5.jpg}}
\mfDeclareImage{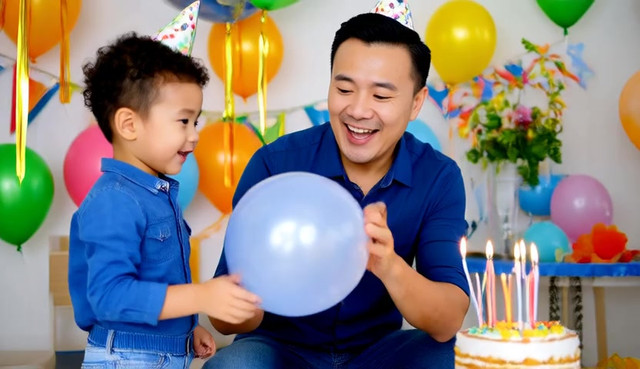}{\includegraphics[width=\teaserFrameW]{assets/teaser/self-forcing-0.jpg}}
\mfDeclareImage{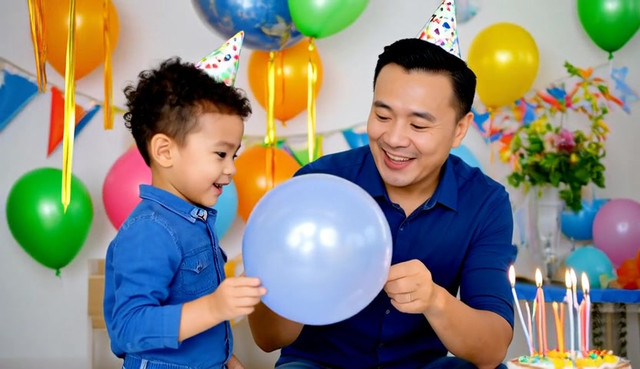}{\includegraphics[width=\teaserFrameW]{assets/teaser/self-forcing-2.5.jpg}}
\mfDeclareImage{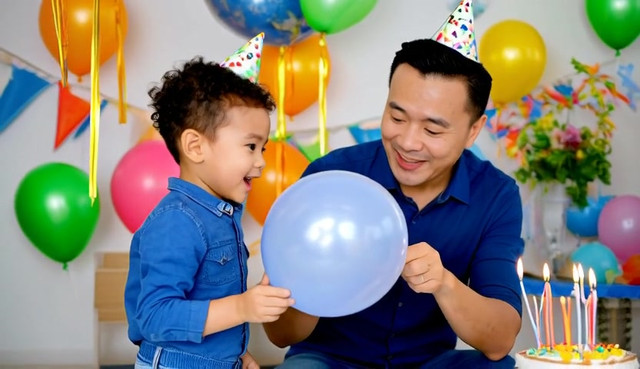}{\includegraphics[width=\teaserFrameW]{assets/teaser/self-forcing-5.jpg}}
\mfDeclareImage{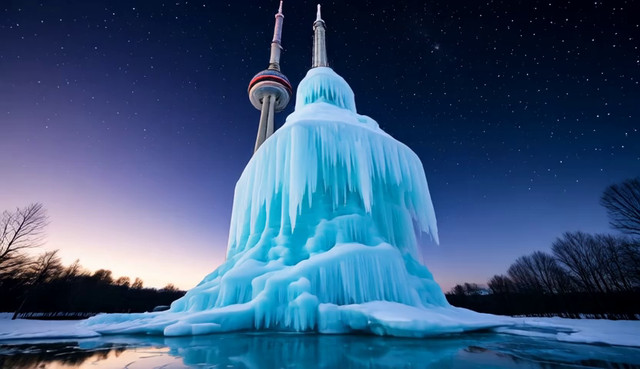}{\includegraphics[width=\qualFrameWidth]{assets/qualitative/prompt_000268_individual/causal_forcing/frame_000_0s.jpg}}
\mfDeclareImage{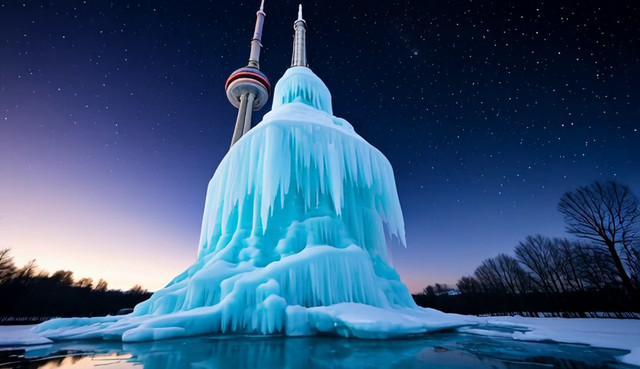}{\includegraphics[width=\qualFrameWidth]{assets/qualitative/prompt_000268_individual/causal_forcing/frame_040_2p5s.jpg}}
\mfDeclareImage{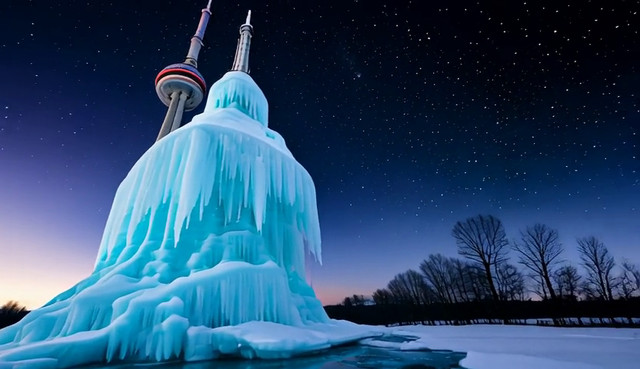}{\includegraphics[width=\qualFrameWidth]{assets/qualitative/prompt_000268_individual/causal_forcing/frame_080_5s.jpg}}
\mfDeclareImage{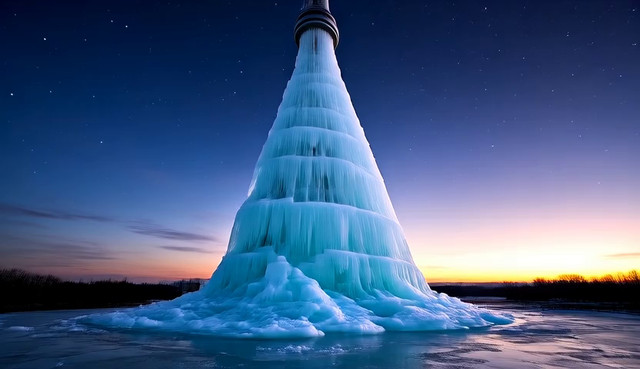}{\includegraphics[width=\qualFrameWidth]{assets/qualitative/prompt_000268_individual/ours/frame_000_0s.jpg}}
\mfDeclareImage{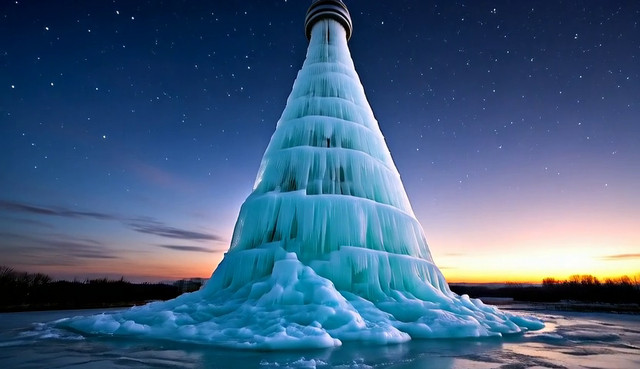}{\includegraphics[width=\qualFrameWidth]{assets/qualitative/prompt_000268_individual/ours/frame_040_2p5s.jpg}}
\mfDeclareImage{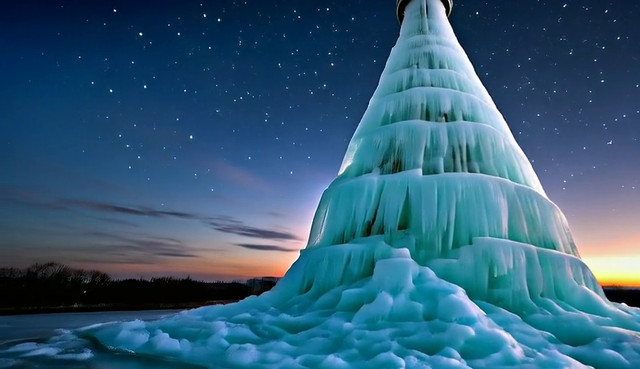}{\includegraphics[width=\qualFrameWidth]{assets/qualitative/prompt_000268_individual/ours/frame_080_5s.jpg}}
\mfDeclareImage{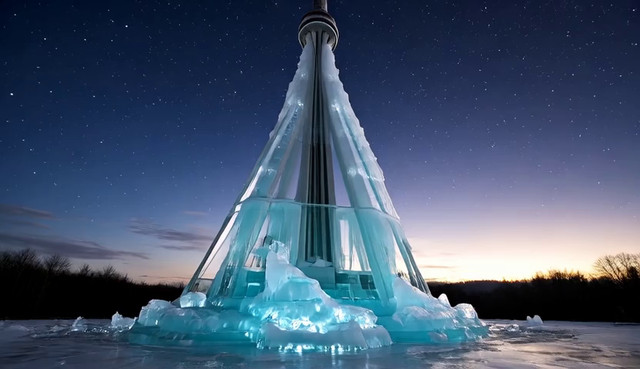}{\includegraphics[width=\qualFrameWidth]{assets/qualitative/prompt_000268_individual/self_forcing/frame_000_0s.jpg}}
\mfDeclareImage{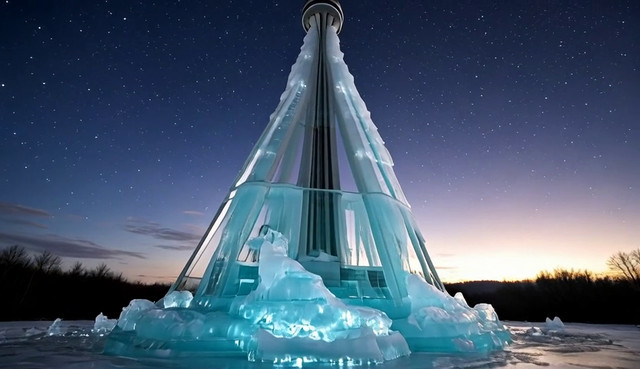}{\includegraphics[width=\qualFrameWidth]{assets/qualitative/prompt_000268_individual/self_forcing/frame_040_2p5s.jpg}}
\mfDeclareImage{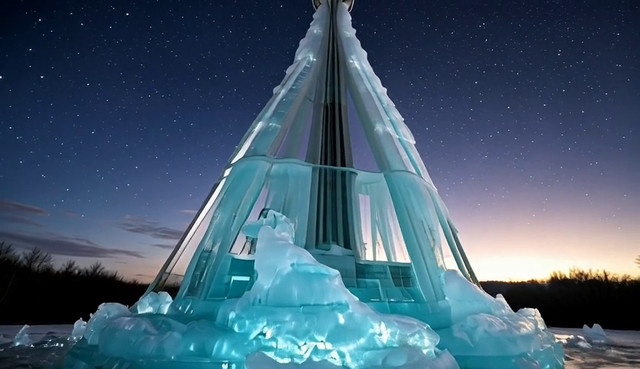}{\includegraphics[width=\qualFrameWidth]{assets/qualitative/prompt_000268_individual/self_forcing/frame_080_5s.jpg}}
\mfDeclareImage{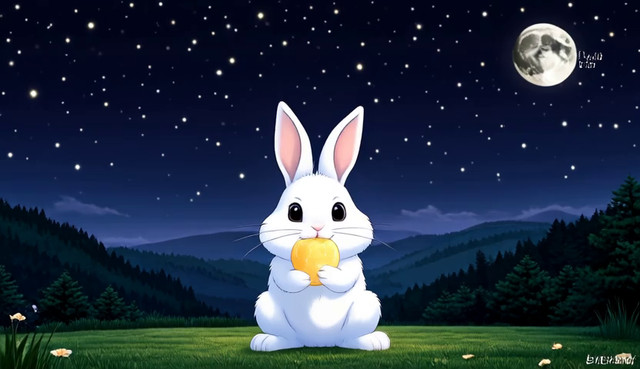}{\includegraphics[width=\qualFrameWidth]{assets/qualitative/seed_324_individual/causal_forcing/frame_000_0s.jpg}}
\mfDeclareImage{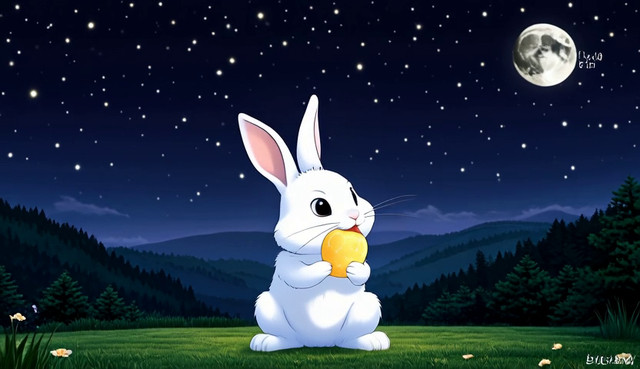}{\includegraphics[width=\qualFrameWidth]{assets/qualitative/seed_324_individual/causal_forcing/frame_040_2p5s.jpg}}
\mfDeclareImage{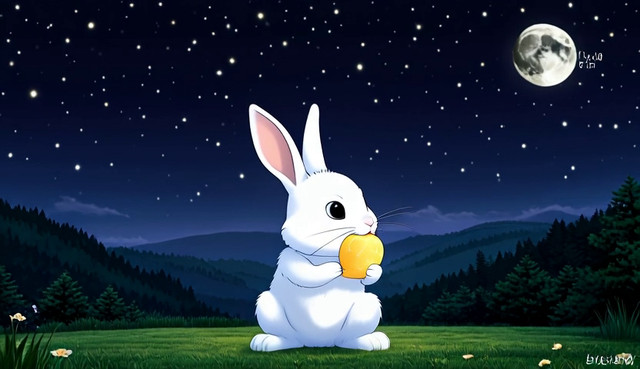}{\includegraphics[width=\qualFrameWidth]{assets/qualitative/seed_324_individual/causal_forcing/frame_080_5s.jpg}}
\mfDeclareImage{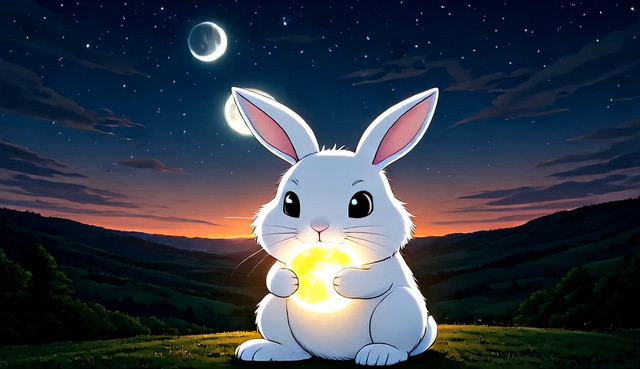}{\includegraphics[width=\qualFrameWidth]{assets/qualitative/seed_324_individual/ours/frame_000_0s.jpg}}
\mfDeclareImage{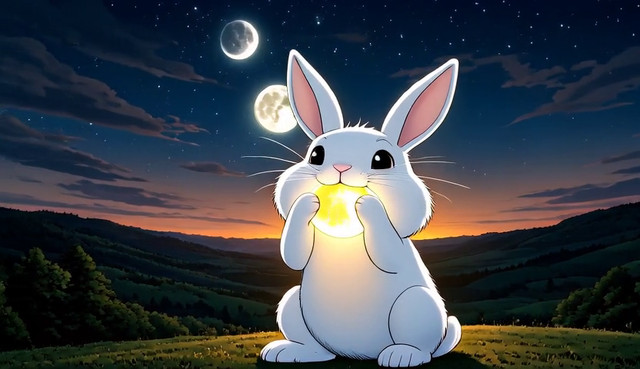}{\includegraphics[width=\qualFrameWidth]{assets/qualitative/seed_324_individual/ours/frame_040_2p5s.jpg}}
\mfDeclareImage{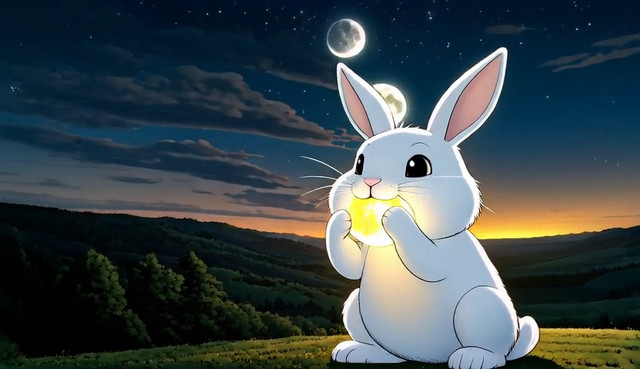}{\includegraphics[width=\qualFrameWidth]{assets/qualitative/seed_324_individual/ours/frame_080_5s.jpg}}
\mfDeclareImage{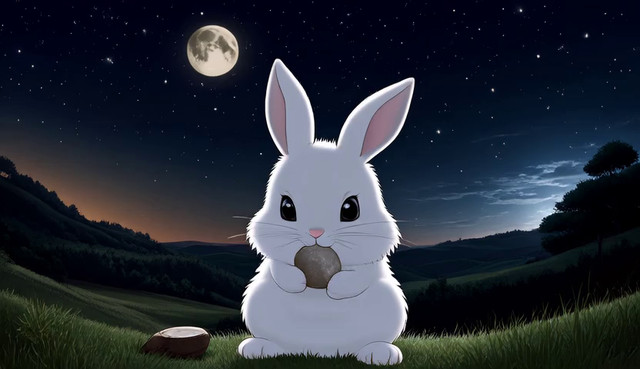}{\includegraphics[width=\qualFrameWidth]{assets/qualitative/seed_324_individual/self_forcing/frame_000_0s.jpg}}
\mfDeclareImage{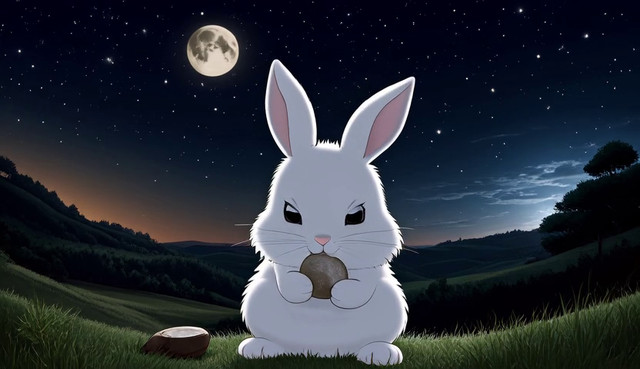}{\includegraphics[width=\qualFrameWidth]{assets/qualitative/seed_324_individual/self_forcing/frame_040_2p5s.jpg}}
\mfDeclareImage{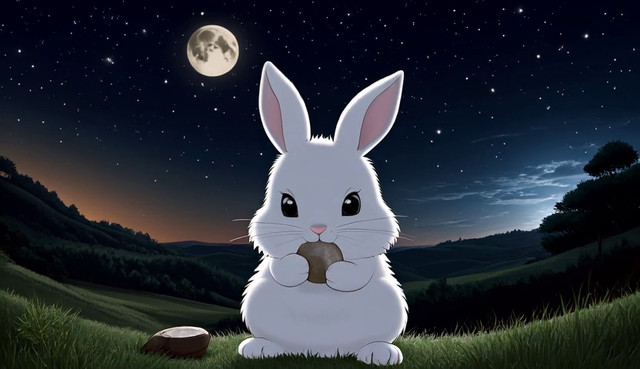}{\includegraphics[width=\qualFrameWidth]{assets/qualitative/seed_324_individual/self_forcing/frame_080_5s.jpg}}
\mfDeclareImage{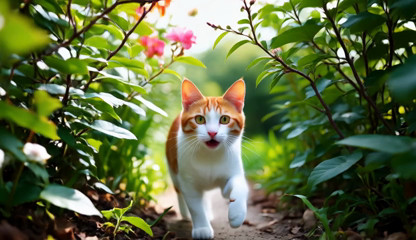}{\includegraphics[width=\longVidFrameWidth]{assets/long-video/prompt_001_seed0/causal_forcing/f1.jpg}}
\mfDeclareImage{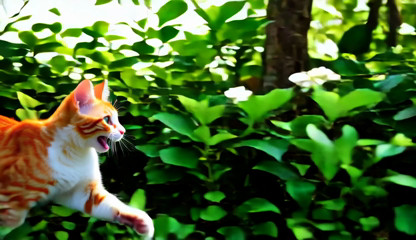}{\includegraphics[width=\longVidFrameWidth]{assets/long-video/prompt_001_seed0/causal_forcing/f2.jpg}}
\mfDeclareImage{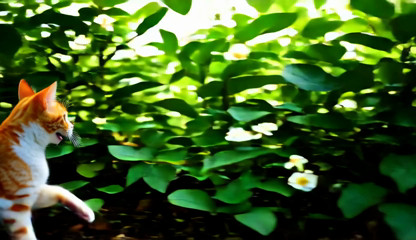}{\includegraphics[width=\longVidFrameWidth]{assets/long-video/prompt_001_seed0/causal_forcing/f3.jpg}}
\mfDeclareImage{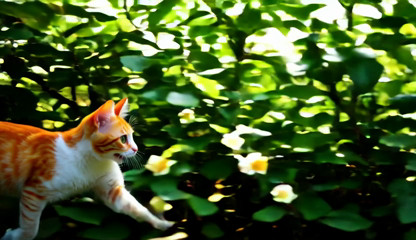}{\includegraphics[width=\longVidFrameWidth]{assets/long-video/prompt_001_seed0/causal_forcing/f4.jpg}}
\mfDeclareImage{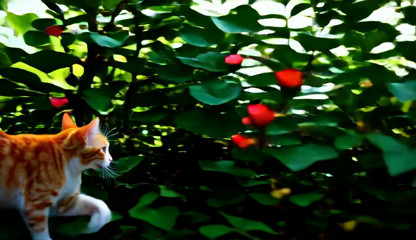}{\includegraphics[width=\longVidFrameWidth]{assets/long-video/prompt_001_seed0/causal_forcing/f5.jpg}}
\mfDeclareImage{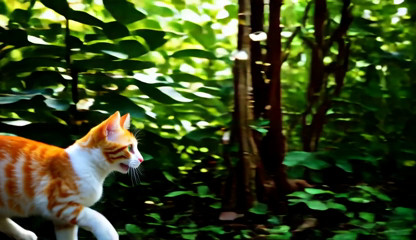}{\includegraphics[width=\longVidFrameWidth]{assets/long-video/prompt_001_seed0/causal_forcing/f6.jpg}}
\mfDeclareImage{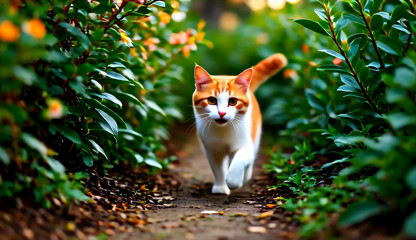}{\includegraphics[width=\longVidFrameWidth]{assets/long-video/prompt_001_seed0/ours/f1.jpg}}
\mfDeclareImage{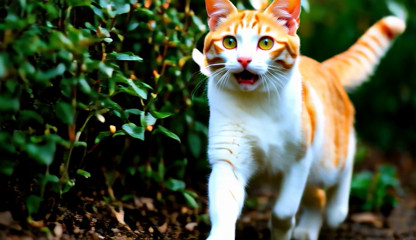}{\includegraphics[width=\longVidFrameWidth]{assets/long-video/prompt_001_seed0/ours/f2.jpg}}
\mfDeclareImage{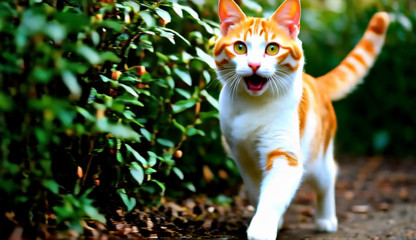}{\includegraphics[width=\longVidFrameWidth]{assets/long-video/prompt_001_seed0/ours/f3.jpg}}
\mfDeclareImage{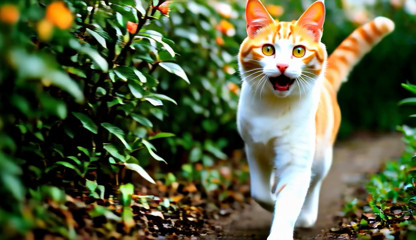}{\includegraphics[width=\longVidFrameWidth]{assets/long-video/prompt_001_seed0/ours/f4.jpg}}
\mfDeclareImage{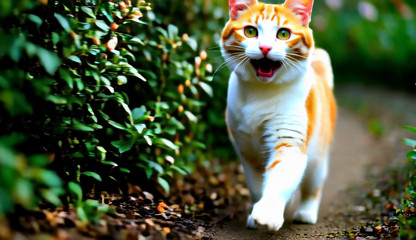}{\includegraphics[width=\longVidFrameWidth]{assets/long-video/prompt_001_seed0/ours/f5.jpg}}
\mfDeclareImage{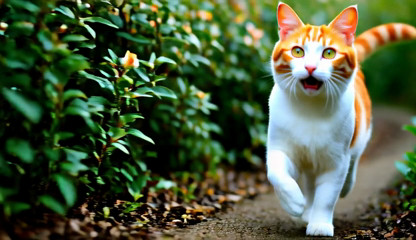}{\includegraphics[width=\longVidFrameWidth]{assets/long-video/prompt_001_seed0/ours/f6.jpg}}
\mfDeclareImage{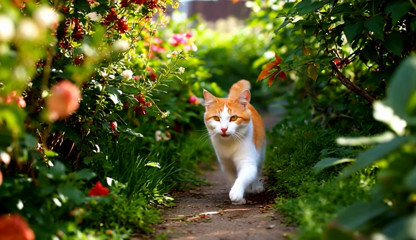}{\includegraphics[width=\longVidFrameWidth]{assets/long-video/prompt_001_seed0/self_forcing/f1.jpg}}
\mfDeclareImage{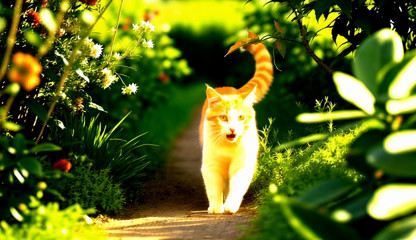}{\includegraphics[width=\longVidFrameWidth]{assets/long-video/prompt_001_seed0/self_forcing/f2.jpg}}
\mfDeclareImage{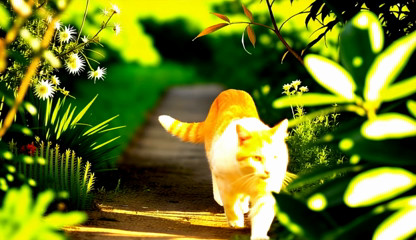}{\includegraphics[width=\longVidFrameWidth]{assets/long-video/prompt_001_seed0/self_forcing/f3.jpg}}
\mfDeclareImage{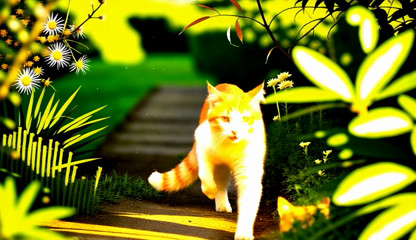}{\includegraphics[width=\longVidFrameWidth]{assets/long-video/prompt_001_seed0/self_forcing/f4.jpg}}
\mfDeclareImage{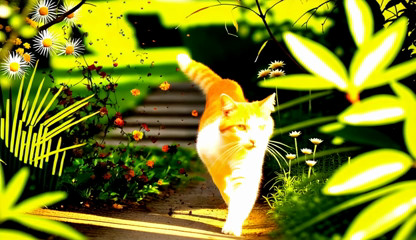}{\includegraphics[width=\longVidFrameWidth]{assets/long-video/prompt_001_seed0/self_forcing/f5.jpg}}
\mfDeclareImage{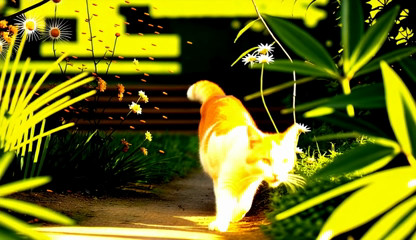}{\includegraphics[width=\longVidFrameWidth]{assets/long-video/prompt_001_seed0/self_forcing/f6.jpg}}
\mfDeclareImage{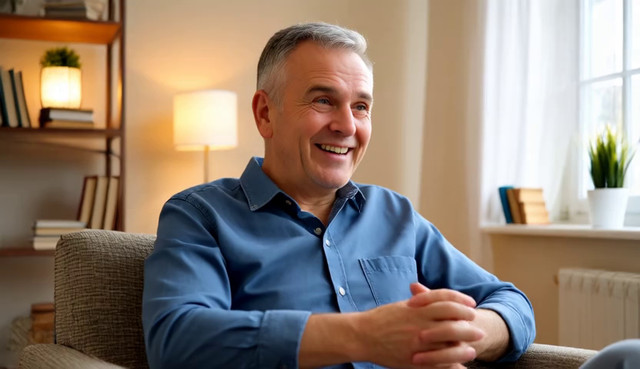}{\includegraphics[width=\extendedQualFrameWidth]{assets/extended-qualitative/prompt_000338/causal_forcing/frame_000_0s.jpg}}
\mfDeclareImage{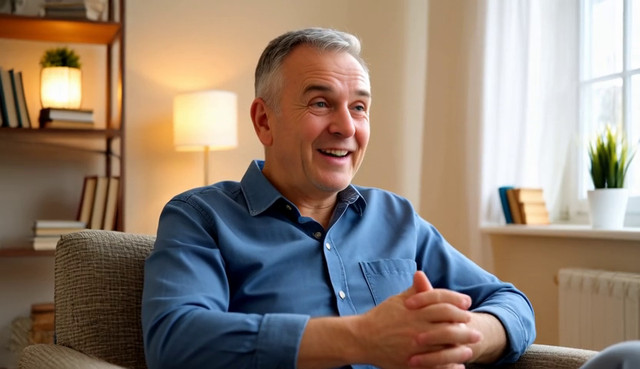}{\includegraphics[width=\extendedQualFrameWidth]{assets/extended-qualitative/prompt_000338/causal_forcing/frame_020_1p25s.jpg}}
\mfDeclareImage{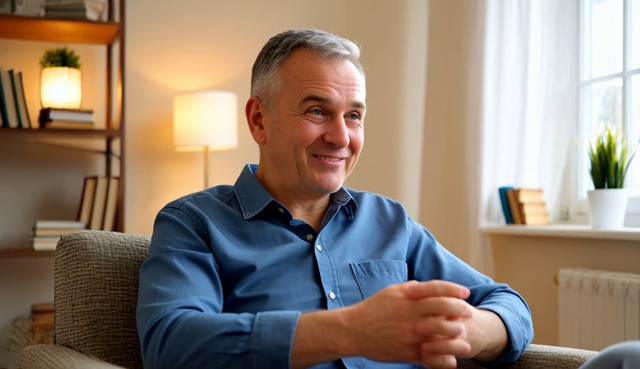}{\includegraphics[width=\extendedQualFrameWidth]{assets/extended-qualitative/prompt_000338/causal_forcing/frame_040_2p5s.jpg}}
\mfDeclareImage{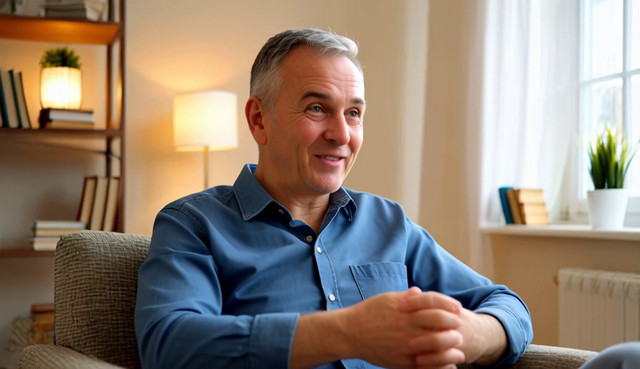}{\includegraphics[width=\extendedQualFrameWidth]{assets/extended-qualitative/prompt_000338/causal_forcing/frame_060_3p75s.jpg}}
\mfDeclareImage{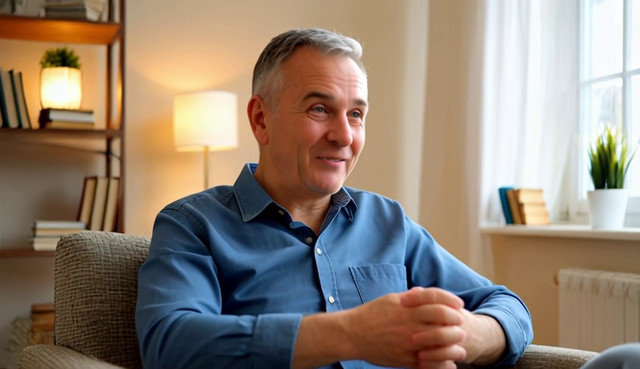}{\includegraphics[width=\extendedQualFrameWidth]{assets/extended-qualitative/prompt_000338/causal_forcing/frame_080_5s.jpg}}
\mfDeclareImage{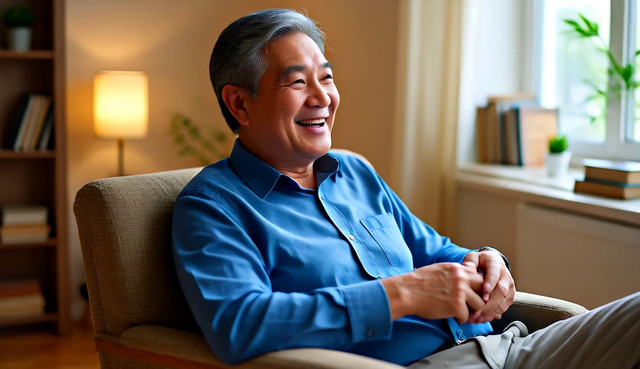}{\includegraphics[width=\extendedQualFrameWidth]{assets/extended-qualitative/prompt_000338/ours/frame_000_0s.jpg}}
\mfDeclareImage{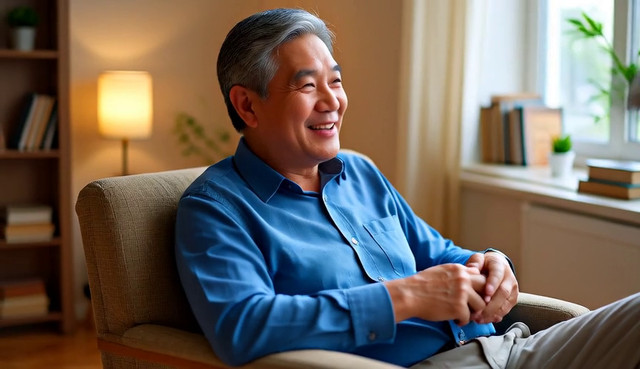}{\includegraphics[width=\extendedQualFrameWidth]{assets/extended-qualitative/prompt_000338/ours/frame_020_1p25s.jpg}}
\mfDeclareImage{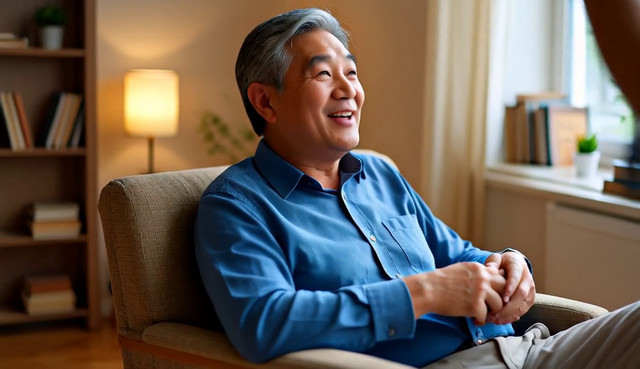}{\includegraphics[width=\extendedQualFrameWidth]{assets/extended-qualitative/prompt_000338/ours/frame_040_2p5s.jpg}}
\mfDeclareImage{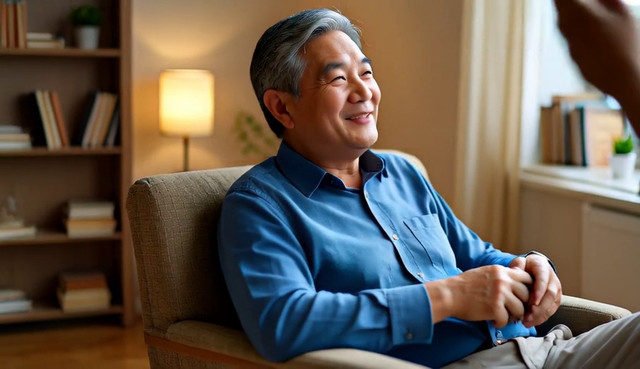}{\includegraphics[width=\extendedQualFrameWidth]{assets/extended-qualitative/prompt_000338/ours/frame_060_3p75s.jpg}}
\mfDeclareImage{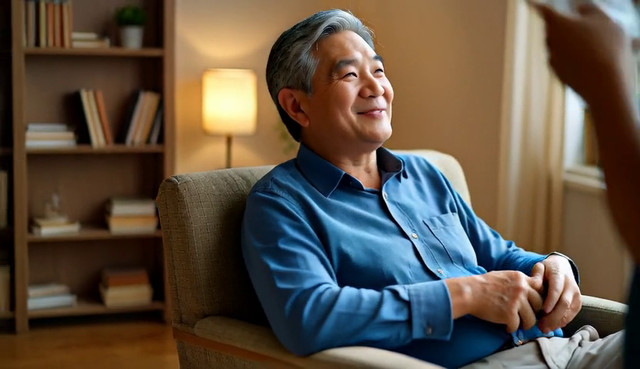}{\includegraphics[width=\extendedQualFrameWidth]{assets/extended-qualitative/prompt_000338/ours/frame_080_5s.jpg}}
\mfDeclareImage{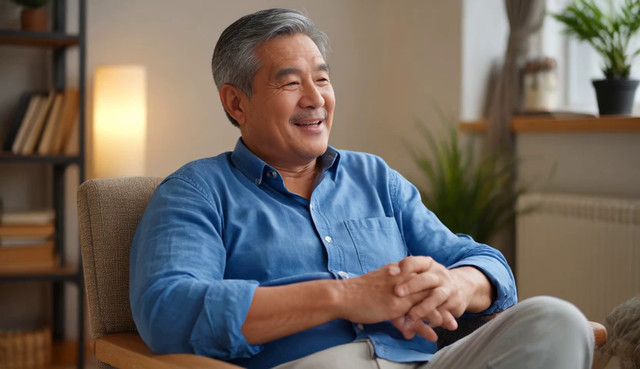}{\includegraphics[width=\extendedQualFrameWidth]{assets/extended-qualitative/prompt_000338/self_forcing/frame_000_0s.jpg}}
\mfDeclareImage{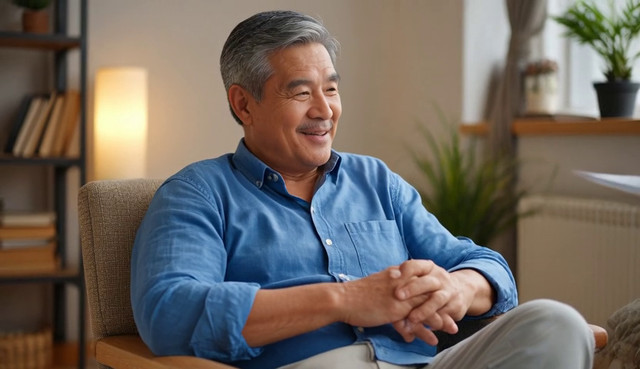}{\includegraphics[width=\extendedQualFrameWidth]{assets/extended-qualitative/prompt_000338/self_forcing/frame_020_1p25s.jpg}}
\mfDeclareImage{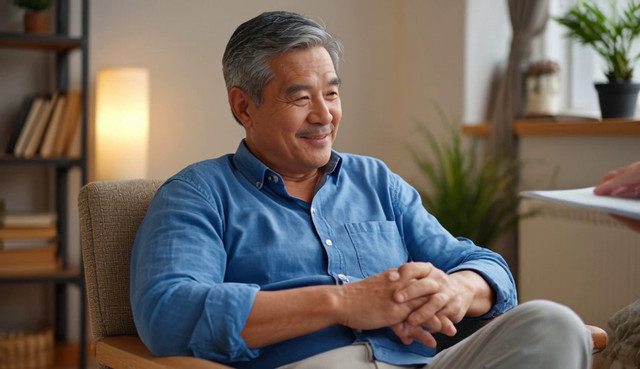}{\includegraphics[width=\extendedQualFrameWidth]{assets/extended-qualitative/prompt_000338/self_forcing/frame_040_2p5s.jpg}}
\mfDeclareImage{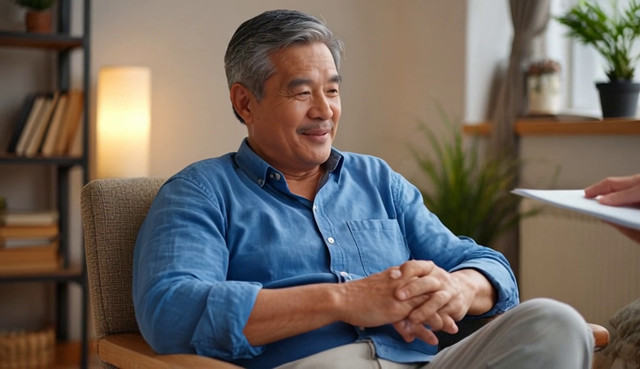}{\includegraphics[width=\extendedQualFrameWidth]{assets/extended-qualitative/prompt_000338/self_forcing/frame_060_3p75s.jpg}}
\mfDeclareImage{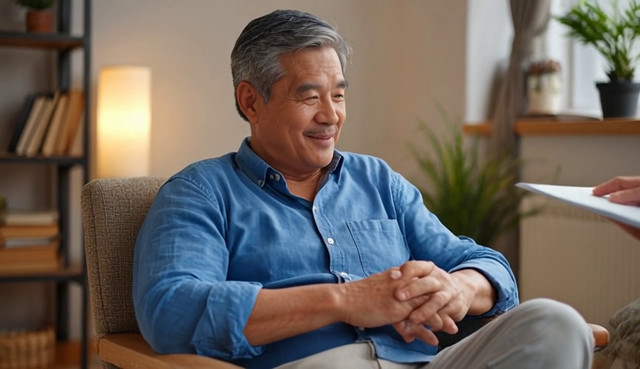}{\includegraphics[width=\extendedQualFrameWidth]{assets/extended-qualitative/prompt_000338/self_forcing/frame_080_5s.jpg}}
\mfDeclareImage{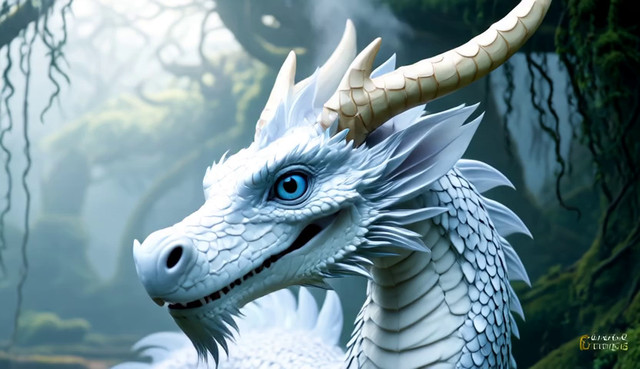}{\includegraphics[width=\extendedQualFrameWidth]{assets/extended-qualitative/prompt_000391/causal_forcing/frame_000_0s.jpg}}
\mfDeclareImage{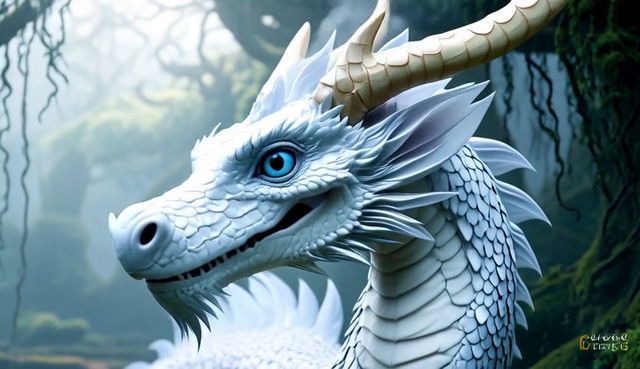}{\includegraphics[width=\extendedQualFrameWidth]{assets/extended-qualitative/prompt_000391/causal_forcing/frame_020_1p25s.jpg}}
\mfDeclareImage{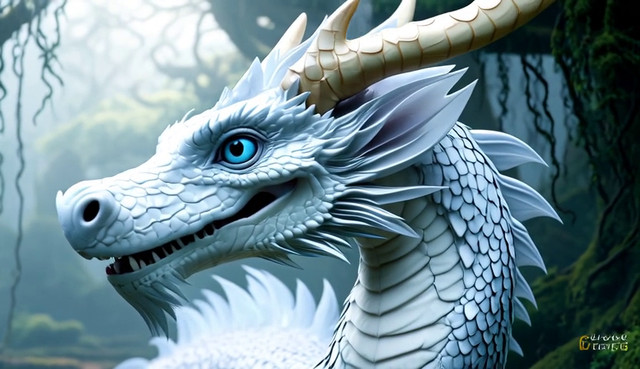}{\includegraphics[width=\extendedQualFrameWidth]{assets/extended-qualitative/prompt_000391/causal_forcing/frame_040_2p5s.jpg}}
\mfDeclareImage{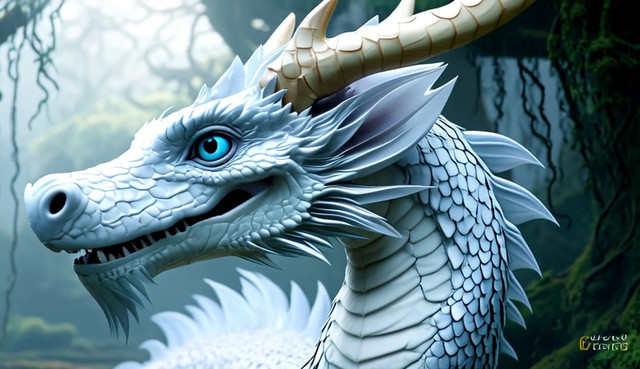}{\includegraphics[width=\extendedQualFrameWidth]{assets/extended-qualitative/prompt_000391/causal_forcing/frame_060_3p75s.jpg}}
\mfDeclareImage{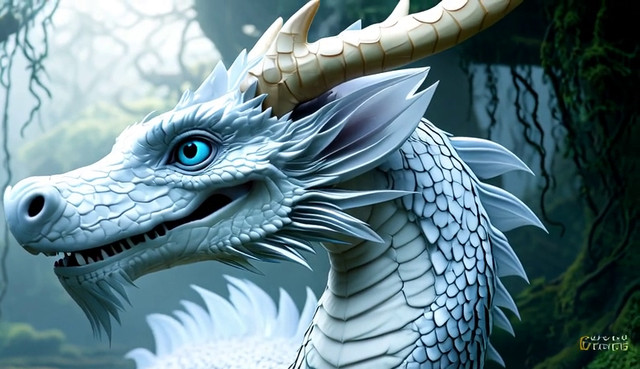}{\includegraphics[width=\extendedQualFrameWidth]{assets/extended-qualitative/prompt_000391/causal_forcing/frame_080_5s.jpg}}
\mfDeclareImage{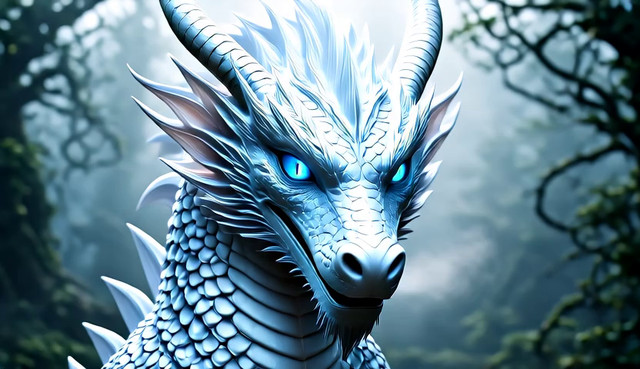}{\includegraphics[width=\extendedQualFrameWidth]{assets/extended-qualitative/prompt_000391/ours/frame_000_0s.jpg}}
\mfDeclareImage{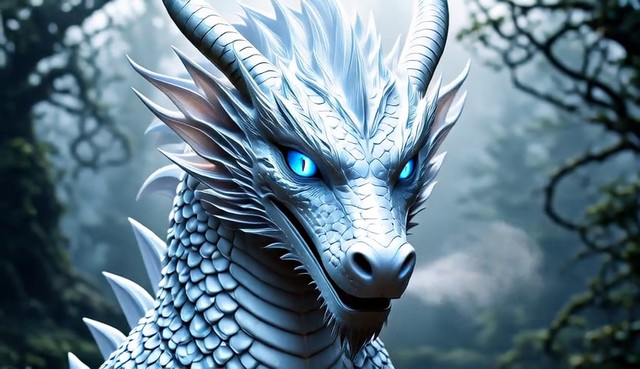}{\includegraphics[width=\extendedQualFrameWidth]{assets/extended-qualitative/prompt_000391/ours/frame_020_1p25s.jpg}}
\mfDeclareImage{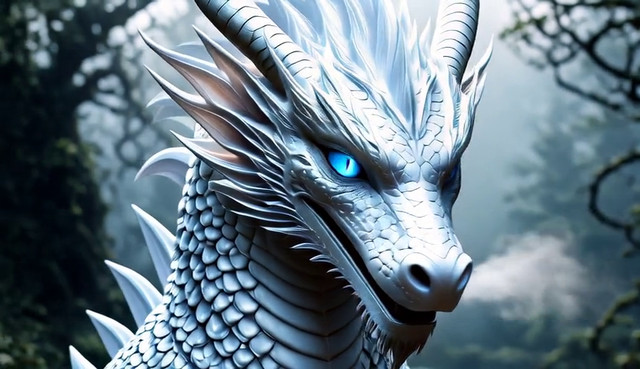}{\includegraphics[width=\extendedQualFrameWidth]{assets/extended-qualitative/prompt_000391/ours/frame_040_2p5s.jpg}}
\mfDeclareImage{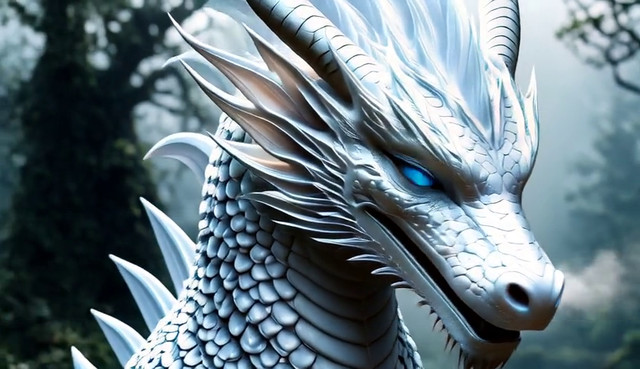}{\includegraphics[width=\extendedQualFrameWidth]{assets/extended-qualitative/prompt_000391/ours/frame_060_3p75s.jpg}}
\mfDeclareImage{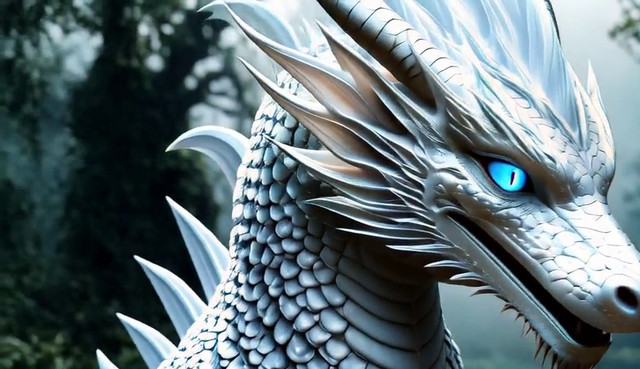}{\includegraphics[width=\extendedQualFrameWidth]{assets/extended-qualitative/prompt_000391/ours/frame_080_5s.jpg}}
\mfDeclareImage{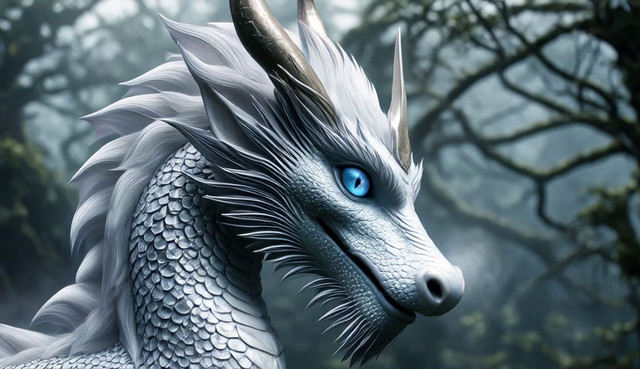}{\includegraphics[width=\extendedQualFrameWidth]{assets/extended-qualitative/prompt_000391/self_forcing/frame_000_0s.jpg}}
\mfDeclareImage{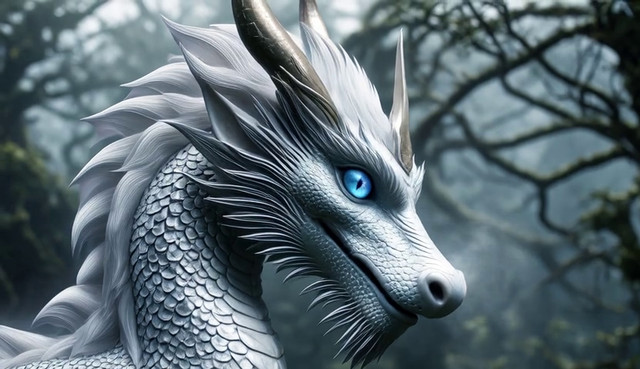}{\includegraphics[width=\extendedQualFrameWidth]{assets/extended-qualitative/prompt_000391/self_forcing/frame_020_1p25s.jpg}}
\mfDeclareImage{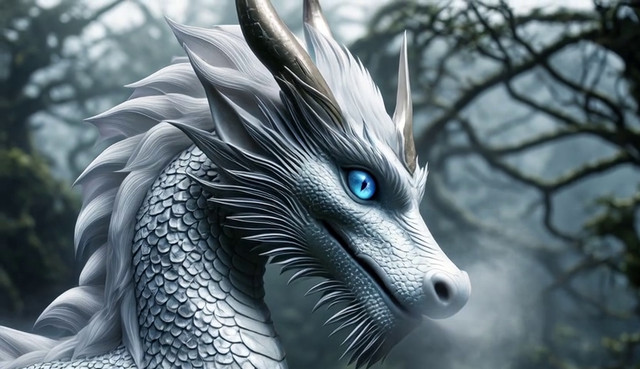}{\includegraphics[width=\extendedQualFrameWidth]{assets/extended-qualitative/prompt_000391/self_forcing/frame_040_2p5s.jpg}}
\mfDeclareImage{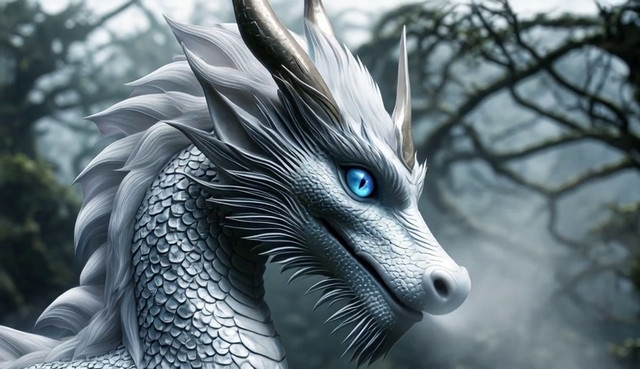}{\includegraphics[width=\extendedQualFrameWidth]{assets/extended-qualitative/prompt_000391/self_forcing/frame_060_3p75s.jpg}}
\mfDeclareImage{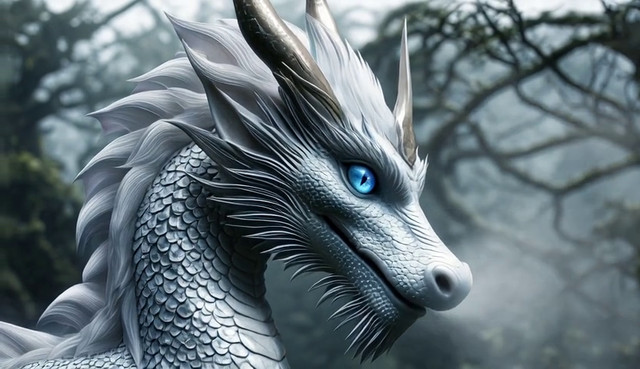}{\includegraphics[width=\extendedQualFrameWidth]{assets/extended-qualitative/prompt_000391/self_forcing/frame_080_5s.jpg}}
\mfDeclareImage{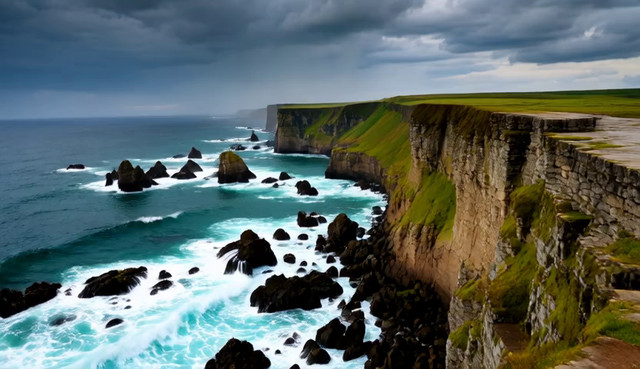}{\includegraphics[width=\extendedQualFrameWidth]{assets/extended-qualitative/prompt_000823/causal_forcing/frame_000_0s.jpg}}
\mfDeclareImage{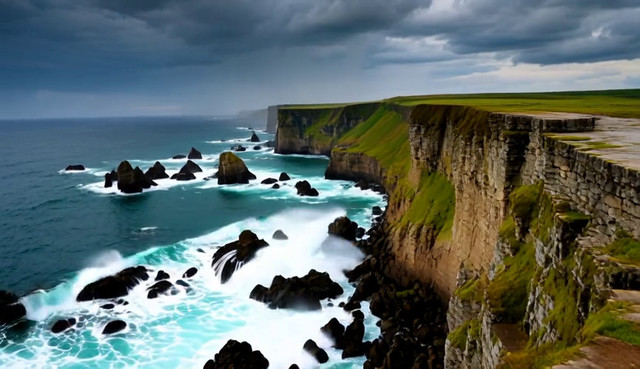}{\includegraphics[width=\extendedQualFrameWidth]{assets/extended-qualitative/prompt_000823/causal_forcing/frame_020_1p25s.jpg}}
\mfDeclareImage{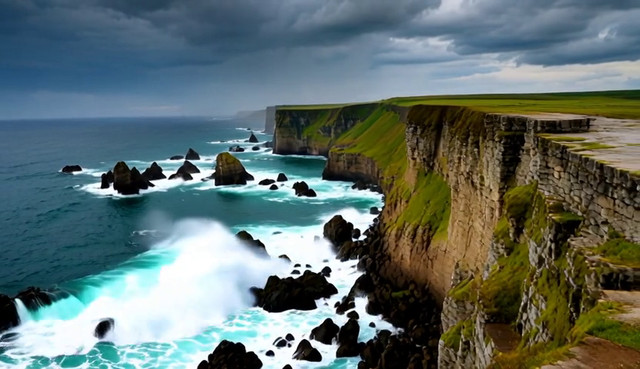}{\includegraphics[width=\extendedQualFrameWidth]{assets/extended-qualitative/prompt_000823/causal_forcing/frame_040_2p5s.jpg}}
\mfDeclareImage{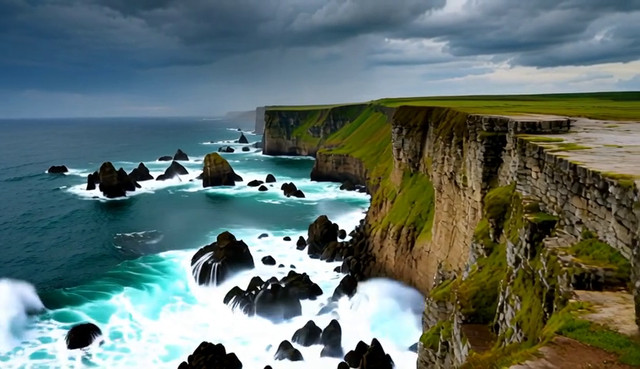}{\includegraphics[width=\extendedQualFrameWidth]{assets/extended-qualitative/prompt_000823/causal_forcing/frame_060_3p75s.jpg}}
\mfDeclareImage{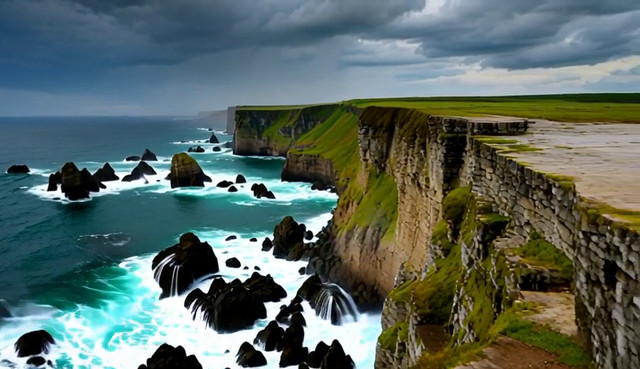}{\includegraphics[width=\extendedQualFrameWidth]{assets/extended-qualitative/prompt_000823/causal_forcing/frame_080_5s.jpg}}
\mfDeclareImage{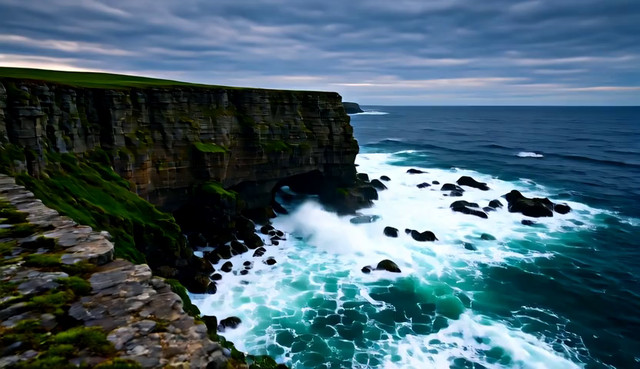}{\includegraphics[width=\extendedQualFrameWidth]{assets/extended-qualitative/prompt_000823/ours/frame_000_0s.jpg}}
\mfDeclareImage{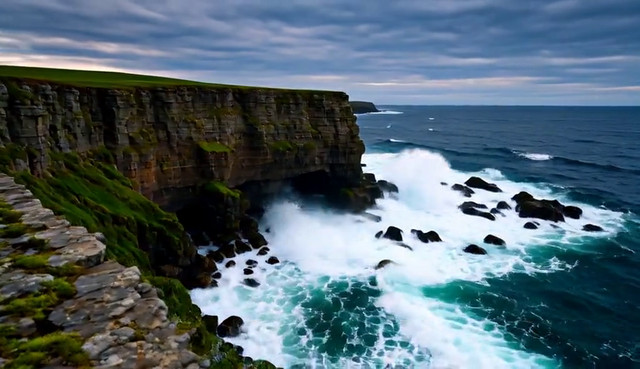}{\includegraphics[width=\extendedQualFrameWidth]{assets/extended-qualitative/prompt_000823/ours/frame_020_1p25s.jpg}}
\mfDeclareImage{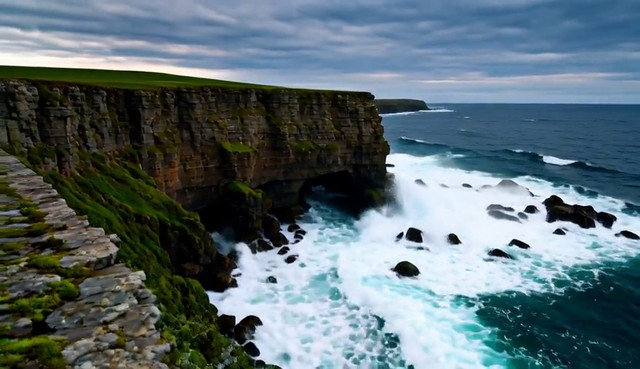}{\includegraphics[width=\extendedQualFrameWidth]{assets/extended-qualitative/prompt_000823/ours/frame_040_2p5s.jpg}}
\mfDeclareImage{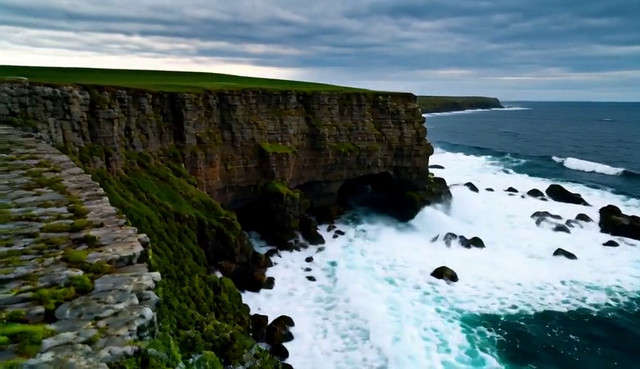}{\includegraphics[width=\extendedQualFrameWidth]{assets/extended-qualitative/prompt_000823/ours/frame_060_3p75s.jpg}}
\mfDeclareImage{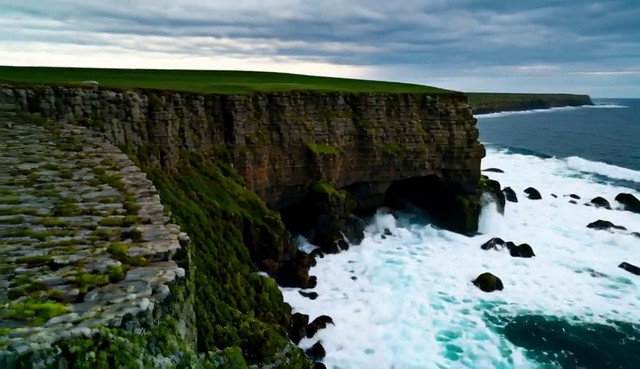}{\includegraphics[width=\extendedQualFrameWidth]{assets/extended-qualitative/prompt_000823/ours/frame_080_5s.jpg}}
\mfDeclareImage{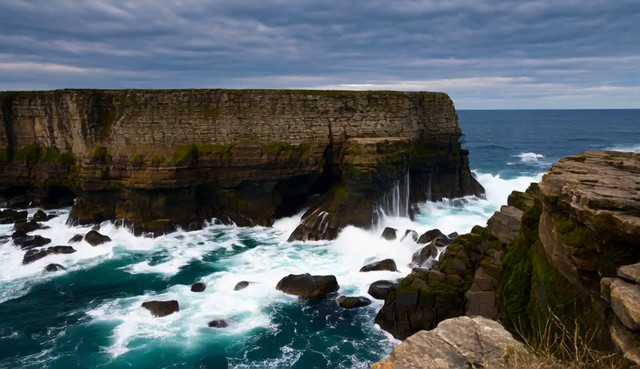}{\includegraphics[width=\extendedQualFrameWidth]{assets/extended-qualitative/prompt_000823/self_forcing/frame_000_0s.jpg}}
\mfDeclareImage{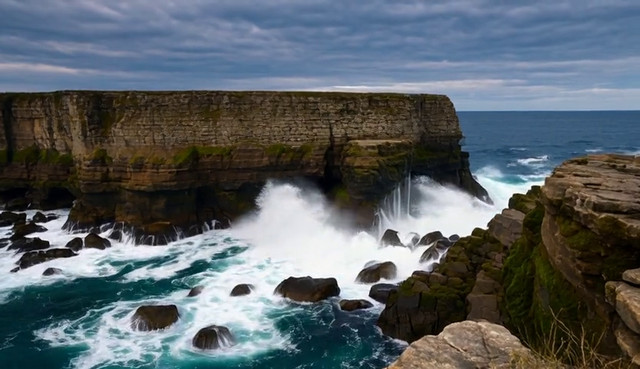}{\includegraphics[width=\extendedQualFrameWidth]{assets/extended-qualitative/prompt_000823/self_forcing/frame_020_1p25s.jpg}}
\mfDeclareImage{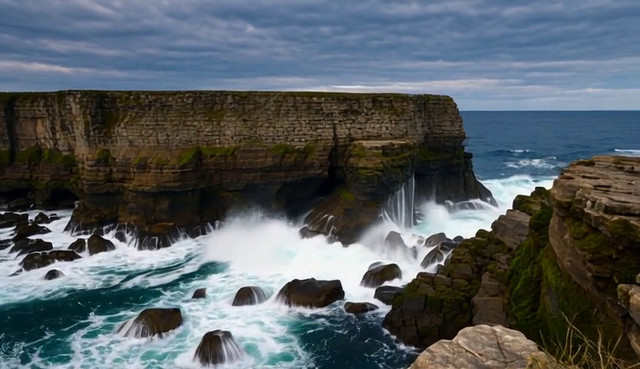}{\includegraphics[width=\extendedQualFrameWidth]{assets/extended-qualitative/prompt_000823/self_forcing/frame_040_2p5s.jpg}}
\mfDeclareImage{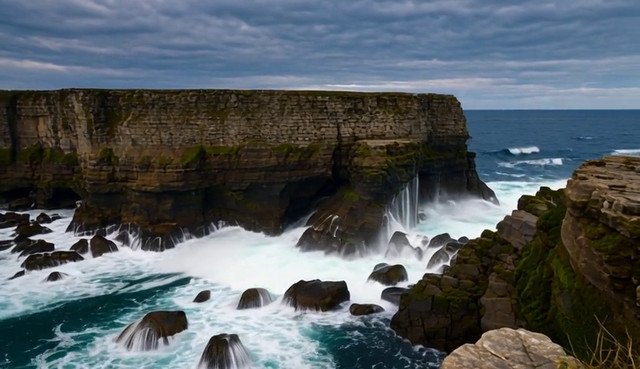}{\includegraphics[width=\extendedQualFrameWidth]{assets/extended-qualitative/prompt_000823/self_forcing/frame_060_3p75s.jpg}}
\mfDeclareImage{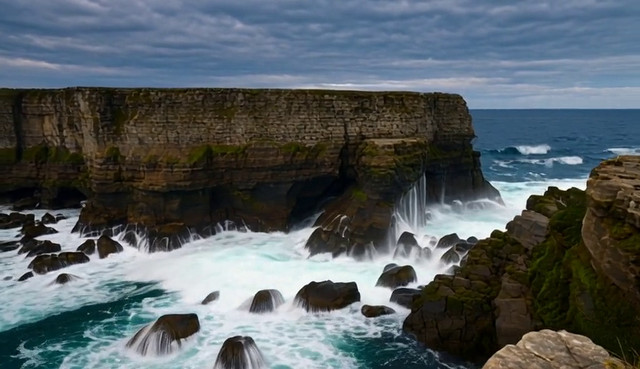}{\includegraphics[width=\extendedQualFrameWidth]{assets/extended-qualitative/prompt_000823/self_forcing/frame_080_5s.jpg}}
\mfDeclareImage{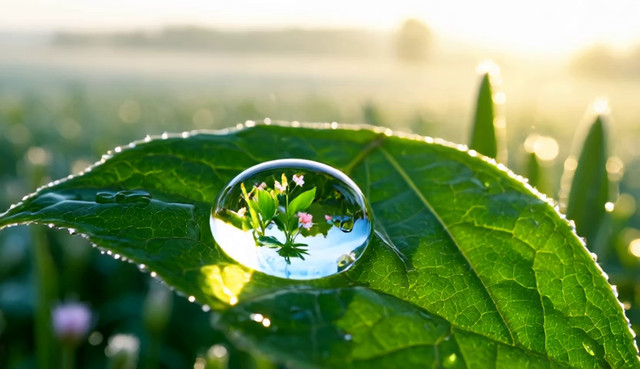}{\includegraphics[width=\extendedQualFrameWidth]{assets/extended-qualitative/prompt_000827/causal_forcing/frame_000_0s.jpg}}
\mfDeclareImage{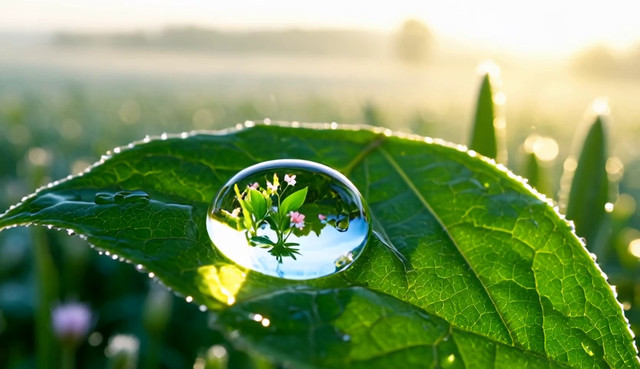}{\includegraphics[width=\extendedQualFrameWidth]{assets/extended-qualitative/prompt_000827/causal_forcing/frame_020_1p25s.jpg}}
\mfDeclareImage{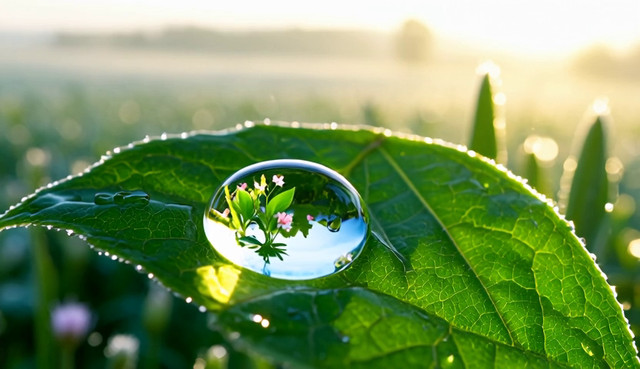}{\includegraphics[width=\extendedQualFrameWidth]{assets/extended-qualitative/prompt_000827/causal_forcing/frame_040_2p5s.jpg}}
\mfDeclareImage{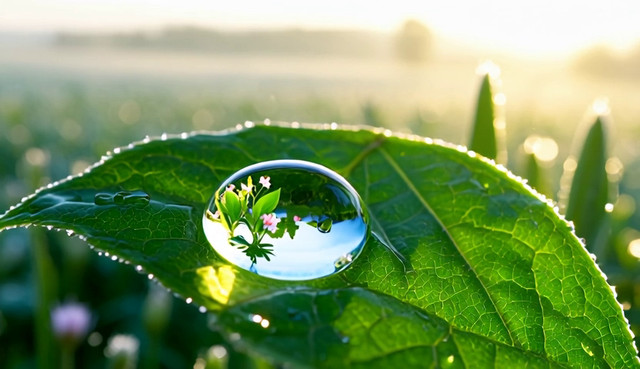}{\includegraphics[width=\extendedQualFrameWidth]{assets/extended-qualitative/prompt_000827/causal_forcing/frame_060_3p75s.jpg}}
\mfDeclareImage{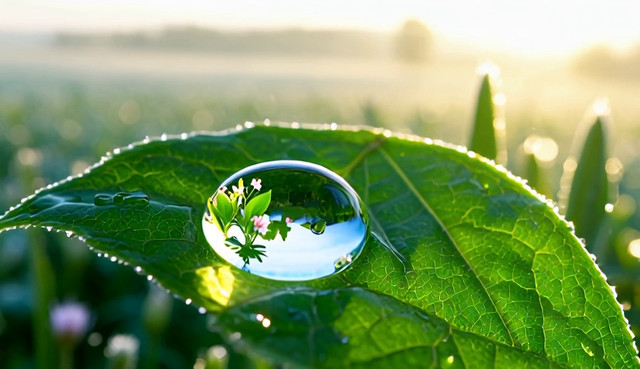}{\includegraphics[width=\extendedQualFrameWidth]{assets/extended-qualitative/prompt_000827/causal_forcing/frame_080_5s.jpg}}
\mfDeclareImage{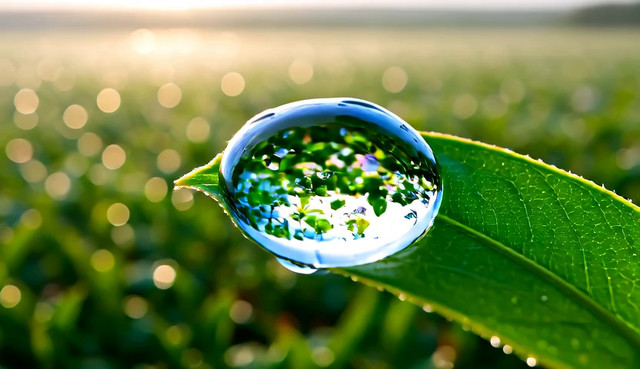}{\includegraphics[width=\extendedQualFrameWidth]{assets/extended-qualitative/prompt_000827/ours/frame_000_0s.jpg}}
\mfDeclareImage{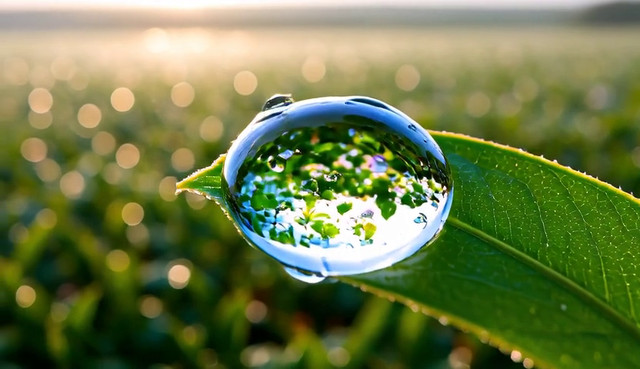}{\includegraphics[width=\extendedQualFrameWidth]{assets/extended-qualitative/prompt_000827/ours/frame_020_1p25s.jpg}}
\mfDeclareImage{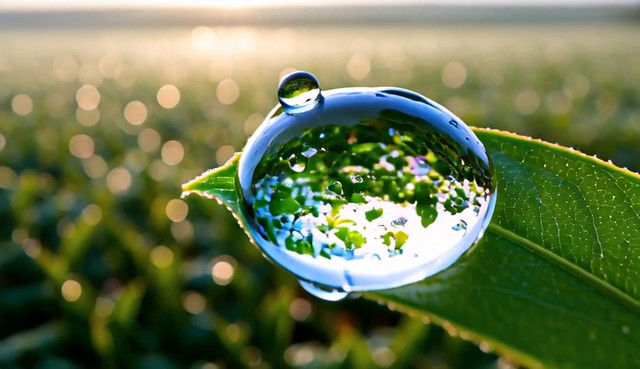}{\includegraphics[width=\extendedQualFrameWidth]{assets/extended-qualitative/prompt_000827/ours/frame_040_2p5s.jpg}}
\mfDeclareImage{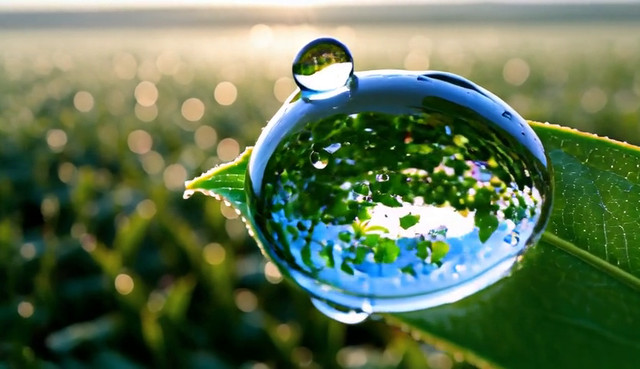}{\includegraphics[width=\extendedQualFrameWidth]{assets/extended-qualitative/prompt_000827/ours/frame_060_3p75s.jpg}}
\mfDeclareImage{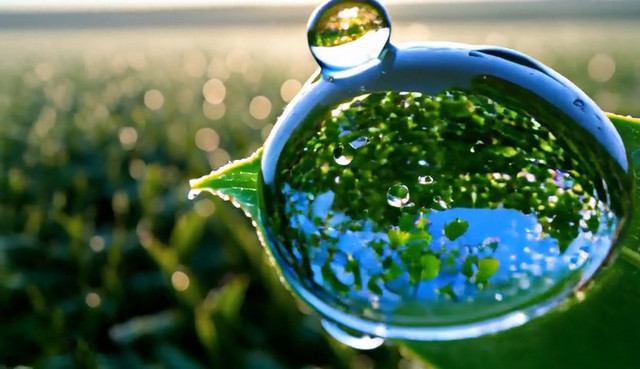}{\includegraphics[width=\extendedQualFrameWidth]{assets/extended-qualitative/prompt_000827/ours/frame_080_5s.jpg}}
\mfDeclareImage{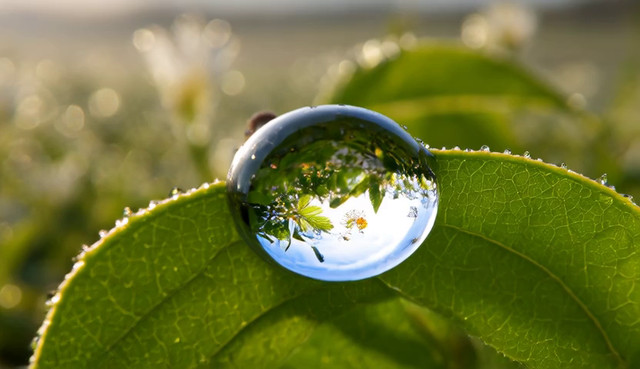}{\includegraphics[width=\extendedQualFrameWidth]{assets/extended-qualitative/prompt_000827/self_forcing/frame_000_0s.jpg}}
\mfDeclareImage{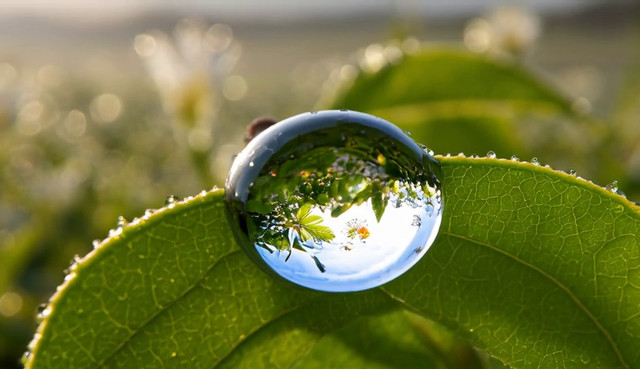}{\includegraphics[width=\extendedQualFrameWidth]{assets/extended-qualitative/prompt_000827/self_forcing/frame_020_1p25s.jpg}}
\mfDeclareImage{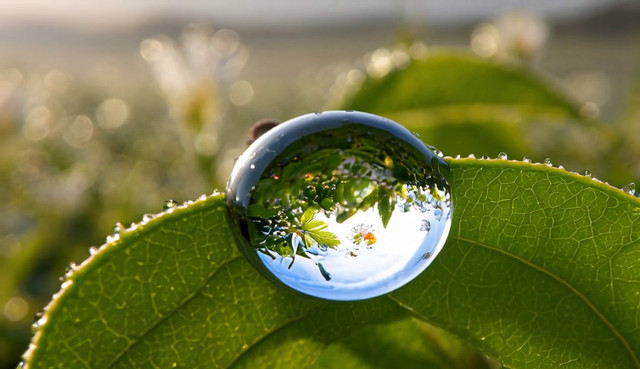}{\includegraphics[width=\extendedQualFrameWidth]{assets/extended-qualitative/prompt_000827/self_forcing/frame_040_2p5s.jpg}}
\mfDeclareImage{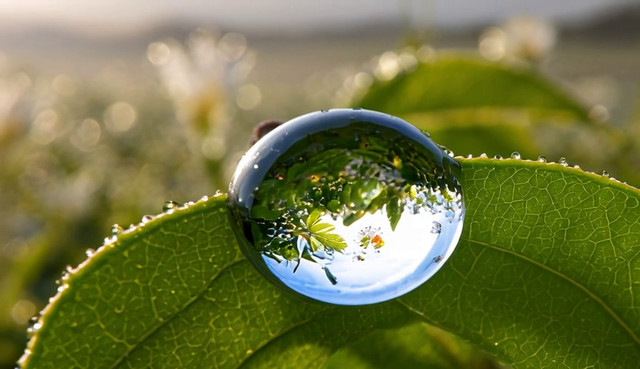}{\includegraphics[width=\extendedQualFrameWidth]{assets/extended-qualitative/prompt_000827/self_forcing/frame_060_3p75s.jpg}}
\mfDeclareImage{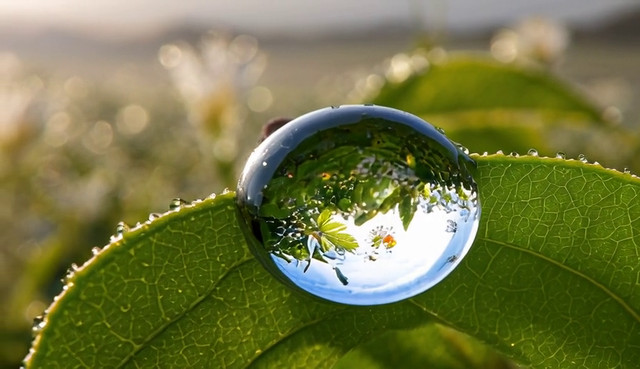}{\includegraphics[width=\extendedQualFrameWidth]{assets/extended-qualitative/prompt_000827/self_forcing/frame_080_5s.jpg}}
\mfDeclareImage{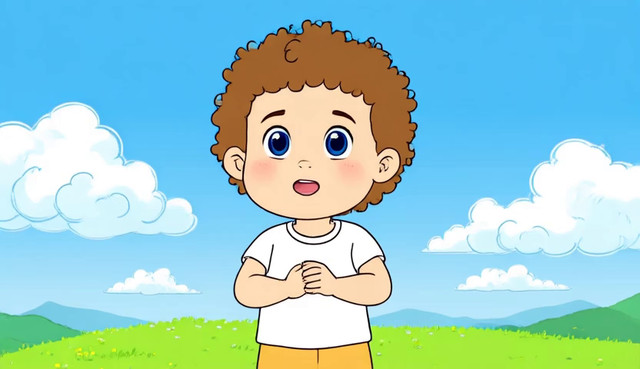}{\includegraphics[width=\extendedQualFrameWidth]{assets/extended-qualitative/prompt_000868/causal_forcing/frame_000_0s.jpg}}
\mfDeclareImage{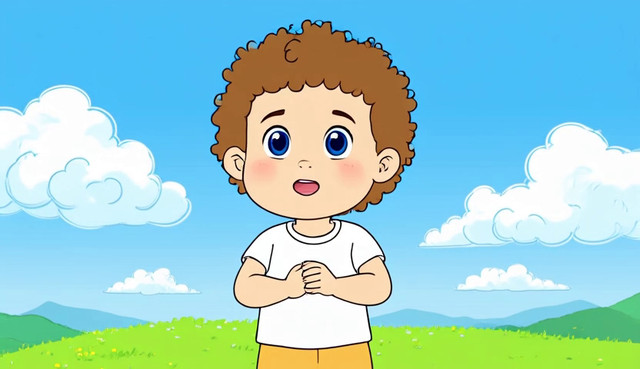}{\includegraphics[width=\extendedQualFrameWidth]{assets/extended-qualitative/prompt_000868/causal_forcing/frame_020_1p25s.jpg}}
\mfDeclareImage{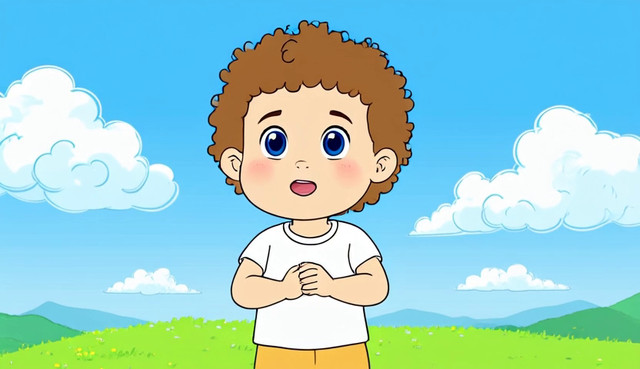}{\includegraphics[width=\extendedQualFrameWidth]{assets/extended-qualitative/prompt_000868/causal_forcing/frame_040_2p5s.jpg}}
\mfDeclareImage{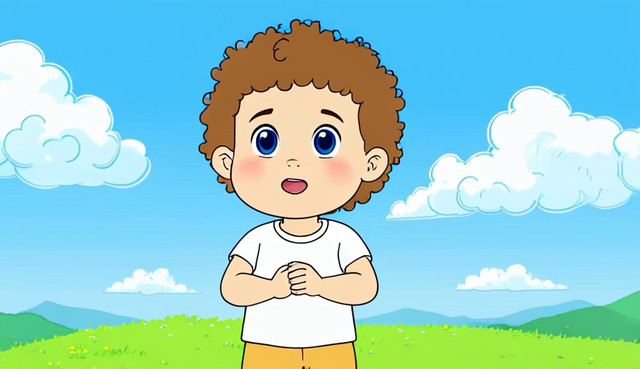}{\includegraphics[width=\extendedQualFrameWidth]{assets/extended-qualitative/prompt_000868/causal_forcing/frame_060_3p75s.jpg}}
\mfDeclareImage{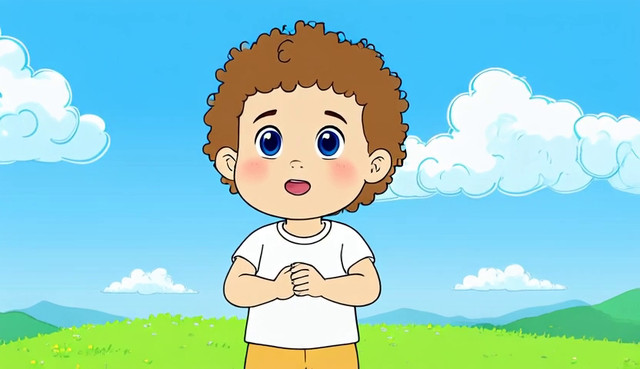}{\includegraphics[width=\extendedQualFrameWidth]{assets/extended-qualitative/prompt_000868/causal_forcing/frame_080_5s.jpg}}
\mfDeclareImage{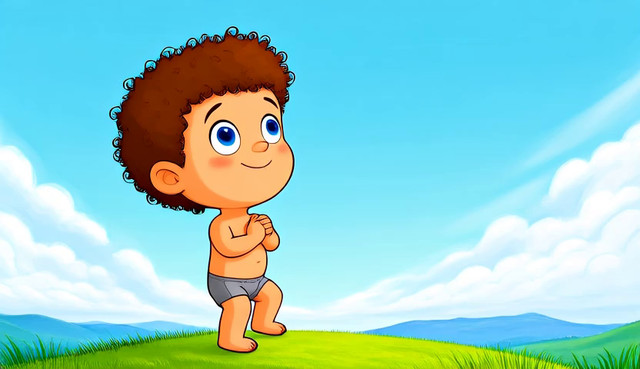}{\includegraphics[width=\extendedQualFrameWidth]{assets/extended-qualitative/prompt_000868/ours/frame_000_0s.jpg}}
\mfDeclareImage{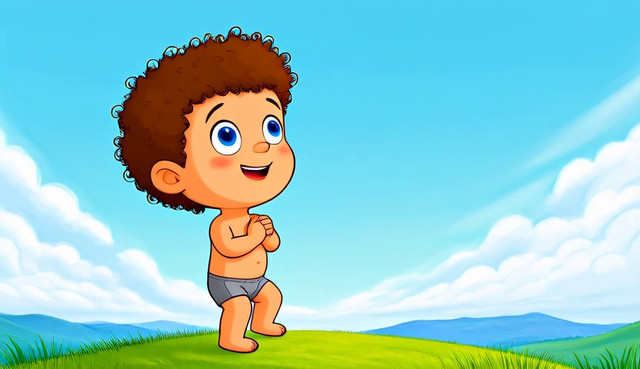}{\includegraphics[width=\extendedQualFrameWidth]{assets/extended-qualitative/prompt_000868/ours/frame_020_1p25s.jpg}}
\mfDeclareImage{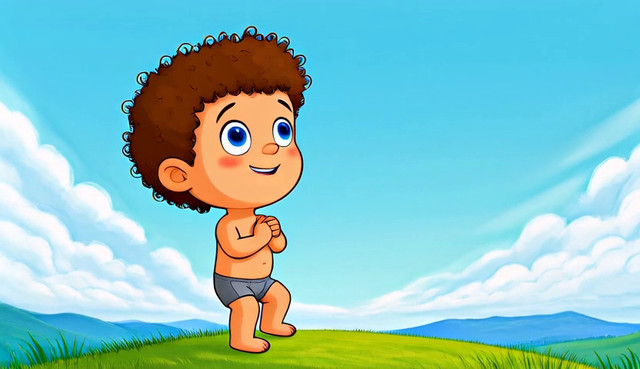}{\includegraphics[width=\extendedQualFrameWidth]{assets/extended-qualitative/prompt_000868/ours/frame_040_2p5s.jpg}}
\mfDeclareImage{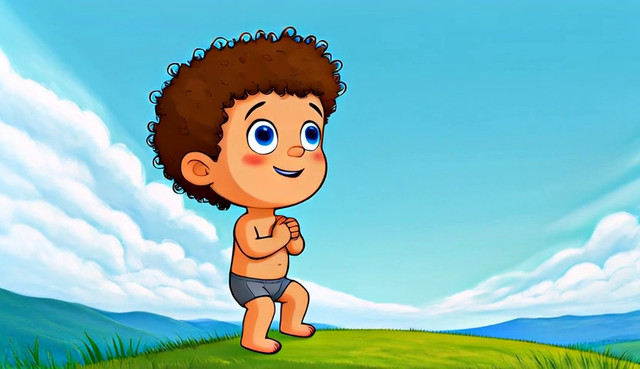}{\includegraphics[width=\extendedQualFrameWidth]{assets/extended-qualitative/prompt_000868/ours/frame_060_3p75s.jpg}}
\mfDeclareImage{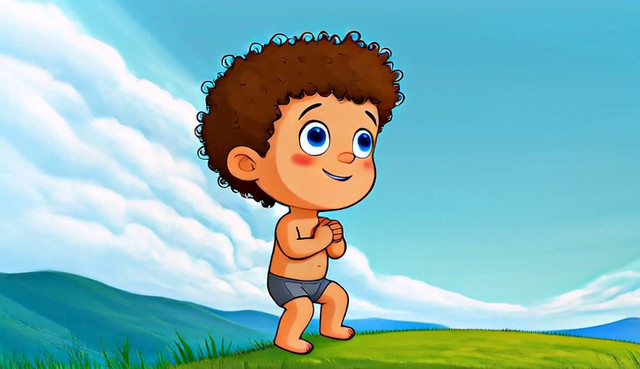}{\includegraphics[width=\extendedQualFrameWidth]{assets/extended-qualitative/prompt_000868/ours/frame_080_5s.jpg}}
\mfDeclareImage{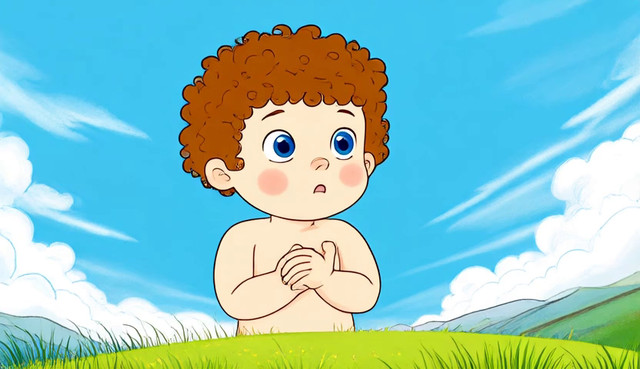}{\includegraphics[width=\extendedQualFrameWidth]{assets/extended-qualitative/prompt_000868/self_forcing/frame_000_0s.jpg}}
\mfDeclareImage{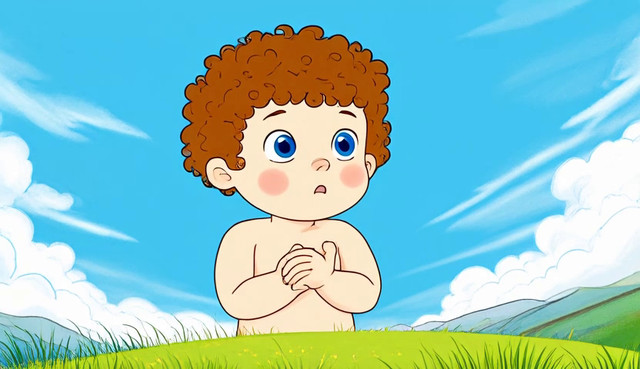}{\includegraphics[width=\extendedQualFrameWidth]{assets/extended-qualitative/prompt_000868/self_forcing/frame_020_1p25s.jpg}}
\mfDeclareImage{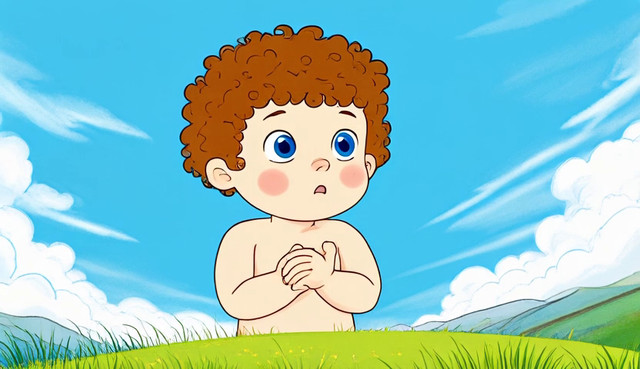}{\includegraphics[width=\extendedQualFrameWidth]{assets/extended-qualitative/prompt_000868/self_forcing/frame_040_2p5s.jpg}}
\mfDeclareImage{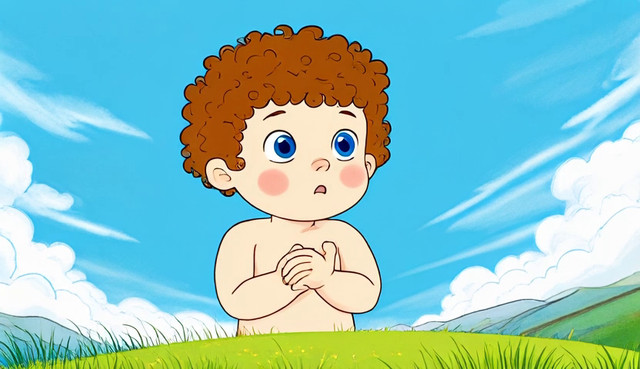}{\includegraphics[width=\extendedQualFrameWidth]{assets/extended-qualitative/prompt_000868/self_forcing/frame_060_3p75s.jpg}}
\mfDeclareImage{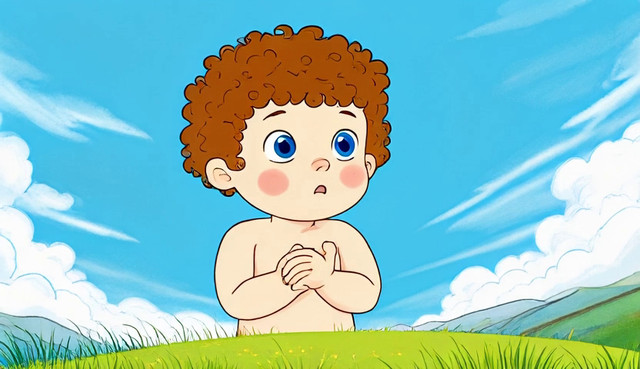}{\includegraphics[width=\extendedQualFrameWidth]{assets/extended-qualitative/prompt_000868/self_forcing/frame_080_5s.jpg}}
\mfDeclareImage{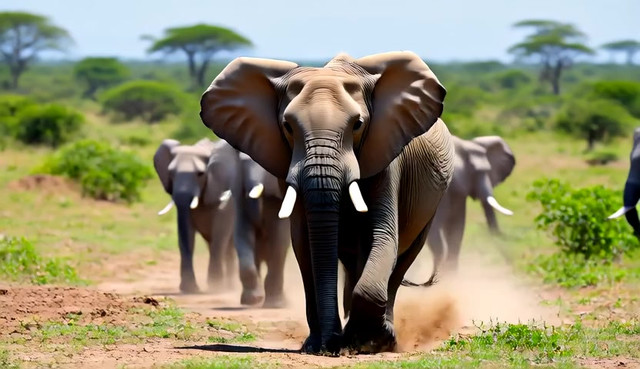}{\includegraphics[width=\ablationQualFrameWidth]{assets/qualitative-ablation/prompt_000319/endpoint_only/frame_000_0s.jpg}}
\mfDeclareImage{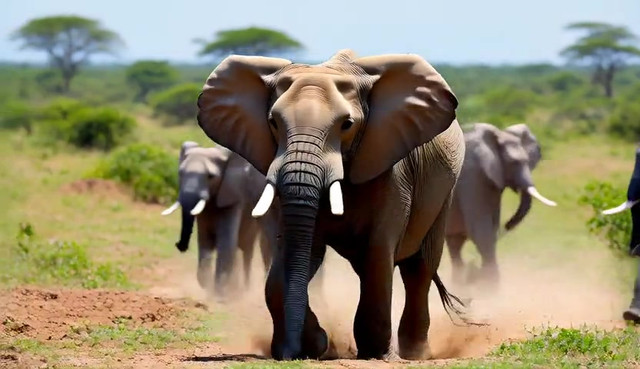}{\includegraphics[width=\ablationQualFrameWidth]{assets/qualitative-ablation/prompt_000319/endpoint_only/frame_020_1p25s.jpg}}
\mfDeclareImage{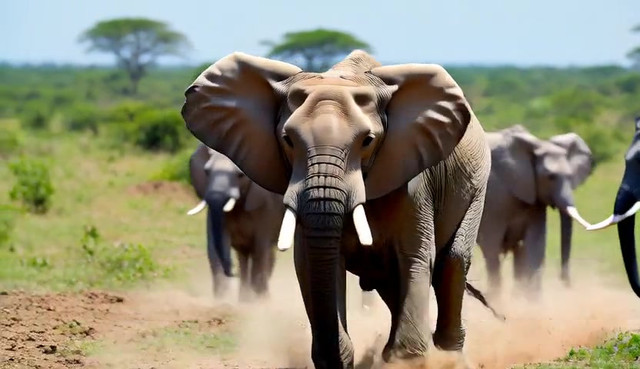}{\includegraphics[width=\ablationQualFrameWidth]{assets/qualitative-ablation/prompt_000319/endpoint_only/frame_040_2p5s.jpg}}
\mfDeclareImage{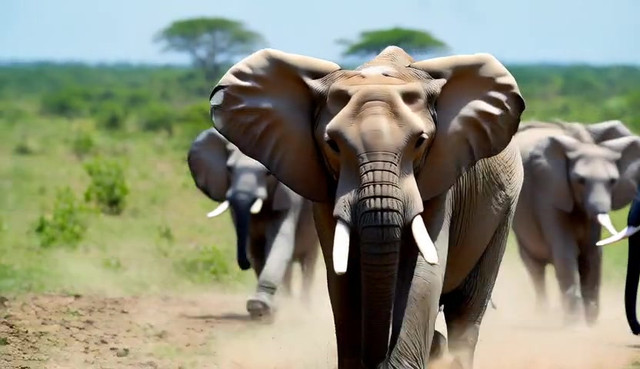}{\includegraphics[width=\ablationQualFrameWidth]{assets/qualitative-ablation/prompt_000319/endpoint_only/frame_060_3p75s.jpg}}
\mfDeclareImage{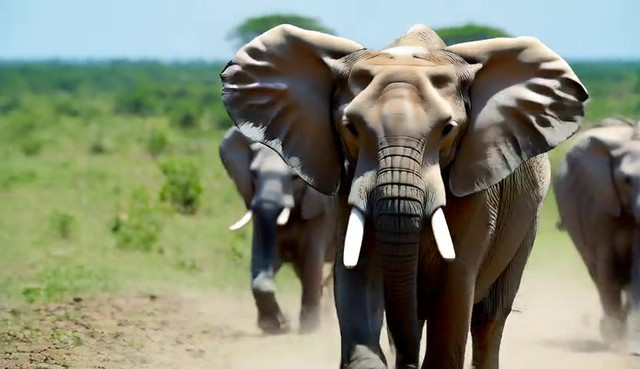}{\includegraphics[width=\ablationQualFrameWidth]{assets/qualitative-ablation/prompt_000319/endpoint_only/frame_080_5s.jpg}}
\mfDeclareImage{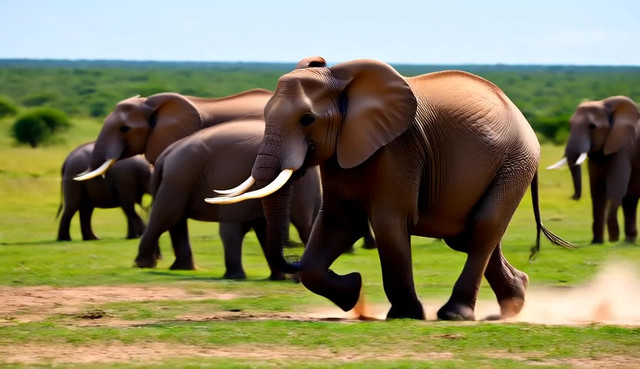}{\includegraphics[width=\ablationQualFrameWidth]{assets/qualitative-ablation/prompt_000319/k2/frame_000_0s.jpg}}
\mfDeclareImage{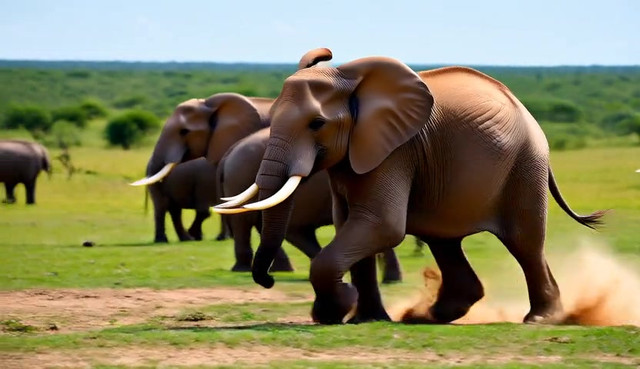}{\includegraphics[width=\ablationQualFrameWidth]{assets/qualitative-ablation/prompt_000319/k2/frame_020_1p25s.jpg}}
\mfDeclareImage{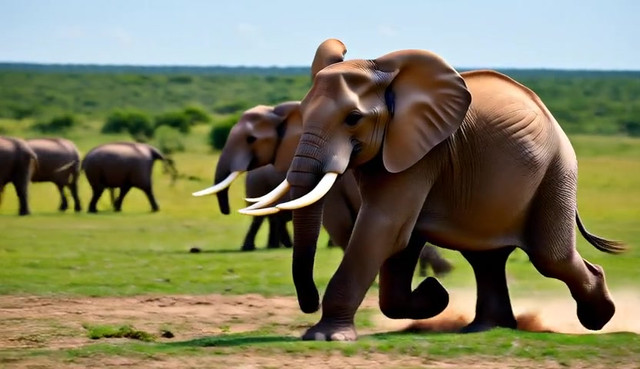}{\includegraphics[width=\ablationQualFrameWidth]{assets/qualitative-ablation/prompt_000319/k2/frame_040_2p5s.jpg}}
\mfDeclareImage{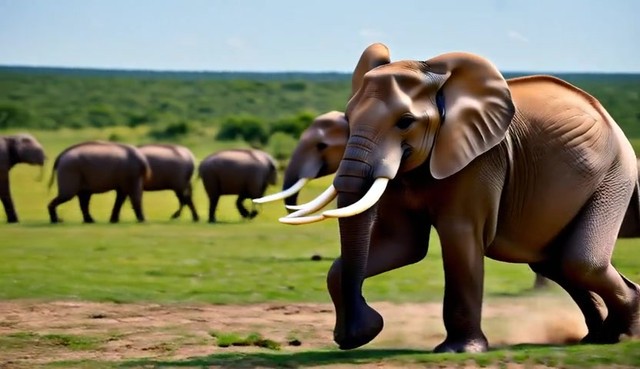}{\includegraphics[width=\ablationQualFrameWidth]{assets/qualitative-ablation/prompt_000319/k2/frame_060_3p75s.jpg}}
\mfDeclareImage{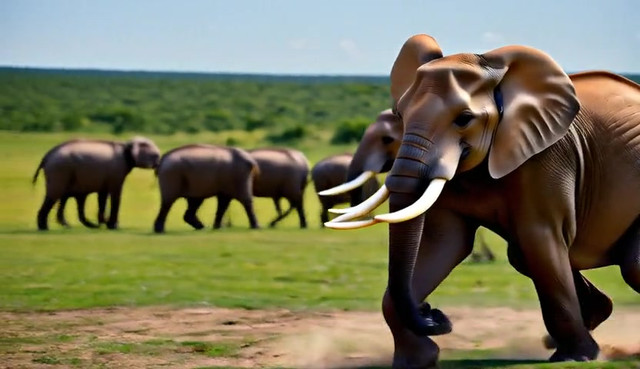}{\includegraphics[width=\ablationQualFrameWidth]{assets/qualitative-ablation/prompt_000319/k2/frame_080_5s.jpg}}
\mfDeclareImage{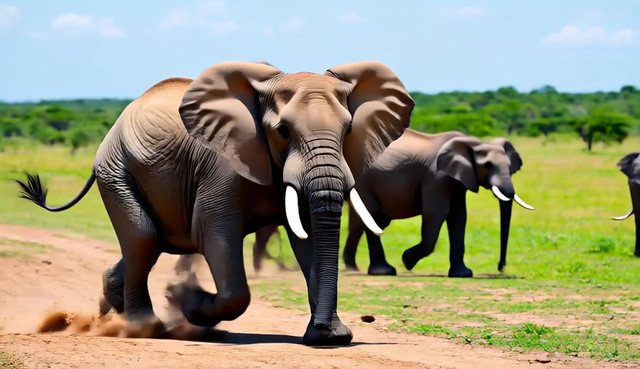}{\includegraphics[width=\ablationQualFrameWidth]{assets/qualitative-ablation/prompt_000319/lambda05/frame_000_0s.jpg}}
\mfDeclareImage{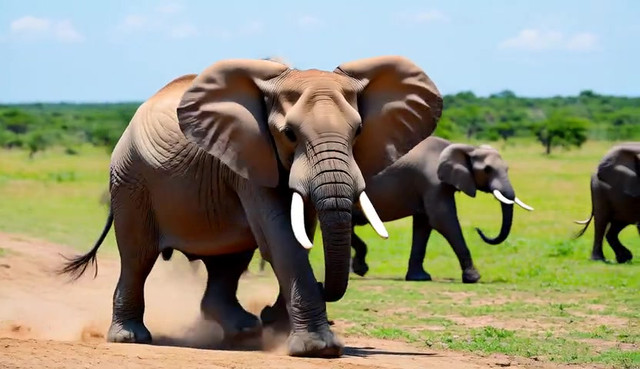}{\includegraphics[width=\ablationQualFrameWidth]{assets/qualitative-ablation/prompt_000319/lambda05/frame_020_1p25s.jpg}}
\mfDeclareImage{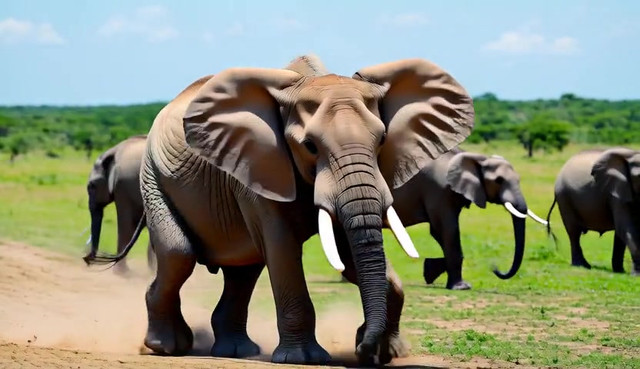}{\includegraphics[width=\ablationQualFrameWidth]{assets/qualitative-ablation/prompt_000319/lambda05/frame_040_2p5s.jpg}}
\mfDeclareImage{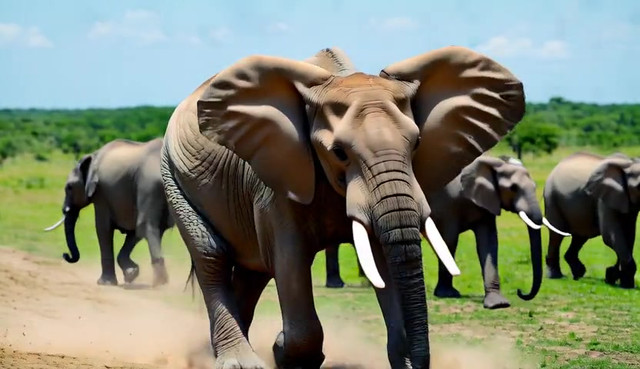}{\includegraphics[width=\ablationQualFrameWidth]{assets/qualitative-ablation/prompt_000319/lambda05/frame_060_3p75s.jpg}}
\mfDeclareImage{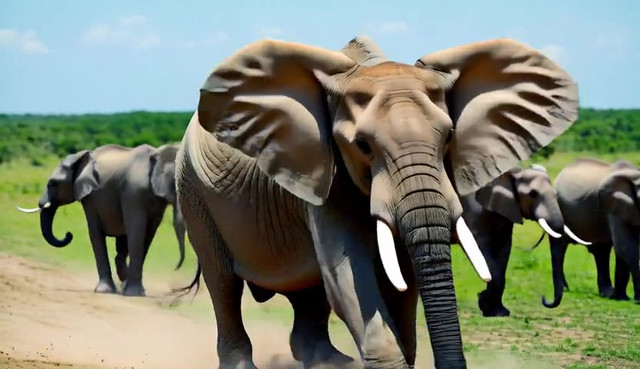}{\includegraphics[width=\ablationQualFrameWidth]{assets/qualitative-ablation/prompt_000319/lambda05/frame_080_5s.jpg}}
\mfDeclareImage{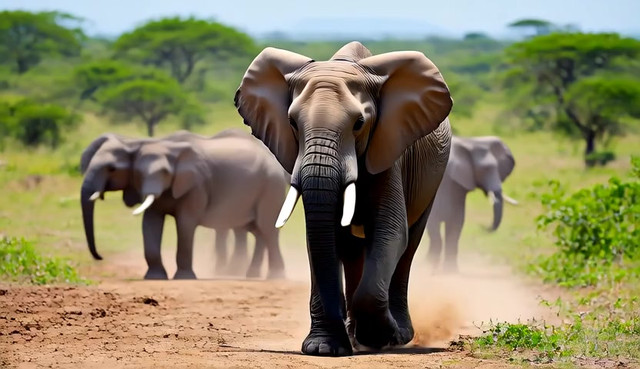}{\includegraphics[width=\ablationQualFrameWidth]{assets/qualitative-ablation/prompt_000319/lambda1/frame_000_0s.jpg}}
\mfDeclareImage{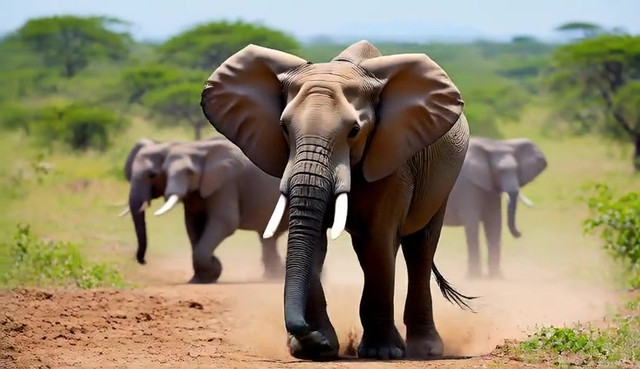}{\includegraphics[width=\ablationQualFrameWidth]{assets/qualitative-ablation/prompt_000319/lambda1/frame_020_1p25s.jpg}}
\mfDeclareImage{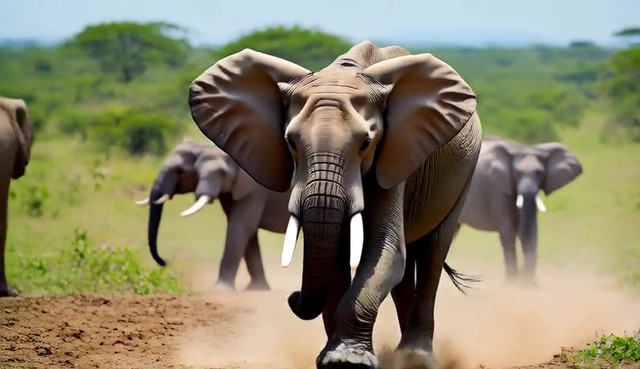}{\includegraphics[width=\ablationQualFrameWidth]{assets/qualitative-ablation/prompt_000319/lambda1/frame_040_2p5s.jpg}}
\mfDeclareImage{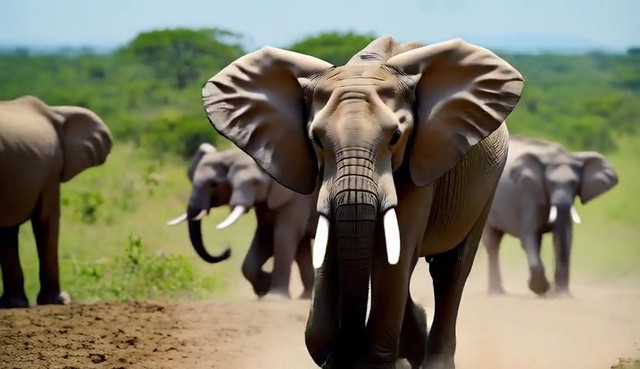}{\includegraphics[width=\ablationQualFrameWidth]{assets/qualitative-ablation/prompt_000319/lambda1/frame_060_3p75s.jpg}}
\mfDeclareImage{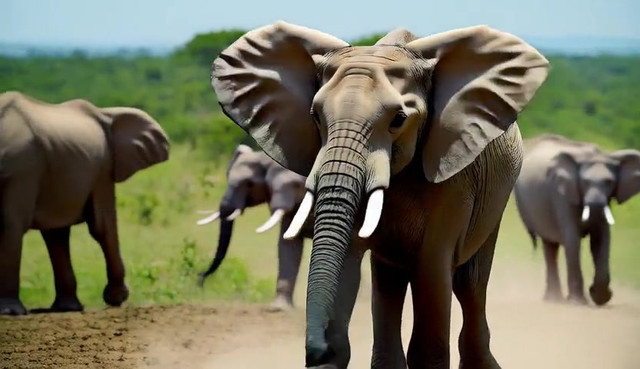}{\includegraphics[width=\ablationQualFrameWidth]{assets/qualitative-ablation/prompt_000319/lambda1/frame_080_5s.jpg}}
\mfDeclareImage{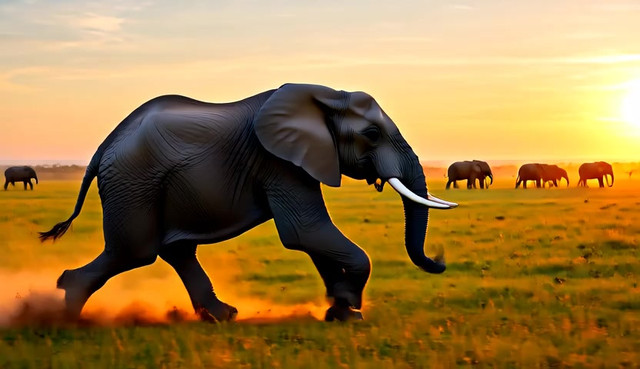}{\includegraphics[width=\ablationQualFrameWidth]{assets/qualitative-ablation/prompt_000319/ours/frame_000_0s.jpg}}
\mfDeclareImage{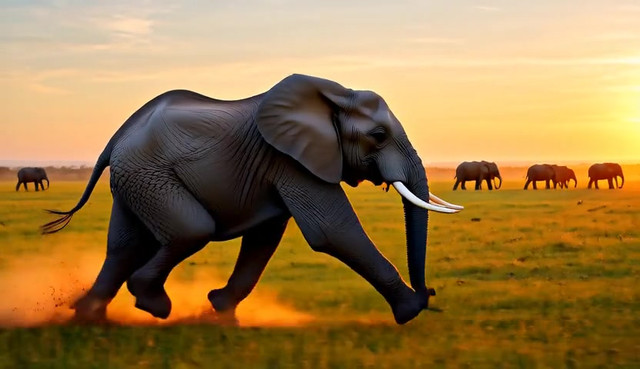}{\includegraphics[width=\ablationQualFrameWidth]{assets/qualitative-ablation/prompt_000319/ours/frame_020_1p25s.jpg}}
\mfDeclareImage{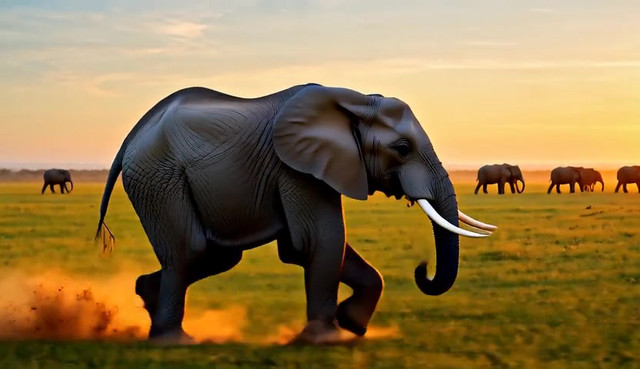}{\includegraphics[width=\ablationQualFrameWidth]{assets/qualitative-ablation/prompt_000319/ours/frame_040_2p5s.jpg}}
\mfDeclareImage{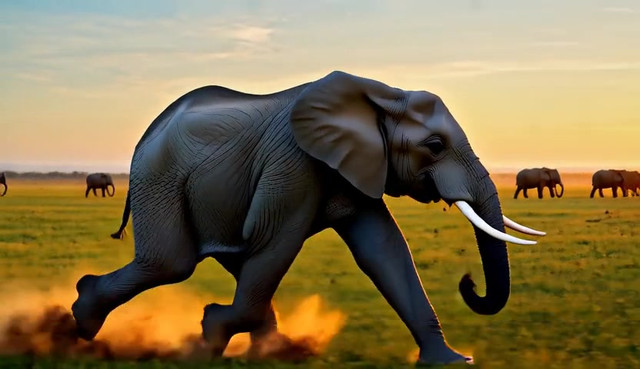}{\includegraphics[width=\ablationQualFrameWidth]{assets/qualitative-ablation/prompt_000319/ours/frame_060_3p75s.jpg}}
\mfDeclareImage{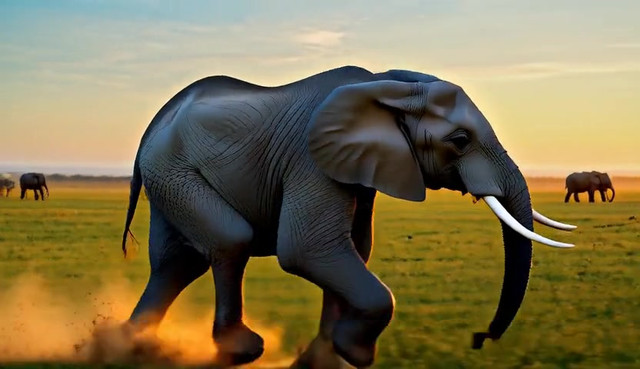}{\includegraphics[width=\ablationQualFrameWidth]{assets/qualitative-ablation/prompt_000319/ours/frame_080_5s.jpg}}
\mfDeclareImage{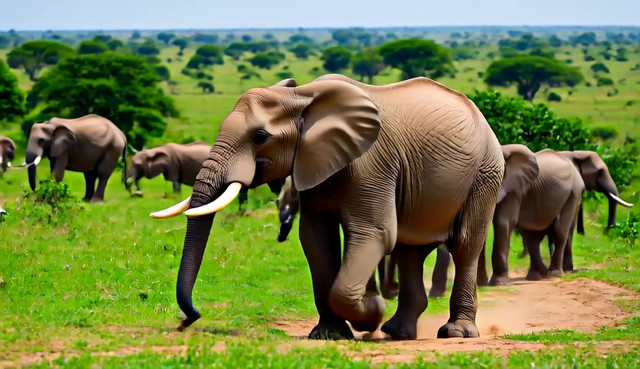}{\includegraphics[width=\ablationQualFrameWidth]{assets/qualitative-ablation/prompt_000319/shuffled_ratio/frame_000_0s.jpg}}
\mfDeclareImage{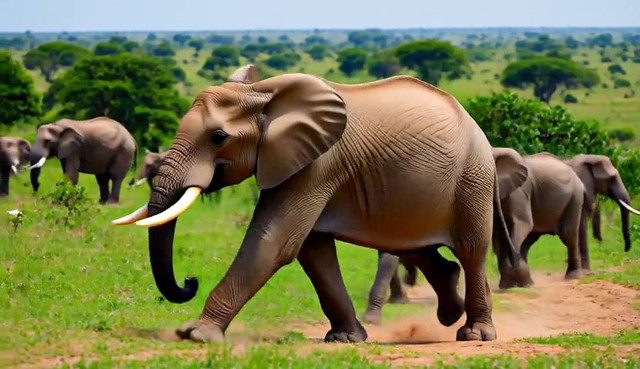}{\includegraphics[width=\ablationQualFrameWidth]{assets/qualitative-ablation/prompt_000319/shuffled_ratio/frame_020_1p25s.jpg}}
\mfDeclareImage{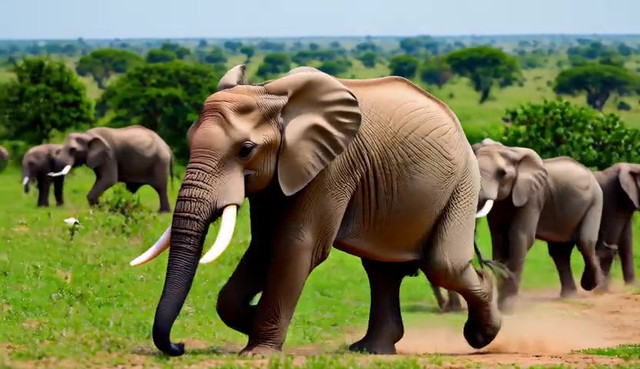}{\includegraphics[width=\ablationQualFrameWidth]{assets/qualitative-ablation/prompt_000319/shuffled_ratio/frame_040_2p5s.jpg}}
\mfDeclareImage{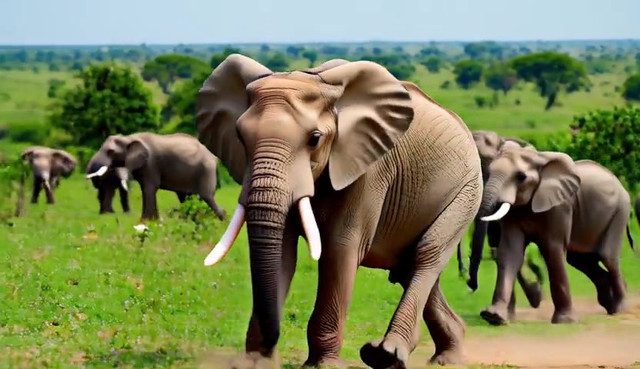}{\includegraphics[width=\ablationQualFrameWidth]{assets/qualitative-ablation/prompt_000319/shuffled_ratio/frame_060_3p75s.jpg}}
\mfDeclareImage{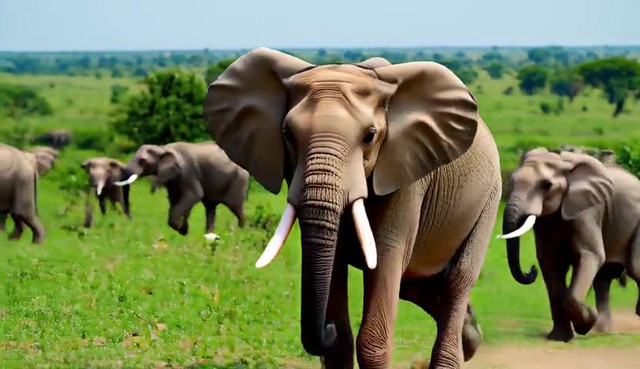}{\includegraphics[width=\ablationQualFrameWidth]{assets/qualitative-ablation/prompt_000319/shuffled_ratio/frame_080_5s.jpg}}
\mfDeclareImage{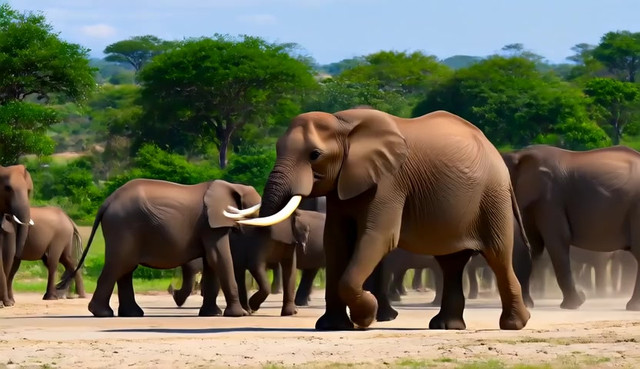}{\includegraphics[width=\ablationQualFrameWidth]{assets/qualitative-ablation/prompt_000319/wo_cfg_decoupling/frame_000_0s.jpg}}
\mfDeclareImage{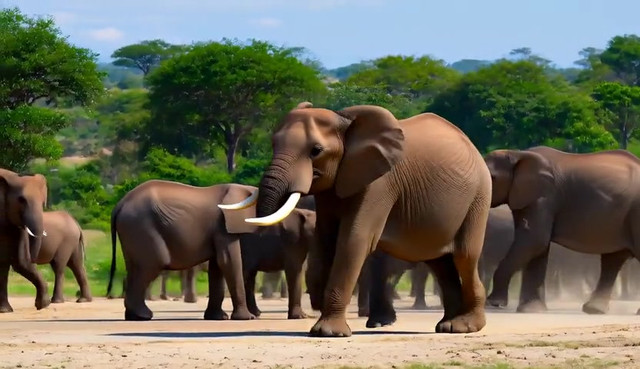}{\includegraphics[width=\ablationQualFrameWidth]{assets/qualitative-ablation/prompt_000319/wo_cfg_decoupling/frame_020_1p25s.jpg}}
\mfDeclareImage{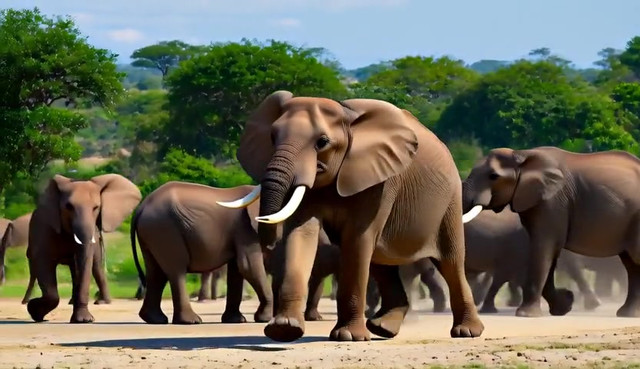}{\includegraphics[width=\ablationQualFrameWidth]{assets/qualitative-ablation/prompt_000319/wo_cfg_decoupling/frame_040_2p5s.jpg}}
\mfDeclareImage{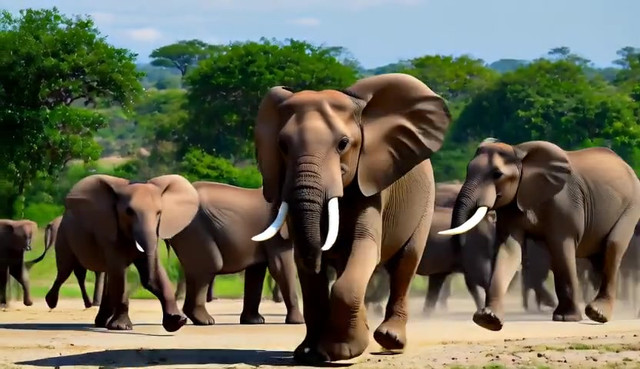}{\includegraphics[width=\ablationQualFrameWidth]{assets/qualitative-ablation/prompt_000319/wo_cfg_decoupling/frame_060_3p75s.jpg}}
\mfDeclareImage{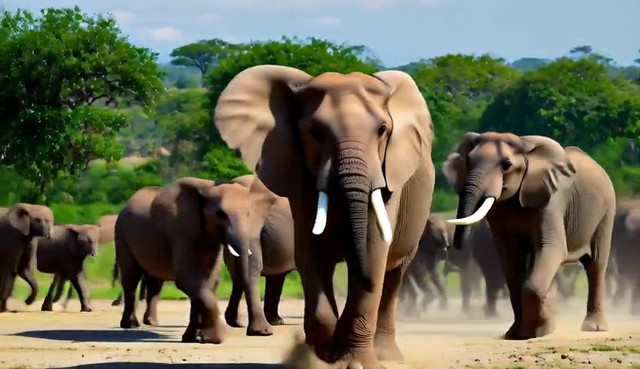}{\includegraphics[width=\ablationQualFrameWidth]{assets/qualitative-ablation/prompt_000319/wo_cfg_decoupling/frame_080_5s.jpg}}
\mfDeclareImage{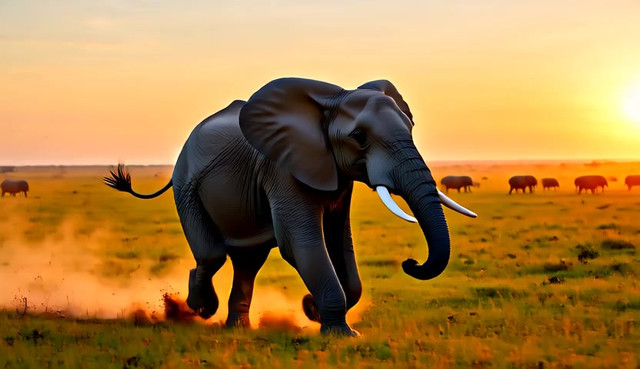}{\includegraphics[width=\ablationQualFrameWidth]{assets/qualitative-ablation/prompt_000319/wo_mmd/frame_000_0s.jpg}}
\mfDeclareImage{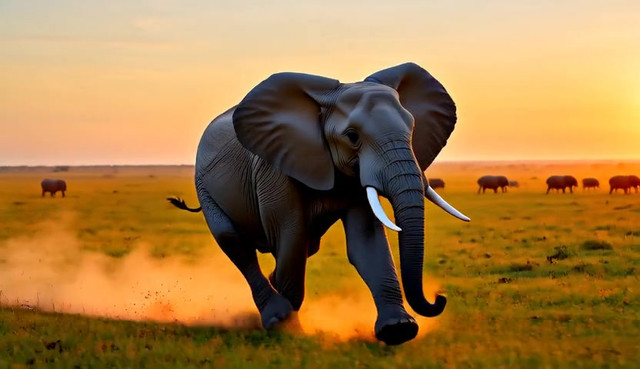}{\includegraphics[width=\ablationQualFrameWidth]{assets/qualitative-ablation/prompt_000319/wo_mmd/frame_020_1p25s.jpg}}
\mfDeclareImage{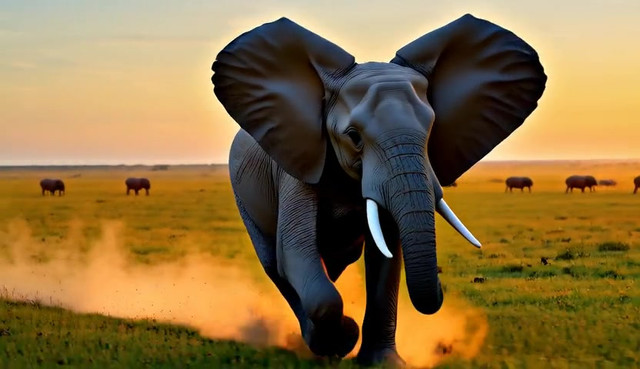}{\includegraphics[width=\ablationQualFrameWidth]{assets/qualitative-ablation/prompt_000319/wo_mmd/frame_040_2p5s.jpg}}
\mfDeclareImage{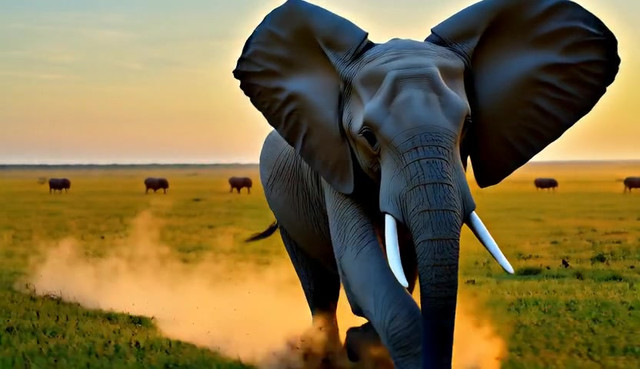}{\includegraphics[width=\ablationQualFrameWidth]{assets/qualitative-ablation/prompt_000319/wo_mmd/frame_060_3p75s.jpg}}
\mfDeclareImage{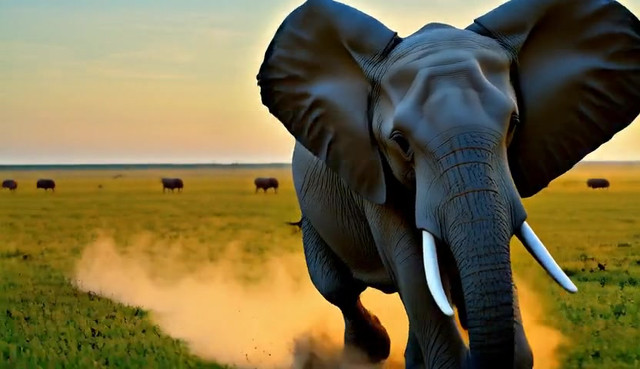}{\includegraphics[width=\ablationQualFrameWidth]{assets/qualitative-ablation/prompt_000319/wo_mmd/frame_080_5s.jpg}}

\begin{document}
\twocolumn[{%
    \renewcommand\twocolumn[1][]{#1}%
    \maketitle
\begin{center}
    \captionsetup{type=figure}
    \vspace*{-16pt}
    \begingroup
    \newcommand{\teaserShowTags}{1}
    \definecolor{teaserBase}{HTML}{4F4F4F}
    \definecolor{teaserOurs}{HTML}{BB4857}
    \definecolor{teaserInk}{HTML}{222222}
    \definecolor{teaserMuted}{HTML}{6B6B6B}
    \pgfmathsetlengthmacro{\teaserTabW}{11bp}   %
    \pgfmathsetlengthmacro{\teaserGap}{1.5bp}   %
    \pgfmathsetlengthmacro{\teaserPanelW}{196bp}%
    \pgfmathsetlengthmacro{\teaserPanelGap}{10bp}
    \pgfmathsetlengthmacro{\teaserBand}{11bp}   %
    \pgfmathsetlengthmacro{\teaserFrameW}{(\linewidth-\teaserPanelW-\teaserPanelGap-\teaserTabW-3*\teaserGap)/3}
    \pgfmathsetlengthmacro{\teaserFrameH}{\teaserFrameW*(480/832)}%
    \pgfmathsetlengthmacro{\teaserFrameX}{\teaserTabW+\teaserGap}
    \pgfmathsetlengthmacro{\teaserFramesW}{3*\teaserFrameW+2*\teaserGap}
    \pgfmathsetlengthmacro{\teaserRowsH}{3*\teaserFrameH+2*\teaserGap}
    \pgfmathsetlengthmacro{\teaserPanelX}{\linewidth-\teaserPanelW}
    \newcommand{\teaserRow}[5]{%
        \pgfmathsetlengthmacro{\teaserY}{-#1*(\teaserFrameH+\teaserGap)}
        \fill[#3, rounded corners=1.5bp] (0pt,\teaserY) rectangle (\teaserTabW,\teaserY-\teaserFrameH);
        \node[rotate=90, text=white, font=\scriptsize\bfseries]
            at (0.5*\teaserTabW,\teaserY-0.5*\teaserFrameH) {#4\vphantom{g}};
        \foreach \teaserTime [count=\teaserCol from 0] in {0,2.5,5} {
            \pgfmathsetlengthmacro{\teaserX}{\teaserFrameX+\teaserCol*(\teaserFrameW+\teaserGap)}
            \begin{scope}
                \clip[rounded corners=1.5bp] (\teaserX,\teaserY) rectangle ++(\teaserFrameW,-\teaserFrameH);
                \node[anchor=north west] at (\teaserX,\teaserY)
                    {\mfImage{assets/teaser/#2-\teaserTime.jpg}};
            \end{scope}
        }
        \ifnum\teaserShowTags=1
            \node[anchor=south east, fill=#3, text=white, rounded corners=1.5bp,
                  inner xsep=2.6bp, inner ysep=1.7bp, font=\fontsize{6}{7}\selectfont\bfseries]
                at (\teaserFrameX+\teaserFramesW-2.5bp,\teaserY-\teaserFrameH+2.5bp) {#5\vphantom{lg}};
        \fi
    }
    \begin{tikzpicture}[x=1bp, y=1bp, inner sep=0pt, outer sep=0pt]
        \useasboundingbox (0pt,12.5bp) rectangle (\linewidth,-\teaserRowsH-\teaserBand);
        \node[anchor=base west, font=\footnotesize, text=teaserInk] at (0pt,4.5bp)
            {\textbf{(a)}\enspace Prompt: \textit{``\ldots a father blowing up a balloon for his excited child\ldots''}};
        \node[anchor=base west, font=\footnotesize, text=teaserInk] at (\teaserPanelX,4.5bp)
            {\textbf{(b)}\enspace VBench Dynamic Degree during DMD training};
        \teaserRow{0}{self-forcing}{teaserBase}{Self Forcing}{nearly static\,\ding{55}}
        \teaserRow{1}{causal-forcing}{teaserBase}{Causal Forcing}{no inflation\,\ding{55}}
        \teaserRow{2}{ours}{teaserOurs}{Mass Forcing}{full inflation\,\ding{51}}
        \pgfmathsetlengthmacro{\teaserAxisY}{-\teaserRowsH-6.2bp}
        \draw[teaserMuted, line width=0.5bp, -{Stealth[length=3.4bp, width=3bp]}]
            (\teaserFrameX,\teaserAxisY) -- (\teaserFrameX+\teaserFramesW,\teaserAxisY);
        \foreach \teaserTime [count=\teaserCol from 0] in {0\,s, 2.5\,s, 5\,s} {
            \pgfmathsetlengthmacro{\teaserX}{\teaserFrameX+\teaserCol*(\teaserFrameW+\teaserGap)+0.5*\teaserFrameW}
            \node[fill=white, inner xsep=3bp, inner ysep=1bp, font=\scriptsize, text=teaserInk]
                at (\teaserX,\teaserAxisY) {\teaserTime};
        }
        \node[anchor=north west] at (\teaserPanelX,0pt)
            {\includegraphics[width=\teaserPanelW]{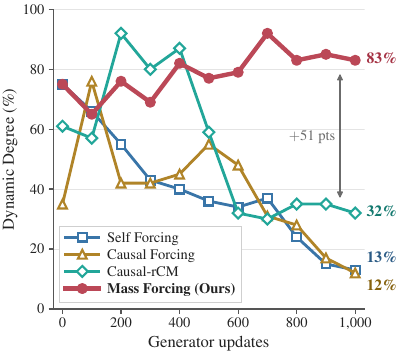}};
    \end{tikzpicture}
    \endgroup
    \vspace{-5.5mm}%
    \caption{\textbf{Mass Forcing sustains temporal dynamics during DMD training.}
    \textbf{(a)} Videos from the same prompt. Self Forcing~\citep{huang_2026_nips_self_forcing} stays nearly static and Causal Forcing~\citep{zhu_2026_icml_causal_forcing} never inflates the balloon, while Mass Forcing shows the full inflation.
    \textbf{(b)} VBench Dynamic Degree, the fraction of videos classified as dynamic, over DMD training. Self Forcing, Causal Forcing, and Causal-rCM~\citep{zheng_2026_causalrcm} start from different initializations, yet all fall to $12$-$32\%$ by update 1{,}000, suggesting that the mode-seeking objective, not the initialization, drives the loss of dynamics. Mass Forcing stays between $65\%$ and $92\%$ and ends at $83\%$, 51 points above the best baseline.
    }
    \label{fig:teaser}
\end{center}
 }]
%
% First-page footnote (the title block is set inside \twocolumn[...], where \thanks does not work).
{\let\thefootnote\relax\footnotetext{\textsuperscript{\dag}Work done during an internship at AMAP, Alibaba Group.}}
\begin{abstract}
Distribution Matching Distillation (DMD) has emerged as a promising paradigm for real-time video synthesis, distilling pretrained diffusion models into few-step autoregressive generators.
However, its reverse-KL objective is mode-seeking and can collapse the student's conditional video distribution onto a narrow subset of modes, leading to reduced sample diversity, diminished temporal dynamics, and color oversaturation.
Although existing methods primarily mitigate this issue via coverage-oriented student initialization, we observe that such initial coverage is transient: across three different initializations, the fraction of dynamic videos drops sharply during subsequent DMD training, indicating that the unchanged mode-seeking objective, rather than the initialization, drives this degradation.
Motivated by this training-time degradation, we propose Mass Forcing, a probability-mass-aware framework that moves on-policy DMD supervision from whole videos to individual states.
Each update branches several continuations from a shared student-generated prefix, making the student's probability mass at that state measurable against the teacher.
First, at the local level, Mass Forcing estimates relative teacher-to-student density ratios among these continuations through latent-space score integration and uses them to rebalance conditional DMD updates, amplifying the training influence of rollouts underrepresented relative to the teacher.
Second, recognizing that mismatched mode probabilities do not necessarily induce strong local score corrections, Mass Forcing aligns generated rollout sets with a reweighted reference distribution defined by the same mass estimates, providing cross-rollout supervision beyond individual gradient reweighting.
Experiments on MovieGen and VBench demonstrate that Mass Forcing improves video fidelity and sustains temporal dynamics throughout training, achieving the highest Total scores among compared methods and a Dynamic Degree of 83\% after 1,000 generator updates, versus 12-32\% for the baselines, and introducing no additional inference-time computational overhead.
\end{abstract}
\section{Introduction}
\label{sec:introduction}

Recent video diffusion models synthesize high-fidelity videos with fine visual detail and complex motion~\citep{brooks_2024_sora,yang_2025_iclr_cogvideox,polyak_2024_moviegen,kong_2024_hunyuanvideo,ma_2025_stepvideo,wan_2025_wan,yang_2026_eccv_omnidance,zhu_2026_dreamx_creator}.
Most of these models, however, are bidirectional: they jointly denoise all frames at every step. Hence, no frames can be shown until the whole clip is generated, and this latency precludes real-time streaming and interaction.
Autoregressive (AR) video models~\citep{yan_2021_videogpt,kondratyuk_2024_icml_videopoet,wang_2024_loong,deng_2025_iclr_nova,jin_2025_iclr_pyramidflow,sandai_2025_magi1} instead generate frames or chunks sequentially, each conditioned on the preceding context, which reduces time-to-first-frame and enables continuous, interactive generation.
Yet AR diffusion models still require many denoising steps per chunk, keeping them far from real time.

Few-step AR distillation closes this gap by matching a causal student, initialized from a pretrained video diffusion model, to the teacher's output distribution via Distribution Matching Distillation (DMD)~\citep{yin_2024_cvpr_dmd,yin_2024_nips_dmd2,yin_2025_cvpr_causvid}.
Self Forcing~\citep{huang_2026_nips_self_forcing} performs this matching \emph{on-policy}: it rolls out the student's own video and applies DMD to the video as a whole.
However, DMD minimizes a \textit{mode-seeking} reverse Kullback--Leibler (KL) divergence, which lets the student concentrate on a few teacher-supported modes, reducing sample diversity and temporal dynamics~\citep{xu_2025_fdistill,you_2026_cvpr_adaptive}.
Existing remedies for AR video mainly improve initialization, either with an AR teacher that bridges the bidirectional-to-causal gap, as in Causal Forcing~\citep{zhu_2026_icml_causal_forcing}, or with a mode-covering start from consistency distillation, as in Causal-rCM~\citep{zheng_2026_causalrcm}, leaving the subsequent mode-seeking objective unchanged.

Our key observation is that the mode coverage gained at initialization is transient under continued on-policy DMD.
\Cref{fig:teaser}b tracks VBench Dynamic Degree (DD)~\citep{huang_2024_cvpr_vbench}, the fraction of generated videos classified as dynamic, during DMD training from three different initializations.
Between $100$ and $1{,}000$ generator updates, DD falls from $66\%$ to $13\%$ for Self Forcing and from $76\%$ to $12\%$ for Causal Forcing, while Causal-rCM, despite its mode-covering initialization, drops from a peak of $92\%$ to $32\%$.
Different initializations thus affect the early trajectory but not the late-training decline, suggesting that the objective (rather than the initialization) drives the collapse.
This behavior is consistent with the well-known \textit{zero-forcing} property of reverse KL~\citep{bishop_2007_PMLR}: because the divergence is evaluated only on student samples, a teacher-supported mode that the student stops generating incurs almost no penalty.
We conjecture that motion-rich continuations are especially exposed in AR video generation: they spread probability across many diverse modes, whereas static continuations concentrate it, so on-policy training can drift toward static content at little cost.
Meanwhile, Self Forcing's Aesthetic and Imaging Quality improve overall as training proceeds (\cref{sec:appendix_quality_trajectory}), revealing an early-stopping trade-off between visual quality and temporal dynamics.
Resolving this dilemma requires supervision that preserves coverage throughout optimization, so that continued distillation improves quality without sacrificing dynamics.

Such supervision must look beyond individual samples: the DMD update checks whether each rollout is plausible under the teacher, but not whether the student generates it as often as the teacher would, so rollouts from underrepresented modes count no more than others and lost coverage is not actively restored.
Video-level on-policy training also leaves this frequency unobservable: each student-generated prefix is followed by a single continuation, so how the student distributes its probability mass at that state cannot be measured.
We therefore introduce \emph{Mass Forcing}, a probability-mass-aware framework that keeps the on-policy states of Self Forcing but supervises them one at a time.
Each update samples a chunk, generates its prefix with the current student, and branches several continuations from that prefix.
These continuations are draws from one conditional distribution and share one prefix, so their relative teacher-to-student density ratios can be estimated, which we do by integrating the score difference between the teacher and fake-score networks along latent-space paths.
Weighting DMD by such ratios yields more mode-covering objectives~\citep{xu_2025_fdistill}.
Unlike prior work, which obtains the ratios from an auxiliary GAN discriminator in image distillation, our estimator reuses the networks already present in DMD.

The same estimates drive two complementary objectives.
Locally, they reweight the conditional distribution-matching term of DMD toward underrepresented rollouts, while the classifier-free guidance (CFG)~\citep{ho_2022_cfg} term stays uniformly weighted, so that guidance strength is unchanged.
Such reweighting, however, can only amplify an existing score correction, which nearly vanishes when well-separated modes differ mainly in probability.
At the set level, we therefore use maximum mean discrepancy (MMD)~\citep{gretton_2012_jmlr_mmd,li_2015_icml_gmmn} to align the rollout set with a detached reference distribution reweighted by the same estimates, pulling rollouts toward those with higher estimated teacher support.
Both objectives reuse the same on-policy rollouts and need no teacher-generated ones.

\begin{figure*}[t]
    \centering
    \resizebox{0.9\linewidth}{!}{\includegraphics[pagebox=cropbox]{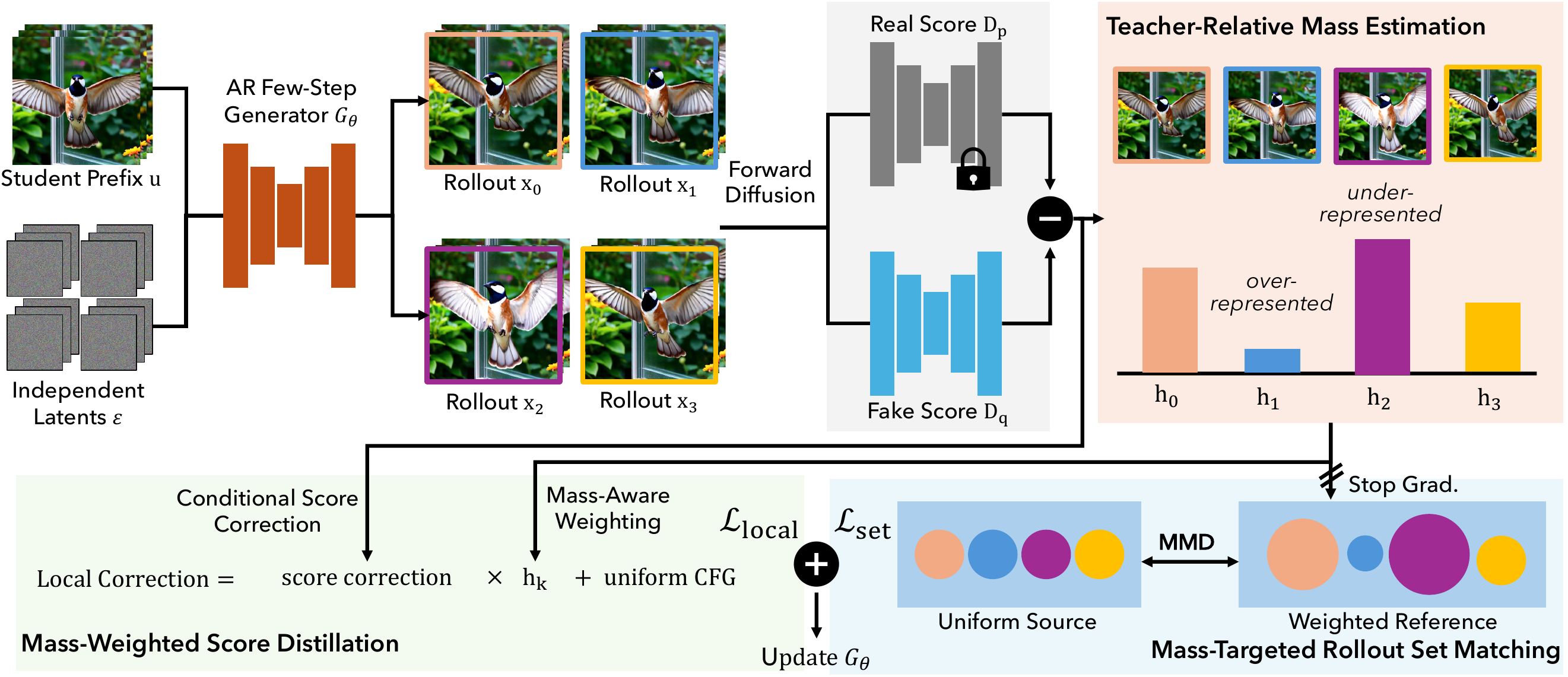}}
    \caption{\textbf{Mass Forcing training.}
    Each update samples one chunk, generates its prefix with the current student, and draws $K$ independent rollouts from that shared prefix.
    Integrating the difference between the frozen teacher score and the fake score along paths between rollouts gives each rollout a teacher-relative mass $h_k$, with $h_k>1$ for rollouts the student underrepresents.
    $\mathcal L_{\mathrm{local}}$ scales each rollout's conditional score correction by $h_k$ and leaves classifier-free guidance uniform, and $\mathcal L_{\mathrm{set}}$ matches the rollout set to an $h_k$-weighted copy of itself.}
    \label{fig:mass_forcing_method}
\end{figure*}

Experiments on MovieGen~\citep{polyak_2024_moviegen} and VBench~\citep{huang_2024_cvpr_vbench} show that Mass Forcing achieves the highest Total and Quality scores among all compared methods, raising the Total score over the strongest baseline, \ie, Causal Forcing, from $82.51$ to $83.38$ on MovieGen and from $84.14$ to $84.98$ on VBench.
Starting from the same ODE initialization as Self Forcing, Mass Forcing keeps DD between $65\%$ and $92\%$ throughout training and ends at $83\%$ (\cref{fig:teaser}b).
All additional supervision is confined to training: the standard four-step AR generator and its IID Gaussian sampling are unchanged, and inference incurs no extra latency.
Our contributions are:
\begin{itemize}[leftmargin=*,itemsep=.5pt,topsep=0pt,parsep=0pt]
    \item We show that the mode coverage provided by improved initialization is transient under on-policy DMD: across three initializations, temporal dynamics decline sharply as training continues, indicating that the objective, not the starting point, drives the collapse.
    \item We propose Mass Forcing, which supervises on-policy DMD per state by branching several continuations from one student-generated prefix, estimates their teacher-relative probability mass via score line integrals over existing DMD networks, and uses these estimates in two complementary objectives: mass-weighted conditional score distillation and mass-targeted rollout-set matching.
    \item Experiments on MovieGen and VBench show improved overall generation quality and sustained temporal dynamics over competitive few-step AR baselines, with no change to the inference pipeline or its cost.
\end{itemize}
\section{Method}
\label{sec:method}

Mass Forcing distills a bidirectional teacher into a few-step autoregressive (AR) student in three parts, each answering one question (\cref{fig:mass_forcing_method}).
\emph{Where to learn} (\cref{sec:method_onpolicy}): training is on-policy and per state, so every update starts from a student-generated prefix and supervises several continuations of the next chunk.
\emph{What is missing there} (\cref{sec:method_ratio}): DMD counts a continuation as often as the student generates it, not as often as the teacher would, and we measure this gap by a teacher-relative mass estimated from DMD's own score networks.
\emph{How to use it} (\cref{sec:method_objectives}): the mass reweights each rollout's score correction and sets a target for the rollout set.

\subsection{On-Policy Distribution Matching}
\label{sec:method_onpolicy}

For a prompt $c$, the student generates a latent video $V$ as $B$ chunks, each conditioned on the chunks before it:
\begin{equation}
    Q_\theta(V\mid c)
    =\prod_{b=1}^{B}q_\theta\bigl(x^{(b)}\mid u^{(b)},c\bigr),
    \label{eq:mf_ar_factorization}
\end{equation}
where $Q_\theta$ is the student's video distribution with parameters $\theta$, $q_\theta$ is its conditional distribution for one chunk, $x^{(b)}\in\mathbb R^d$ is the $b$-th chunk, of dimension $d$, and $u^{(b)}$ is the prefix formed by the first $b-1$ chunks.

\paragraph{Which states to learn on.}
On-policy training supervises a model on inputs it produces itself~\citep{yang_2026_unopd}.
DMD2~\citep{yin_2024_nips_dmd2} does so along denoising time, and Self Forcing~\citep{huang_2026_nips_self_forcing} also along video time, by rolling out a whole video from the student's own history and scoring it as a whole.
Its supervision is thus at the level of the trajectory, with a single continuation per state.
We keep the same on-policy states and supervise per state instead.
Each update samples one chunk index $b$ uniformly, generates the prefix $u=u^{(b)}$ with the current student, detaches it together with its KV cache, and branches $K$ independent continuations of chunk $b$ from this shared prefix.
The score networks see only the prefix and the current chunk, so distribution matching targets the conditional distribution of the next chunk given the state $(u,c)$.
By itself, this change does not improve over Self Forcing (\cref{sec:exp_ablation}).
Its role is to make mass estimable: the $K$ continuations are draws from one conditional distribution and share one prefix, so the student's mass at a state becomes observable and their density ratios directly comparable (\cref{sec:method_ratio}).
\Cref{sec:appendix_long_video} evaluates 60-second rollouts, twelve times the training length.

\paragraph{What is matched at each state.}
Fix a state $(u,c)$ and a noise level $t$, and let $y$ be a noised continuation,
\begin{equation}
    y=\alpha_t x+\sigma_t\eta,
    \qquad x=G_\theta(\epsilon;u,c),
    \label{eq:mf_noised_continuation}
\end{equation}
where $x$ is a continuation produced by the generator $G_\theta$ from sampler noise $\epsilon$, $\eta$ is standard Gaussian noise, $\sigma_t$ is the noise scale at level $t$, and $\alpha_t=1-\sigma_t$.
Let $q_{\theta,t}$ and $p_t$ be the densities of $y$ under the student and the teacher, both conditioned on $(u,c)$, and let $s_{q,t}$ and $s_{p,t}$ be their scores, that is, the gradients of their logarithms with respect to $y$.
DMD~\citep{yin_2024_cvpr_dmd,yin_2024_nips_dmd2} minimizes the reverse KL divergence from $q_{\theta,t}$ to $p_t$, averaged over noise levels with weights $\omega(t)$.
We denote this objective by $\mathcal L_{\mathrm{RKL}}$, and its gradient is
\begin{equation}
    \nabla_\theta\mathcal L_{\mathrm{RKL}}
    =\mathbb E_{t,\epsilon,\eta}\!\left[
      \omega(t)\alpha_tJ_\theta^\top
      (s_{q,t}(y)-s_{p,t}(y))\right],
    \label{eq:mf_rkl_gradient}
\end{equation}
where $J_\theta$ is the Jacobian of $G_\theta$ with respect to $\theta$.
The prefix is held fixed within an update, and \cref{sec:app_scope} relates this fixed-prefix objective to full-video DMD.

\subsection{Teacher-Relative Mass}
\label{sec:method_ratio}

\begin{figure}[t]
    \centering
    \includegraphics[width=\linewidth]{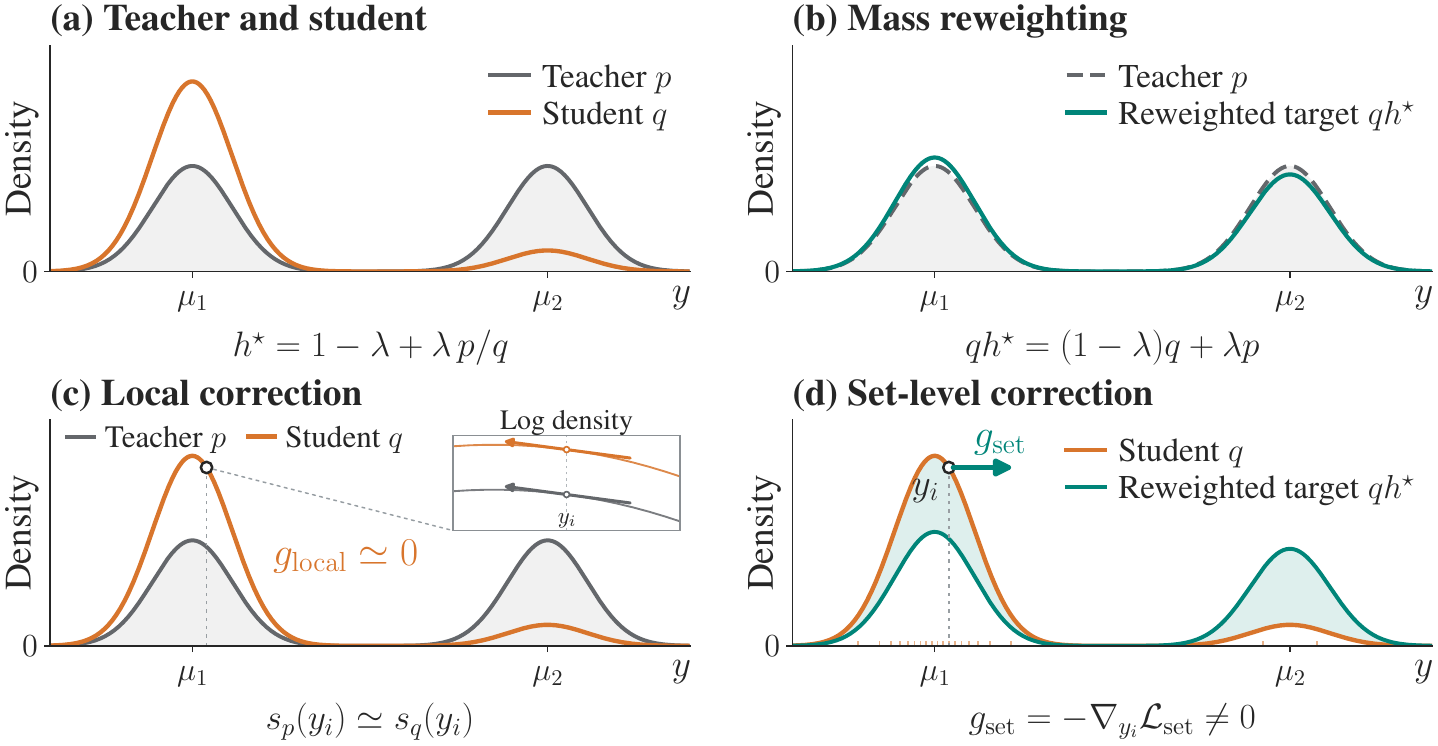}
    \caption{\textbf{Mass-aware local and set-level supervision.}
    (a,b) Weighting student samples by $h^\star$ moves the effective measure from $q$ to $qh^\star$, restoring mass on the underrepresented mode $\mu_2$.
    (c,d) Within a mode, the local correction is weak, while set-level matching still moves rollouts toward $\mu_2$.}
    \label{fig:mass_forcing_core_idea}
\end{figure}
 
On-policy training decides where \cref{eq:mf_rkl_gradient} is applied, but not what it measures.
Its expectation runs over student samples, so at any state a mode enters the gradient in proportion to the student's mass on it, not the teacher's: a teacher-supported continuation that the student rarely samples contributes few corrections, and one it no longer samples contributes none.

\paragraph{Ideal weight.}
At a fixed state and noise level, write $q$, $p$, $s_q$, and $s_p$ for the densities and scores above.
What is missing is the ratio $r$ of the teacher density to the student density.
Where $r>1$, the teacher assigns more probability than the student, so the student underrepresents that region; where $r<1$, the student overrepresents it.
Suppose $r$ is known, and weight each student sample's correction by $h^*$:
\begin{equation}
    \begin{aligned}
    h^*(y)&=1-\lambda+\lambda\,r(y),
    \qquad r(y)=\frac{p(y)}{q(y)},\\
    q(y)\,h^*(y)&=(1-\lambda)\,q(y)+\lambda\,p(y),
    \end{aligned}
    \label{eq:mf_ideal_mass}
\end{equation}
with a mixing coefficient $\lambda\in[0,1]$.
By the second line, an $h^*$-weighted average over student samples is an average over a mixture of student and teacher samples:
\begin{equation}
    \begin{aligned}
    &\mathbb E_{y\sim q}\bigl[h^*(y)\,(s_q-s_p)\bigr]\\
    &\quad=\underbrace{(1-\lambda)\,\mathbb E_{y\sim q}[s_q-s_p]}_{\text{on student samples}}
      +\underbrace{\lambda\,\mathbb E_{y\sim p}[s_q-s_p]}_{\text{on teacher samples}}.
    \end{aligned}
    \label{eq:mf_coverage}
\end{equation}
The first term is the DMD correction.
The second evaluates it where the teacher places its mass, so a teacher-supported mode keeps its share of the gradient however rarely the student visits it.
With the Jacobian of \cref{eq:mf_rkl_gradient} included, this is exactly the gradient of a mixture of reverse and forward KL with weights $1-\lambda$ and $\lambda$~\citep{xu_2025_fdistill} (\cref{sec:app_kl}).
Per sample, the weight is above one where $r>1$ and below one where $r<1$, so underrepresented rollouts gain influence and overrepresented ones lose it.
The mean weight stays one, and the inference sampler is unchanged (\cref{fig:mass_forcing_core_idea}(a,b)).

\paragraph{Estimation from DMD's own scores.}
The ratio $r$ is unknown, and prior work estimates it with an auxiliary GAN discriminator~\citep{xu_2025_fdistill}.
We instead read it off what DMD already computes:
\begin{equation}
    s_p(y)-s_q(y)
    =\nabla_y\log\frac{p(y)}{q(y)}
    =\nabla_y\log r(y).
    \label{eq:mf_score_ratio}
\end{equation}
\emph{The score difference that drives the DMD gradient is the gradient of the log ratio.}
Integrating it between two samples gives their log-ratio difference, and normalizing within a group removes the unknown constant, in four steps (\cref{fig:mass_forcing_method}).

\textbf{(i) Sample.}
We detach the $K$ rollouts of \cref{sec:method_onpolicy} and add one shared noise draw, giving the noisy rollouts $\bar y_0,\ldots,\bar y_{K-1}$ (overbars mark detached samples).
Sharing the noise keeps its level constant along paths between them.

\textbf{(ii) Score difference.}
The frozen teacher and a fake denoiser, trained on current student rollouts, predict the clean chunk from a noisy clip as $D_p$ and $D_q$, respectively, and their difference gives an estimate $\Delta$ of the score difference:
\begin{equation}
    \Delta(y)=\frac{\alpha_t}{\sigma_t^2}\bigl(D_p(y)-D_q(y)\bigr).
    \label{eq:mf_score_field}
\end{equation}
Both networks receive the prefix at the same noise level as the chunk, so $\Delta$ estimates the gradient of $\log\tilde r$, where $\tilde r$ is the ratio conditioned on the noisy prefix, a proxy for the clean-prefix ratio $r$ (\cref{sec:app_scores}).

\textbf{(iii) Integrate.}
With the medoid of the group as anchor $a$, we integrate $\Delta$ along the straight path from $a$ to each rollout $k$ by three-point Simpson quadrature in latent space (\cref{eq:mf_simpson}), which gives its relative log ratio $\ell_k$:
\begin{equation}
    \ell_k\approx\int_a^{\bar y_k}\!\Delta(y)^\top\mathrm dy
    \approx\log\tilde r(\bar y_k)-\log\tilde r(a).
    \label{eq:mf_line_integral}
\end{equation}
The second relation is an equality for the exact score difference.

\textbf{(iv) Normalize.}
The anchor term is shared by the group and cancels under normalization:
\begin{equation}
    \widehat r_k=K\frac{\exp(v_k)}{\sum_j\exp(v_j)},
    \qquad h_k=1-\lambda_{\mathrm{eff}}+\lambda_{\mathrm{eff}}\,\widehat r_k.
    \label{eq:mf_weights}
\end{equation}
Here $\widehat r_k$ is the normalized ratio estimate of rollout $k$, $h_k$ is its weight, and the sum runs over the group.
The logit $v_k$ is $\ell_k$ centered at the group mean, divided by a temperature $\tau$, and clipped to $[-\kappa,\kappa]$ with a bound $\kappa$ for stability (\cref{eq:mf_ratio_logits}).
Without tempering and clipping, and for exact log ratios, $\widehat r_k$ is the ratio of rollout $k$ divided by the group average.
Thus $h_k$ is $h^*$ with the estimate $\widehat r_k$ in place of $r$ and, in place of $\lambda$, a group-shared coefficient $\lambda_{\mathrm{eff}}$ that is bounded by a maximum $\lambda_{\max}$ and set by the training ramp and reliability gates (\cref{sec:impl_mass}).
The weights are detached and average to one over the group.
Computed from a finite group, a learned field, and a noisy prefix, they are a coverage-oriented surrogate, not exact forward-KL optimization (\cref{sec:app_ratio}).

\subsection{Mass-Aware Objectives}
\label{sec:method_objectives}

The weights enter training twice: locally, they rescale each rollout's score correction as in \cref{eq:mf_coverage}, and at the set level, they prescribe the mass each rollout should carry.

\paragraph{Mass-weighted score distillation.}
\label{sec:method_local}
Since the mass estimates target the conditional teacher, we separate conditional distribution matching from classifier-free guidance (CFG)~\citep{liu_2025_decoupled_dmd} and weight only the former.
In clean-latent coordinates, the correction applied to rollout $k$ is
\begin{equation}
    g_k=\frac1{n_k}\left[
       h_k\underbrace{(D_{q,k}-D_{p,k})}_{\text{conditional matching}}
       +\underbrace{(D_{p,k}-D_{\mathrm{cfg},k})}_{\text{guidance}}
    \right],
    \label{eq:mf_local_gradient}
\end{equation}
where subscript $k$ denotes evaluation at $\bar y_k$, $D_{\mathrm{cfg}}$ is the CFG-guided teacher prediction, and $n_k$ is a detached DMD normalizer (\cref{sec:app_local}).
With unit weights, this is the standard guided DMD correction; otherwise, only conditional matching is rescaled.
As $g_k$ is computed on detached rollouts, we regenerate each one from the same randomness and prefix cache to obtain a differentiable copy $x_k$ of the clean rollout, equal in value, and minimize the local loss
\begin{equation}
    \mathcal L_{\mathrm{local}}
    =\frac{1}{2Kd}\sum_{k=0}^{K-1}
      \left\|x_k-\operatorname{sg}(x_k-g_k)\right\|_2^2,
    \label{eq:mf_local_loss}
\end{equation}
where $\operatorname{sg}$ stops gradients, so the gradient in $x_k$ is $g_k/(Kd)$.

\paragraph{Mass-targeted rollout-set matching.}
\label{sec:method_set}
A weight can only rescale a correction that is already there.
When teacher and student share well-separated component shapes but assign them different probabilities, their within-mode scores nearly coincide (\cref{fig:mass_forcing_core_idea}(c)), so the weight multiplies an almost vanishing correction.
The weights themselves remain informative, because the line integral accumulates score differences along paths \emph{between} modes.
We therefore use them a second time, as a target for the rollout set.
The uniformly weighted source $\mu_\theta$ and the reweighted reference $\nu$ are
\begin{equation}
    \mu_\theta=\frac1K\sum_k\delta_{y_k},
    \qquad \nu=\sum_k\frac{h_k}{K}\,\delta_{\bar y_k},
    \label{eq:mf_empirical_target}
\end{equation}
where $\delta_y$ is a point mass, $y_k$ is the differentiable copy of $\bar y_k$, and the reference is detached.
We match them with maximum mean discrepancy (MMD)~\citep{gretton_2012_jmlr_mmd,li_2015_icml_gmmn} and add this set-level loss $\mathcal L_{\mathrm{set}}$ to the local loss for the total loss $\mathcal L_{\mathrm{MF}}$:
\begin{equation}
    \mathcal L_{\mathrm{set}}
    =\operatorname{MMD}_{\mathcal K_\beta}^2(\mu_\theta,\nu),
    \quad
    \mathcal L_{\mathrm{MF}}
    =\mathcal L_{\mathrm{local}}+\gamma\mathcal L_{\mathrm{set}},
    \label{eq:mf_total_loss}
\end{equation}
where $\mathcal K_\beta$ is a Gaussian kernel with detached scale $\beta$ and $\gamma$ weights the set-level term.
At the start of an update, $y_k=\bar y_k$ for every $k$, so the two measures coincide in location and differ only in mass, and the gradient on rollout $i$ is
\begin{equation}
    \left.\nabla_{y_i}\mathcal L_{\mathrm{set}}\right|_{y=\bar y}
    =\frac{2}{K^2}\sum_j(1-h_j)
      \nabla_1\mathcal K_\beta(\bar y_i,\bar y_j),
    \label{eq:mf_set_replay_gradient}
\end{equation}
where $\nabla_1$ differentiates the first kernel argument and $j$ indexes the references.
Under gradient descent, each term attracts the rollout toward a reference with $h_j>1$ or repels it from one with $h_j<1$.
The sum can be nonzero even where the local correction vanishes (\cref{fig:mass_forcing_core_idea}(d); \cref{sec:app_set}).

\paragraph{Training and inference.}
Generator updates hold prefixes, score predictions, weights, and references fixed, while the fake denoiser is trained separately on unweighted student clips (\cref{sec:app_training}; \cref{alg:mass_forcing_update}).
With unit weights and without the set-level term, this procedure reduces to unweighted conditional-chunk DMD, not the holistic Self Forcing data path.
Both objectives reuse the same on-policy rollouts without teacher-generated continuations, and all added computation is training-only, preserving standard four-step AR inference with IID Gaussian noise.

\section{Experiments}
\label{sec:exp}

\begin{table*}[t]
\caption{Quantitative comparison with existing methods. Mass Forcing obtains the best Total and Quality on both benchmarks while retaining real-time inference speed. Best results are in \textbf{bold}, second best \underline{underlined}.}
\label{tab:main-table}
\centering
\small
\hypersetup{hidelinks}
\setlength{\tabcolsep}{4.45pt}
\setlength{\heavyrulewidth}{0.75pt}
\setlength{\lightrulewidth}{0.35pt}
\setlength{\cmidrulewidth}{0.4pt}
\renewcommand{\arraystretch}{1.05}
\newlength{\mainTableHeaderShift}
\setlength{\mainTableHeaderShift}{\dimexpr\aboverulesep+\belowrulesep+\cmidrulewidth\relax}
\setlength{\mainTableHeaderShift}{-0.5\mainTableHeaderShift}
\newcommand{\mainTableCite}[1]{{\footnotesize\color{black!60}\citep{#1}}}
\newcommand{\mainTableMissing}{\textcolor{black!45}{--}}
\resizebox{0.9\linewidth}{!}{%
\begin{tabular}{@{}l*{10}{c}@{}}
\toprule
\multirow{2}{*}[\mainTableHeaderShift]{\textbf{Model}} &
\multirow{2}{*}[\mainTableHeaderShift]{\shortstack{\textbf{Throughput}\,$\uparrow$\\{\footnotesize\color{black!65}(FPS)}}} &
\multirow{2}{*}[\mainTableHeaderShift]{\shortstack{\textbf{Latency}\,$\downarrow$\\{\footnotesize\color{black!65}(s)}}} &
\multicolumn{4}{c}{\textbf{MovieGen Evaluation}\,$\uparrow$} &
\multicolumn{4}{c@{}}{\textbf{VBench Evaluation}\,$\uparrow$} \\
\cmidrule(l{\tabcolsep}r{\tabcolsep}){4-7}\cmidrule(l{\tabcolsep}r{0pt}){8-11}
& & & \textbf{Total} & \textbf{Quality} & \textbf{Semantic} & \textbf{Dynamic}
    & \textbf{Total} & \textbf{Quality} & \textbf{Semantic} & \textbf{Dynamic} \\
\midrule
\rowcolor{black!4}
\multicolumn{11}{l}{\color{black!75}\textit{Bidirectional Video Diffusion Models}} \\
\addlinespace[1pt]
LTX-Video~\mainTableCite{hacohen_2024_ltxvideo} & 8.98 & 13.5 & 80.40 & 82.44 & 72.21 & 31.90 & 81.71 & 83.78 & 73.42 & 43.06 \\
Wan2.1-1.3B~\mainTableCite{wan_2025_wan} & 0.24 & 330 & 81.47 & 82.73 & 76.41 & 59.02 & 83.95 & 85.17 & 79.04 & 62.50 \\
AVD~\mainTableCite{you_2026_cvpr_adaptive} & 4.61 & 17.55 & 80.17 & 81.54 & 74.69 & 61.11 & 82.07 & 83.51 & 76.32 & 73.61 \\
\addlinespace[1pt]
\rowcolor{black!4}
\multicolumn{11}{l}{\color{black!75}\textit{Autoregressive Video Diffusion Models}} \\
\addlinespace[1pt]
NOVA~\mainTableCite{deng_2025_iclr_nova} & 0.88 & 4.1 & 77.32 & 78.36 & 73.14 & 50.45 & 79.19 & 79.96 & 76.11 & 44.44 \\
Pyramid Flow~\mainTableCite{jin_2025_iclr_pyramidflow} & 6.70 & 2.5 & 80.11 & 81.07 & 76.24 & 34.40 & 82.05 & 83.51 & 76.21 & 59.72 \\
SkyReels-V2~\mainTableCite{chen_2025_skyreelsv2} & 0.49 & 112 & 80.67 & 82.51 & 73.33 & 50.25 & 82.73 & 84.86 & 74.19 & 63.89 \\
MAGI-1~\mainTableCite{sandai_2025_magi1} & 0.19 & 282 & 81.04 & 81.81 & 77.95 & 22.13 & 80.42 & 81.38 & 76.59 & 16.67 \\
\addlinespace[1pt]
\rowcolor{black!4}
\multicolumn{11}{l}{\color{black!75}\textit{Distilled Autoregressive Video Models}} \\
\addlinespace[1pt]
CausVid~\mainTableCite{yin_2025_cvpr_causvid} & 17.0 & 0.69 & 80.50 & 81.72 & 75.62 & 57.33 & 82.56 & 83.92 & 77.11 & 65.28 \\
Self Forcing~\mainTableCite{huang_2026_nips_self_forcing} & 17.0 & 0.69 & 81.75 & 82.77 & 77.70 & 63.41 & 83.49 & 84.51 & 79.40 & 61.11 \\
Causal Forcing~\mainTableCite{zhu_2026_icml_causal_forcing} & 17.0 & 0.69 & \underline{82.51} & \underline{83.28} & \textbf{79.42} & \underline{77.57} & \underline{84.14} & \underline{85.22} & \underline{79.83} & \underline{86.11} \\
Causal-rCM~\mainTableCite{zheng_2026_causalrcm} & 17.0 & 0.69 & 81.68 & 82.80 & 77.21 & 48.26 & 83.85 & 85.10 & 78.84 & 69.44 \\
\midrule
\textbf{Mass Forcing (Ours)} & 17.0 & 0.69 & \textbf{83.38} & \textbf{84.53} & \underline{78.78} & \textbf{83.35} & \textbf{84.98} & \textbf{86.17} & \textbf{80.23} & \textbf{88.89} \\
\bottomrule
\end{tabular}%
}
\end{table*}

We evaluate whether coverage-oriented supervision improves video generation quality and sustains temporal dynamics during continued DMD training.
We first compare Mass Forcing with existing video generators, then examine training trajectories and the contributions of local mass reweighting and set-level matching.

\subsection{Experimental Setup}
\label{sec:exp_setup}

\paragraph{Model and training.}
We build on the Self Forcing implementation~\citep{huang_2026_nips_self_forcing}, using a causal Wan2.1-T2V-1.3B generator, a frozen bidirectional Wan2.1-T2V-14B teacher, and a bidirectional 1.3B residual network for fake-score estimation~\citep{wan_2025_wan}.
Mass Forcing is applied during DMD training from the ODE-initialized generator, rather than as post-training of a converged DMD checkpoint.
Following Self Forcing, we train on a filtered subset of VidProM~\citep{wang_2024_nips_vidprom} containing 248{,}221 text prompts.
The default configuration uses $K=4$ rollouts per prefix, temperature $\tau=2$, logit bound $\kappa=2$, and maximum mixture coefficient $\lambda_{\max}=0.25$.
The set-level reference uses the mixed weights $h_k/K$ in \cref{eq:mf_empirical_target}.
Optimizer settings, rollout sampling, and mass-activation rules are detailed in \cref{sec:implementation_details}.

\paragraph{Benchmarks and metrics.}
We evaluate on $1{,}003$ MovieGen~\citep{polyak_2024_moviegen} prompts and $946$ VBench~\citep{huang_2024_cvpr_vbench} prompts, with all videos generated at 480p and assessed using the standard VBench quality and semantic evaluators.
\Cref{tab:main-table} reports aggregate Quality, Semantic, and Total scores on the evaluator's normalized scale.
\Cref{sec:appendix_dimensions} reports the individual dimension scores before aggregate-score normalization, multiplied by 100 for readability.
MovieGen does not provide the metadata required for VBench semantic evaluation, so we construct it using Qwen3-VL-32B-Instruct.
We exclude Spatial Relationship because no prompts satisfying VBench's criteria for this dimension were identified in the evaluated MovieGen set.
All compared methods use the same metadata and eight-dimensional protocol, retaining VBench's per-dimension normalization and relative weights while renormalizing the retained weights to sum to one (\cref{sec:appendix_moviegen}).
We report Dynamic Degree (DD), the fraction of videos classified as dynamic, alongside aggregate quality, since motion alone is not sufficient evidence of generation quality.
We treat DD as a measure of temporal dynamics, not of mode coverage.

\paragraph{Comparisons and inference.}
The comparisons cover bidirectional generators, AR generators, and distilled AR generators, as grouped in \cref{tab:main-table}.
The closest baselines are CausVid, Self Forcing, Causal Forcing, and Causal-rCM, which allow us to compare with distribution matching, on-policy training, and improved initialization.
For each baseline, we use its officially released checkpoint and native inference pipeline.
Mass Forcing retains four-step AR generation with IID Gaussian noise.
Neither score estimation nor rollout-set matching is used at inference.
Throughput (FPS) and latency are measured on a single NVIDIA H100 GPU with batch size 1 and averaged after model loading, excluding VAE decoding.
For AR methods, the first chunk is excluded from timing, so the reported latency measures subsequent-chunk generation rather than time to first frame.

\begin{figure*}[t]
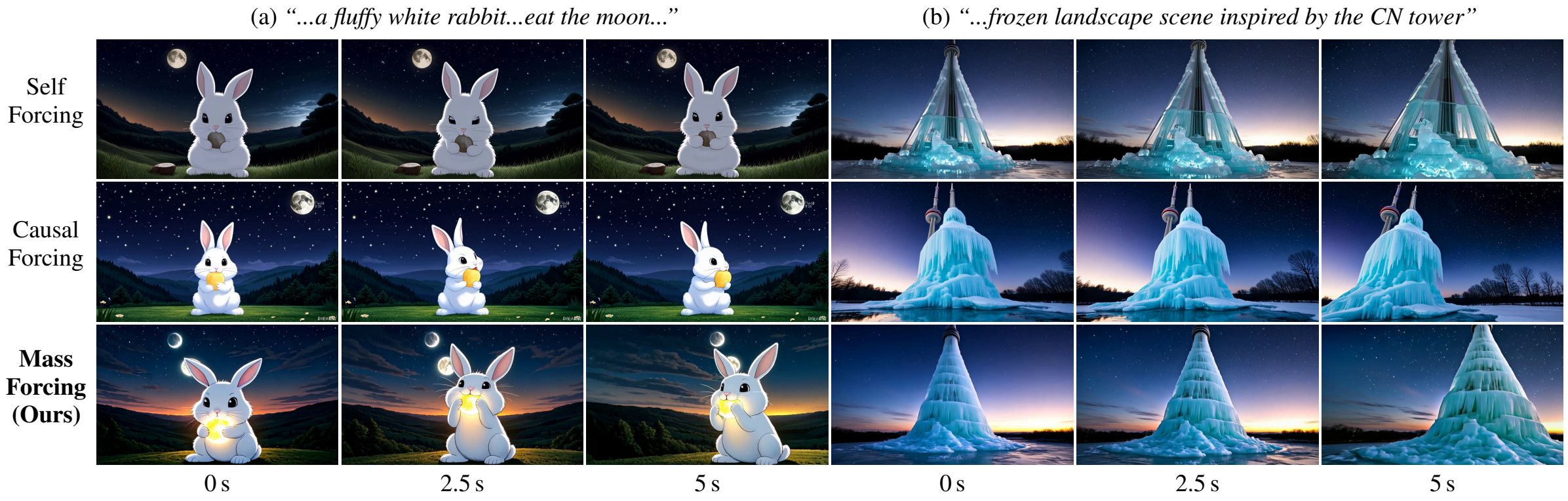

    \centering
    \begingroup
    \scriptsize
    \setlength{\tabcolsep}{0pt}
    \setlength{\extrarowheight}{0pt}
    \renewcommand{\arraystretch}{1}
    \newlength{\qualLabelWidth}
    \newlength{\qualFrameWidth}
    \newlength{\qualImageGap}
    \newlength{\qualSceneGap}
    \settowidth{\qualLabelWidth}{\textbf{Forcing}}
    \setlength{\qualImageGap}{1pt}
    \setlength{\qualSceneGap}{\qualImageGap}
    \setlength{\qualFrameWidth}{\dimexpr(0.9\linewidth-\qualLabelWidth-3pt-4\qualImageGap-\qualSceneGap)/6\relax}
    \newcommand{\qualFrame}[1]{%
        \raisebox{-0.5\height}{\mfImage{assets/qualitative/#1}}}
    \newcommand{\qualMethod}[1]{%
        \parbox[c]{\qualLabelWidth}{\centering#1}}
    \begin{tabular}{@{}c@{\hspace{3pt}}c@{\hspace{\qualImageGap}}c@{\hspace{\qualImageGap}}c@{\hspace{\qualSceneGap}}c@{\hspace{\qualImageGap}}c@{\hspace{\qualImageGap}}c@{}}
        & \multicolumn{3}{@{}c@{\hspace{\qualSceneGap}}}{%
            \parbox[b]{\dimexpr3\qualFrameWidth+2\qualImageGap\relax}{\centering
            (a) \textit{``...a fluffy white rabbit...eat the moon...''}}}
        & \multicolumn{3}{@{}c@{}}{%
            \parbox[b]{\dimexpr3\qualFrameWidth+2\qualImageGap\relax}{\centering
            (b) \textit{``...frozen landscape scene inspired by the CN tower''}}} \\
        \noalign{\vskip2pt}
        \qualMethod{Self\\Forcing} &
        \qualFrame{seed_324_individual/self_forcing/frame_000_0s.jpg} &
        \qualFrame{seed_324_individual/self_forcing/frame_040_2p5s.jpg} &
        \qualFrame{seed_324_individual/self_forcing/frame_080_5s.jpg} &
        \qualFrame{prompt_000268_individual/self_forcing/frame_000_0s.jpg} &
        \qualFrame{prompt_000268_individual/self_forcing/frame_040_2p5s.jpg} &
        \qualFrame{prompt_000268_individual/self_forcing/frame_080_5s.jpg} \\
        \noalign{\vskip\qualImageGap}
        \qualMethod{Causal\\Forcing} &
        \qualFrame{seed_324_individual/causal_forcing/frame_000_0s.jpg} &
        \qualFrame{seed_324_individual/causal_forcing/frame_040_2p5s.jpg} &
        \qualFrame{seed_324_individual/causal_forcing/frame_080_5s.jpg} &
        \qualFrame{prompt_000268_individual/causal_forcing/frame_000_0s.jpg} &
        \qualFrame{prompt_000268_individual/causal_forcing/frame_040_2p5s.jpg} &
        \qualFrame{prompt_000268_individual/causal_forcing/frame_080_5s.jpg} \\
        \noalign{\vskip\qualImageGap}
        \qualMethod{\textbf{Mass}\\\textbf{Forcing}\\\textbf{(Ours)}} &
        \qualFrame{seed_324_individual/ours/frame_000_0s.jpg} &
        \qualFrame{seed_324_individual/ours/frame_040_2p5s.jpg} &
        \qualFrame{seed_324_individual/ours/frame_080_5s.jpg} &
        \qualFrame{prompt_000268_individual/ours/frame_000_0s.jpg} &
        \qualFrame{prompt_000268_individual/ours/frame_040_2p5s.jpg} &
        \qualFrame{prompt_000268_individual/ours/frame_080_5s.jpg} \\
        \noalign{\vskip2pt}
        & 0\,s & 2.5\,s & 5\,s & 0\,s & 2.5\,s & 5\,s \\
    \end{tabular}
    \endgroup
    \caption{\textbf{Qualitative comparison with Self Forcing and Causal Forcing.}
    Two prompts illustrate character-pose changes and scene evolution, with frames at 0, 2.5, and 5 seconds and methods aligned by row.
    Mass Forcing preserves detailed appearance as pose and viewpoint change.}
    \label{fig:qualitative}
\end{figure*}
 
\subsection{Evaluation Results}
\label{sec:exp_results}

\paragraph{Quantitative comparison.}
Mass Forcing achieves the highest Total and Quality scores among the compared methods on both benchmarks (\cref{tab:main-table}).
On MovieGen, it obtains 83.38 Total and 84.53 Quality, exceeding the strongest baseline, Causal Forcing, by 0.87 and 1.25 points, respectively.
On VBench, it obtains 84.98 Total, 86.17 Quality, and 80.23 Semantic, improving over Causal Forcing by 0.84, 0.95, and 0.40 points.
Relative to Self Forcing, the Quality gains are 1.76 points on MovieGen and 1.66 points on VBench.
These improvements do not imply dominance on every metric: MovieGen Semantic remains below Causal Forcing (78.78 versus 79.42), while the overall Total score improves.

\paragraph{Quality and dynamics.}
The dimension-level results show that the aggregate gains coexist with substantial motion rather than depending on nearly static outputs.
Compared with Self Forcing, Mass Forcing increases DD from 63.41\% to 83.35\% on MovieGen and from 61.11\% to 88.89\% on VBench, while improving Aesthetic Quality from 63.88 to 65.92 and from 65.90 to 67.08, respectively.
It also exceeds Causal Forcing in DD on both benchmarks with higher aggregate Quality.
Some individual metrics trade off: for example, VBench Imaging Quality is lower than Self Forcing's (66.93 versus 69.53; see \crefrange{tab:moviegen-quality-dimensions}{tab:vbench-semantic-dimensions}).

\paragraph{Qualitative comparison.}
\Cref{fig:qualitative} complements \cref{fig:teaser} with two further comparisons of Self Forcing, Causal Forcing, and Mass Forcing at matched timestamps.
In \cref{fig:qualitative}(a), Mass Forcing depicts the rabbit eating the moon, whereas Self Forcing and Causal Forcing do not visibly realize the requested eating action.
\Cref{fig:qualitative}(b) highlights scene-level movement: Mass Forcing produces a changing view of the frozen, CN-tower-inspired landscape while preserving layered ice detail.
Together, these examples illustrate prompt-aligned action and scene movement alongside visual detail, complementing the aggregate measurements.
More five-frame comparisons with full prompts appear in \cref{sec:extended_qualitative}.

\subsection{Behavior During Continued Training}
\label{sec:exp_trajectories}

\paragraph{Temporal dynamics.}
\Cref{fig:teaser} compares DD over 1,000 generator updates using $100$ MovieGen prompts with $10$ seeds per prompt at each evaluation point.
Between updates 100 and 1,000, Self Forcing declines from 66\% to 13\% DD and Causal Forcing from 76\% to 12\%, whereas Mass Forcing remains between 65\% and 92\% and finishes at 83\%.
The trajectories are not monotonic, but they distinguish a sustained dynamic regime from the baselines' pronounced late-training degradation.
Beyond these training trajectories, the full-benchmark evaluations in \cref{tab:main-table} show that Mass Forcing improves DD over Self Forcing from 63.41\% to 83.35\% on MovieGen and from 61.11\% to 88.89\% on VBench.
\Cref{fig:extended_scenes,fig:extended_subjects} in \cref{sec:extended_qualitative} provide further qualitative evidence that Mass Forcing produces more dynamic videos than state-of-the-art baselines.

\subsection{Ablation Studies}
\label{sec:exp_ablation}

\Cref{tab:ablation} examines the conditional-chunk data path, guidance decoupling, set-level supervision, and mass-weighting design choices on MovieGen and VBench, with Ours from \cref{tab:main-table} as the reference.
Dimension-level results and video counts appear in \cref{sec:appendix_ablation}, with a visual comparison in \cref{sec:qualitative_ablation}.

\paragraph{Matched unweighted baseline.}
Mass Forcing changes the training data path as well as the objective, replacing holistic rollouts with conditional-chunk updates on rollouts that share a prefix.
To isolate the data path, we train a uniform conditional-chunk baseline that is identical to Mass Forcing except $\lambda_{\max}=0$, which gives $h_k=1$, and $\gamma=0$ (\cref{sec:appendix_matched_baseline}).
Its Dynamic Degree falls from 65\% at update 100 to 11\% at update 1{,}000, like Self Forcing.
On the full benchmarks it reaches 81.55 Total on MovieGen and 81.98 on VBench, 1.83 and 3.00 points below Mass Forcing.
The data path alone therefore explains neither the quality gain nor the sustained dynamics, and the remaining ablations isolate the components that act on it.

\begin{table}[t]
\caption{Ablation results on MovieGen and VBench.}
\label{tab:ablation}
\centering
\footnotesize
\setlength{\tabcolsep}{2pt}
\setlength{\heavyrulewidth}{0.75pt}
\setlength{\lightrulewidth}{0.35pt}
\setlength{\cmidrulewidth}{0.4pt}
\renewcommand{\arraystretch}{1.0}
\begin{tabular*}{\linewidth}{@{\extracolsep{\fill}}l*{4}{c}@{}}
\toprule
\multirow{2}{*}{\textbf{Variant}} &
\multicolumn{2}{c}{\textbf{MovieGen}\,$\uparrow$} &
\multicolumn{2}{c@{}}{\textbf{VBench}\,$\uparrow$} \\
\cmidrule(l{\tabcolsep}r{\tabcolsep}){2-3}\cmidrule(l{\tabcolsep}r{0pt}){4-5}
& \textbf{Quality} & \textbf{Semantic} & \textbf{Quality} & \textbf{Semantic} \\
\midrule
\textbf{Ours (full method)} & \textbf{84.53} & \textbf{78.78} & \textbf{86.17} & \textbf{80.23} \\
\midrule
Uniform conditional-chunk & 82.57 & 77.47 & 84.53 & 71.78 \\
w/o CFG decoupling & 83.42 & 75.24 & 84.77 & 80.16 \\
w/o MMD & 83.35 & \underline{78.59} & 85.19 & 78.27 \\
Shuffled-ratio control & 83.22 & 77.48 & 84.68 & 79.89 \\
$K=2$ & 83.43 & 77.90 & 85.29 & 77.48 \\
Endpoint-only & 83.59 & 77.63 & \underline{85.34} & \underline{80.18} \\
$\lambda_{\max}=0.5$ & 83.22 & 76.97 & 84.25 & 80.00 \\
$\lambda_{\max}=1.0$ & \underline{83.67} & 78.38 & 85.21 & 79.17 \\
\bottomrule
\end{tabular*}
\par\vspace{2pt}
{\scriptsize w/o CFG decoupling also omits MMD.\par}
\end{table}
 
\paragraph{Guidance decoupling.}
The w/o CFG decoupling variant omits both decoupling and MMD, whereas w/o MMD retains decoupling.
With MMD absent in both, the decoupled variant has higher DD: 65.00\% versus 61.81\% on MovieGen and 77.78\% versus 65.28\% on VBench.
Quality changes little on MovieGen (83.35 versus 83.42) and improves on VBench (85.19 versus 84.77).
This supports decoupling for sustaining dynamics, and the gap to Ours reflects both.

\paragraph{Set-level supervision.}
The w/o MMD variant removes the set-level term by setting $\gamma=0$, retaining local mass-weighted score distillation.
Removing MMD reduces Quality from 84.53 to 83.35 on MovieGen and from 86.17 to 85.19 on VBench.
DD also decreases from 83.35\% to 65.00\% and from 88.89\% to 77.78\%, respectively, supporting the contribution of set-level matching to both quality and dynamics.
The loss of dynamics despite retaining local mass reweighting supports the need for interactions across rollouts, not only rescaled pointwise corrections.

\paragraph{Weight assignment.}
The shuffled-ratio control permutes the ratio weights among rollouts within each prefix group, preserving the group weight distribution while breaking the correspondence between a rollout and its estimated teacher-relative support.
Quality decreases to 83.22 on MovieGen and 84.68 on VBench, drops of 1.31 and 1.49 points from the reference.
MovieGen Semantic also decreases from 78.78 to 77.48, supporting the importance of rollout-specific weight assignment for both quality and semantic alignment.
DD also falls to 59.62\% and 69.44\%, linking correct weight assignment to sustained motion rather than only improvements in aggregate quality.
Since shuffling preserves the weight values, this comparison suggests that the benefit depends on which rollouts receive additional influence, not just the presence of nonuniform weights.

\paragraph{Rollout count and integration.}
Reducing the rollout count from $K=4$ to $K=2$ yields Quality scores of 83.43 on MovieGen and 85.29 on VBench, retaining much of the reference configuration's quality while trailing the larger rollout group.
The $K=2$ model retains high DD on VBench (87.50\%) but has lower Aesthetic Quality (66.13 versus 67.08), suggesting that additional rollout alternatives benefit visual quality beyond preserving motion.
Replacing Simpson integration with an endpoint-only estimate yields Quality scores of 83.59 and 85.34, respectively, also below the reference.
Endpoint-only weighting discards information accumulated between the anchor and each rollout.
Its lower scores favor the path over the endpoint alone.

\paragraph{Mixture strength.}
Increasing $\lambda_{\max}$ from 0.25 to 0.5 gives Quality scores of 83.22 on MovieGen and 84.25 on VBench.
Setting it to 1.0 gives 83.67 and 85.21, respectively.
Both settings remain below the default in Quality and Semantic on both benchmarks, favoring the moderate default.
Stronger reweighting therefore does not translate into larger gains: both higher coefficients also lower DD, suggesting that preserving a uniform student component helps balance motion and quality under finite-group mass estimates.

\section{Related Work}
\label{sec:related_work}

\paragraph{Bidirectional video generation.}
Bidirectional video diffusion models jointly denoise entire clips to capture visual detail and temporal coherence~\citep{yang_2025_iclr_cogvideox,kong_2024_hunyuanvideo,wan_2025_wan}.
Full temporal context helps coordinate appearance and motion across frames, but repeated clip-wide denoising makes high-resolution generation computationally expensive.
LTX-Video addresses this cost through latent compression~\citep{hacohen_2024_ltxvideo}, while rCM and Adaptive Video Distillation combine distillation with regularization to balance efficiency and quality~\citep{zheng_2026_iclr_rcm,you_2026_cvpr_adaptive}.
Clip-level processing, however, limits low-latency streaming.

\paragraph{Autoregressive video generation and distillation.}
Autoregressive models support streaming by conditioning generation on preceding context, but AR diffusion models retain iterative denoising costs~\citep{deng_2025_iclr_nova,jin_2025_iclr_pyramidflow,chen_2025_skyreelsv2,sandai_2025_magi1}.
CausVid enables few-step generation by matching a causal student's distribution to a bidirectional teacher's through DMD~\citep{yin_2025_cvpr_causvid}.
Self Forcing reduces exposure bias using student-generated histories~\citep{huang_2026_nips_self_forcing}, while Causal Forcing improves ODE initialization with an AR teacher~\citep{zhu_2026_icml_causal_forcing} and Causal-rCM broadens initial coverage through consistency distillation~\citep{zheng_2026_causalrcm}.
Further extensions improve long-video training and context handling~\citep{yang_2026_iclr_longlive,liu_2026_iclr_rolling_forcing,chen_2026_icml_context_forcing,fu_2026_eccv_ispa,dreamx_2026_dreamx_world}.
Infinity-RoPE~\citep{yesiltepe_2026_cvpr_infinity_rope} is a training-free inference technique for long rollouts of Self-Forcing-style models.
These DMD-based pipelines retain the mode-seeking reverse-KL objective.
Mass Forcing instead introduces coverage-oriented supervision throughout on-policy optimization.
See \cref{sec:appendix_related_work} for an extended discussion.
\section{Conclusions}
\label{sec:conclusion}

We introduced Mass Forcing, a probability-mass-aware framework that provides coverage-oriented supervision throughout on-policy DMD training for few-step autoregressive video generation.
Rather than relying on initialization alone, it makes continued distillation responsive to teacher-relative mass imbalances through complementary local and set-level objectives: estimated density ratios rebalance conditional score corrections while keeping CFG uniformly weighted, and define a rollout-set reference that provides distributional feedback beyond pointwise gradient reweighting.
Experiments on MovieGen and VBench show improved overall generation quality and sustained temporal dynamics during continued training, with no change to the standard four-step inference pipeline or its cost.

\paragraph{Limitations.}
All continuations come from the current student, so a teacher-supported mode it no longer generates cannot be up-weighted. Mass Forcing thus preserves coverage, which is why it is applied from the ODE initialization, but does not recover lost modes. The weights also rely on a learned fake score and a finite group of $K$ rollouts, so they are a coverage-oriented surrogate, not exact ratios.

 {
    \small

}

\clearpage
\onecolumn
\begin{center}
    \Large
    \textbf{\thetitle}\\
    \vspace{0.5em}Appendix
\end{center}

\appendix
\crefalias{section}{appendix}
\crefalias{subsection}{subappendix}
\crefalias{subsubsection}{subsubappendix}
\startcontents[appendix]
\begingroup
\hypersetup{hidelinks}
\section*{Appendix Contents}
\printcontents[appendix]{}{1}{\setcounter{tocdepth}{2}}
\endgroup

\paragraph{Organization.}
\Cref{sec:appendix_related_work,sec:appendix_derivations,sec:implementation_details} support the method: extended related work, a derivation of every equation in \cref{sec:method}, and implementation details with the training algorithm and its cost.
\Cref{sec:appendix_dimensions} gives the detailed results behind \cref{sec:exp}: dimension-level scores, visual quality during training, and the matched unweighted baseline.
\Cref{sec:appendix_analysis} analyzes the mechanism: conditional mode coverage measured against the teacher, and the accuracy and stability of the mass estimator.
\Cref{sec:appendix_additional_eval} reports further evaluations: 60-second videos, a user study, and a second training seed.
\Cref{sec:appendix_qualitative} collects additional qualitative comparisons.
\section{Extended Related Work}
\label{sec:appendix_related_work}

\subsection{Bidirectional Video Diffusion Models}
\label{sec:app_related_bidirectional}
Bidirectional video diffusion models synthesize clips via joint spatiotemporal denoising, leveraging full attention across frames to capture appearance and temporal coherence.
Early work extends image diffusion to video through joint image--video training~\citep{ho_2022_nips_video_diffusion} or temporal layers added to pretrained image models~\citep{blattmann_2023_cvpr_align_latents}.
Subsequent work improves visual fidelity by fine-tuning spatial modules with high-quality images while preserving motion learned from videos~\citep{chen_2024_cvpr_videocrafter2}.
Transformer-based models, including CogVideoX~\citep{yang_2025_iclr_cogvideox}, Sora~\citep{brooks_2024_sora}, HunyuanVideo~\citep{kong_2024_hunyuanvideo}, and Wan~\citep{wan_2025_wan}, scale this formulation to improve text-conditioned synthesis.
To reduce sampling cost, LTX-Video~\citep{hacohen_2024_ltxvideo} lowers per-step computation through spatiotemporal latent compression, while video distillation reduces the number of denoising steps.
To balance fidelity and diversity in few-step generation, rCM~\citep{zheng_2026_iclr_rcm} combines consistency distillation with score regularization, and Adaptive Video Distillation~\citep{you_2026_cvpr_adaptive} uses adaptive regression and temporal regularization to mitigate oversaturation and motion degradation.
Despite these efficiency gains, bidirectional models still jointly denoise entire clips, incurring latency and memory costs that increase with clip length and limit their suitability for low-latency interactive and streaming applications.

\subsection{Autoregressive Video Generation and Distillation}
\label{sec:app_related_autoregressive}
Autoregressive video pretraining models future frames or chunks conditioned on preceding context.
Early methods use discrete tokens for sequence modeling~\citep{hong_2023_iclr_cogvideo,kondratyuk_2024_icml_videopoet} or masked prediction with causal video tokenization~\citep{villegas_2023_iclr_phenaki}.
In continuous latent spaces, NOVA~\citep{deng_2025_iclr_nova} combines frame-wise autoregression with spatial set-wise prediction, while Pyramid Flow~\citep{jin_2025_iclr_pyramidflow} improves efficiency through pyramidal denoising and history compression.
Diffusion Forcing~\citep{chen_2024_nips_diffusion_forcing} introduces independent per-token noise levels, while SkyReels-V2 and MAGI-1~\citep{chen_2025_skyreelsv2,sandai_2025_magi1} scale temporally structured denoising to long videos.
However, these AR diffusion models still require many denoising steps per frame or chunk, limiting real-time generation.

Few-step AR distillation seeks to reduce denoising cost while preserving generation quality under autoregressive rollout.
AAPT~\citep{lin_2025_nips_aapt} explores adversarial student-forcing training for one-step streaming generation, whereas CausVid~\citep{yin_2025_cvpr_causvid} uses DMD~\citep{yin_2024_cvpr_dmd,yin_2024_nips_dmd2} to match a causal student's output distribution to that of a bidirectional teacher.
Within DMD-based pipelines, subsequent work improves both causal initialization and training under generated histories.
Causal Forcing~\citep{zhu_2026_icml_causal_forcing} addresses the bidirectional-to-causal architectural gap by using an AR teacher for ODE initialization, while Causal-rCM~\citep{zheng_2026_causalrcm} uses teacher-forcing consistency distillation to improve initial mode coverage.
During DMD training, Self Forcing~\citep{huang_2026_nips_self_forcing} reduces exposure bias by conditioning rollouts on student-generated histories.
For longer generation, LongLive~\citep{yang_2026_iclr_longlive} introduces streaming training and cache management, Rolling Forcing~\citep{liu_2026_iclr_rolling_forcing} uses joint rolling-window denoising, and Context Forcing~\citep{chen_2026_icml_context_forcing} provides long-context teacher supervision.
Despite improved initialization and temporal context handling, these DMD-based approaches retain reverse-KL distribution matching and its mode-seeking tendency.
In contrast, Mass Forcing modifies the training objective to explicitly promote mode coverage throughout on-policy DMD optimization.

\clearpage
\numberwithin{equation}{section}
\section{Mathematical Derivations of Mass Forcing}
\label{sec:appendix_derivations}

This appendix derives the statements of \cref{sec:method} in the order in which they appear there.
\Cref{sec:app_scope} formalizes the on-policy, per-state setting of \cref{sec:method_onpolicy}.
\Cref{sec:app_kl} derives the ideal weight and its coverage identity (\cref{sec:method_ratio}, ideal weight).
\Cref{sec:app_scores,sec:app_ratio} derive the four steps of the estimator (\cref{sec:method_ratio}, estimation).
\Cref{sec:app_local,sec:app_set} derive the two objectives of \cref{sec:method_objectives}, and \cref{sec:app_training} combines them and specifies the training of the fake denoiser.
Throughout, we distinguish exact population identities from the finite-group, learned-score surrogate used in training.
All parameter gradients are loss gradients, and gradient descent follows their negative.

\paragraph{Notation.}
We use the notation of the main text.
A state is a pair $(u,c)$ of a student-generated prefix and a prompt.
A continuation is $x=G_\theta(\epsilon;u,c)\in\mathbb R^d$ with sampler noise $\epsilon$, and its noised version is $y=\alpha_tx+\sigma_t\eta$ with corruption noise $\eta$ (\cref{eq:mf_noised_continuation}).
At a fixed state and noise level, $q$ and $p$ are the densities of $y$ under the student and the teacher, $s_q$ and $s_p$ are their scores, $r=p/q$ is their ratio, and $h^*$ is the ideal weight; we write $q_{\theta,t}$, $p_t$, $s_{q,t}$, and $s_{p,t}$ only where an average over noise levels is taken.
A tilde, as in $\tilde r$, marks a quantity conditioned on the noisy prefix.
$D_p$ and $D_q$ are the teacher and fake predictions, $\Delta$ is the estimated score difference, and $\Delta^*$ is the exact field that it estimates.
Rollouts are indexed by $k=0,\ldots,K-1$, overbars mark detached samples, and $x_k,y_k$ are the differentiable copies of $\bar x_k,\bar y_k$.
We abbreviate the reference weights by $\pi_k=h_k/K$.

\subsection{On-Policy States and the Fixed-Prefix Objective}
\label{sec:app_scope}

This subsection supports \cref{sec:method_onpolicy}: it states the trajectory-level objective of Self Forcing, the per-state objective analyzed in this paper, and how the two are related.

\paragraph{Autoregressive sampling and on-policy histories.}
In \cref{eq:mf_ar_factorization}, $B$ chunks of $d$ scalar latent entries form the vectorized video $V\in\mathbb R^{Bd}$.
Following CausVid and Self Forcing~\citep{yin_2025_cvpr_causvid,huang_2026_nips_self_forcing}, each chunk alternates denoising and re-noising at decreasing noise levels, then enters the clean history represented by a key--value (KV) cache.
The sampler noise $\epsilon$ in \cref{eq:mf_noised_continuation} includes both initial and intermediate re-noising draws.
Teacher Forcing conditions on clean data histories, while Diffusion Forcing uses noise-corrupted data histories with independently sampled noise levels.
Self Forcing instead constructs these histories from its own generated chunks, reducing exposure bias and enabling holistic distribution matching on the completed rollout.
Mass Forcing keeps these on-policy histories but supervises one sampled chunk per update, conditioned on its detached prefix (\cref{sec:method_onpolicy}).

\paragraph{Trajectory-level objective.}
Corrupting the full video as $V_t=\alpha_tV+\sigma_t\eta_V$, with $\eta_V\sim\mathcal N(0,I_{Bd})$, induces the noisy student distribution $Q_{\theta,t}$.
Let $P_t$ denote the corresponding noisy teacher distribution.
Both are conditioned on prompt $c$, and $I_{Bd}$ is the full-video identity matrix.
The holistic Self Forcing objective is
\begin{equation}
    \mathcal L_{\mathrm{DMD}}^{\mathrm{video}}
    =\mathbb E_t\!\left[\omega(t)
      D_{\mathrm{KL}}(Q_{\theta,t}\|P_t)\right],
    \label{eq:mf_video_rkl}
\end{equation}
where the frozen teacher provides its score and a separate fake denoiser estimates the student score from generated clips.
ODE and consistency-based initialization improve the starting generator~\citep{yin_2025_cvpr_causvid,zhu_2026_icml_causal_forcing,zheng_2026_causalrcm}, but do not change this subsequent reverse-KL objective.

\paragraph{Per-state setting.}
For a state $(u,c)$, let $x=G_\theta(\epsilon;u,c)\in\mathbb R^d$ and $y=\alpha_t x+\sigma_t\eta$ as in \cref{eq:mf_noised_continuation}, where $\epsilon$ contains the generator-rollout randomness and $\eta\sim\mathcal N(0,I_d)$ is independent score-query corruption.
Unless specified otherwise, $(u,c,t)$ are fixed, the teacher distribution does not depend on $\theta$, and $\sigma_t>0$.
For the population gradient derivations, assume differentiable densities and generator maps, valid interchange of differentiation and integration, and vanishing boundary terms in integration by parts.
Gaussian corruption supplies positive smooth densities even when the clean generator law is not described by a smooth density.
The remaining regularity and integrability conditions are separate assumptions.

\paragraph{Relation between the two objectives.}
Let $P(V\mid c)$ be the clean teacher video distribution and let $p(x^{(b)}\mid u^{(b)},c)$ denote its chain-rule conditionals.
This factorization does not require the teacher network itself to be autoregressive.
When the clean-video KL is finite, its chain-rule decomposition is
\begin{align}
 D_{\mathrm{KL}}(Q_\theta\|P)
 &=\mathbb E_{V\sim Q_\theta}\!\left[
   \sum_{b=1}^B\log\frac{q_\theta(x^{(b)}\mid u^{(b)},c)}
                              {p(x^{(b)}\mid u^{(b)},c)}\right] \notag\\
 &=\sum_{b=1}^B\mathbb E_{u^{(b)}\sim Q_\theta^{(<b)}}\!\left[
   D_{\mathrm{KL}}\!\left(q_\theta(\cdot\mid u^{(b)},c)
                  \|p(\cdot\mid u^{(b)},c)\right)\right],
 \label{eq:mf_app_ar_kl}
\end{align}
where $Q_\theta^{(<b)}$ is the student's prefix marginal and prompt conditioning is suppressed in the outer densities.
The video-level divergence is thus a sum over chunk positions of per-state divergences, each averaged over on-policy prefixes, which is the sense in which per-state supervision keeps the on-policy states of Self Forcing.
Differentiating this full objective also differentiates the prefix distribution.
Our per-state analysis instead holds the sampled prefix fixed, matching the detached-prefix update in the implementation.
The same KL chain rule applies to noisy full videos, but then the conditionals use noisy prefixes and are not the clean-prefix conditionals obtained by noising only the next chunk.

\paragraph{Exact fields and learned estimates.}
The scores $s_q$ and $s_p$ refer to population densities in the ideal per-state formulation.
The denoisers $D_q,D_p$ and the residual $R_\phi$ are network predictions.
Likewise, $\Delta^*$ denotes an exact conditional score-difference field, whereas $\Delta$ is its learned estimate.
Overbars on $\bar x_k,\bar y_k$ mark detached reference samples.
The unbarred $x_k,y_k$ are differentiable replays with the same forward values at the start of a generator update.

\subsection{The Ideal Weight and Its Coverage Identity}
\label{sec:app_kl}

This subsection supports the ideal weight of \cref{sec:method_ratio}.
It derives the reverse-KL gradient of \cref{eq:mf_rkl_gradient}, shows that weighting it by the ratio $r$ gives the forward-KL gradient, and obtains the weight $h^*$ of \cref{eq:mf_ideal_mass} and the coverage identity of \cref{eq:mf_coverage} from their mixture.

\paragraph{A transport identity for implicit generators.}
Fix $(u,c,t)$ and write $q$, $p$, and $r=p/q$ as in the main text.
The noisy sample is $y_\theta=\alpha_tG_\theta(\epsilon;u,c)+\sigma_t\eta$, with $\partial y_\theta/\partial\theta=\alpha_tJ_\theta$.
Define the conditional mean Jacobian
\begin{equation}
 \mathcal A_\theta(y)
 =\mathbb E[\alpha_tJ_\theta\mid y_\theta=y].
 \label{eq:mf_app_transport_matrix}
\end{equation}
It has one column per generator parameter.
For any smooth compactly supported test function $\psi$, differentiation through the sample and integration by parts give
\begin{align}
 \nabla_\theta\int q(y)\psi(y)\,\mathrm dy
 &=\mathbb E[\alpha_tJ_\theta^\top\nabla_y\psi(y_\theta)] \notag\\
 &=\int q(y)\mathcal A_\theta(y)^\top\nabla_y\psi(y)\,\mathrm dy \notag\\
 &=-\int\psi(y)\operatorname{div}_y
                  [q(y)\mathcal A_\theta(y)]\,\mathrm dy.
 \label{eq:mf_app_test_function}
\end{align}
Thus the density derivative satisfies
\begin{equation}
 \nabla_\theta q(y)
 =-\operatorname{div}_y[q(y)\mathcal A_\theta(y)],
 \label{eq:mf_app_continuity}
\end{equation}
where divergence is applied columnwise and returned as a parameter-gradient vector.
This argument does not require an invertible generator or a one-to-one map from $\epsilon$ to $y$.

\paragraph{Reverse KL.}
The per-state DMD objective is
\begin{equation}
 \mathcal L_{\mathrm{RKL}}
 =\mathbb E_t\!\left[\omega(t)
     D_{\mathrm{KL}}(q_{\theta,t}\|p_t)\right].
 \label{eq:mf_rkl}
\end{equation}
At a fixed noise level, the teacher is constant with respect to $\theta$ and
\begin{align}
 \nabla_\theta D_{\mathrm{KL}}(q\|p)
 &=\int\left(\log\frac{q}{p}+1\right)\nabla_\theta q\,\mathrm dy \notag\\
 &=-\int\left(\log\frac{q}{p}+1\right)
       \operatorname{div}_y(q\mathcal A_\theta)\,\mathrm dy \notag\\
 &=\int q\mathcal A_\theta^{\top}
            \nabla_y\log\frac{q}{p}\,\mathrm dy \notag\\
 &=\mathbb E_{\epsilon,\eta}\!\left[
       \alpha_tJ_\theta^{\top}(s_q-s_p)\right].
 \label{eq:mf_app_rkl_gradient_steps}
\end{align}
The constant term vanishes because $\int\nabla_\theta q\,\mathrm dy=0$.
Averaging over $t$ with $\omega(t)$ recovers \cref{eq:mf_rkl_gradient}, the standard score-difference gradient underlying DMD~\citep{yin_2024_cvpr_dmd,yin_2024_nips_dmd2}.
The expectation is taken over student samples, which is the starting point of \cref{sec:method_ratio}.

\paragraph{Forward KL.}
Keeping the same generator distribution and fixed teacher,
\begin{align}
 \nabla_\theta D_{\mathrm{KL}}(p\|q)
 &=-\int\frac{p}{q}\nabla_\theta q\,\mathrm dy \notag\\
 &=\int r\operatorname{div}_y(q\mathcal A_\theta)\,\mathrm dy \notag\\
 &=-\int q\mathcal A_\theta^{\top}\nabla_y r\,\mathrm dy.
 \label{eq:mf_app_fkl_steps}
\end{align}
Since
\begin{equation}
 \nabla_y r=r\nabla_y\log(p/q)=r(s_p-s_q),
 \label{eq:mf_app_ratio_derivative}
\end{equation}
the forward-KL gradient is
\begin{equation}
 \nabla_\theta D_{\mathrm{KL}}(p\|q)
 =\mathbb E_{\epsilon,\eta}\!\left[
   \alpha_tJ_\theta^\top r(y)(s_q(y)-s_p(y))\right].
 \label{eq:mf_app_fkl_gradient}
\end{equation}
Relative to reverse KL, the correction is weighted by the teacher-to-student density ratio, consistent with the general divergence-gradient analysis of $f$-Distill~\citep{xu_2025_fdistill}.

\paragraph{Mixed divergence and the ideal weight.}
For a fixed $\lambda\in[0,1]$, define
\begin{align}
 \mathcal L_{\mathrm{mix}}
 &=\mathbb E_t\!\left[\omega(t)\bigl(
      (1-\lambda)D_{\mathrm{KL}}(q_{\theta,t}\|p_t)
      +\lambda D_{\mathrm{KL}}(p_t\|q_{\theta,t})\bigr)\right], \notag\\
 \nabla_\theta\mathcal L_{\mathrm{mix}}
 &=\mathbb E_{t,\epsilon,\eta}\!\left[
      \omega(t)\alpha_tJ_\theta^\top
      h^*(y)(s_{q,t}(y)-s_{p,t}(y))\right],
 \label{eq:mf_app_mixed_gradient}
\end{align}
with the ideal weight $h^*=1-\lambda+\lambda r$ of \cref{eq:mf_ideal_mass}, evaluated at the noise level of $y$.
This is \cref{eq:mf_rkl_gradient} with each student sample weighted by $h^*$.
Multiplying the sampling density by this weight gives
\begin{equation}
 q\,h^*
 =(1-\lambda)q+\lambda p,
 \qquad
 \mathbb E_{q}[h^*]=1,
 \label{eq:mf_app_mass_identity}
\end{equation}
which is the second line of \cref{eq:mf_ideal_mass} and the statement that the mean weight is one.

\paragraph{Coverage identity.}
Applying \cref{eq:mf_app_mass_identity} to the per-sample correction $\mathcal A_\theta^\top(s_q-s_p)$ splits the weighted gradient at a fixed noise level into an average over student samples and an average over teacher samples,
\[
 \mathbb E_{y\sim q}\!\left[h^*\mathcal A_\theta^\top(s_q-s_p)\right]
 =(1-\lambda)\,\mathbb E_{y\sim q}\!\left[\mathcal A_\theta^\top(s_q-s_p)\right]
  +\lambda\,\mathbb E_{y\sim p}\!\left[\mathcal A_\theta^\top(s_q-s_p)\right],
\]
which is \cref{eq:mf_coverage} with the Jacobian factor included.
The first term is $(1-\lambda)$ times the reverse-KL gradient of \cref{eq:mf_app_rkl_gradient_steps}, and the second is $\lambda$ times the forward-KL gradient of \cref{eq:mf_app_fkl_gradient}, rewritten as an expectation over teacher samples through $q\,r=p$.
This is an identity for the effective measure in the gradient estimate, not a change to the generator's inference sampler.
The derivation already accounts for the dependence of $q$ and $r$ on $\theta$.
Once this gradient field has been evaluated, differentiating through its score or ratio factors again would not reproduce the derived first-order gradient.
The implementation instead injects the detached field through a surrogate loss (\cref{sec:app_local}).

\subsection{Estimation, Steps (i) and (ii): From Denoising Predictions to the Score Difference}
\label{sec:app_scores}

This subsection supports steps (i) and (ii) of the estimator in \cref{sec:method_ratio}.
It specifies the shared-prefix, shared-noise queries, derives \cref{eq:mf_score_ratio,eq:mf_score_field} from the Gaussian denoising identity, and shows why the estimated ratio is the noisy-prefix proxy $\tilde r$.

\paragraph{Step (i): shared-prefix, shared-noise queries.}
For $K$ detached continuations $\bar x_k$ generated with independent sampler noises, the score inputs are
\begin{equation}
    \bar y_k=\alpha_t\bar x_k+\sigma_t\eta,
    \qquad u_t=\alpha_tu+\sigma_t\eta_u.
    \label{eq:mf_query_noise}
\end{equation}
The current-chunk corruption $\eta\sim\mathcal N(0,I_d)$ is shared across the group.
The prefix corruption $\eta_u$ is an independent standard Gaussian tensor with the prefix's shape and is also held fixed across queries.
This sharing applies only to score-query corruption, not to the independent sampler noises $\epsilon_k$.
The networks receive the joint clip $[u_t,\bar y_k]$ and return clean-latent predictions.
Only the continuation block enters the generator correction.

\paragraph{Gaussian denoising identity.}
For a fixed conditioning state, write the noisy student density as
\begin{equation}
 q(y)
 =\int\mathcal N(y;\alpha_t x,\sigma_t^2 I_d)\,q_\theta(\mathrm dx),
 \label{eq:mf_app_gaussian_convolution}
\end{equation}
where $q_\theta(\mathrm dx)$ denotes the clean student probability measure.
Differentiating the Gaussian kernel yields
\begin{align}
 \nabla_y\log q(y)
 &=\frac{1}{q(y)}\int
    \frac{\alpha_t x-y}{\sigma_t^2}
    \mathcal N(y;\alpha_t x,\sigma_t^2 I_d)\,q_\theta(\mathrm dx) \notag\\
 &=\frac{\alpha_t\mathbb E_q[x\mid y]-y}{\sigma_t^2}.
 \label{eq:mf_app_denoising_score}
\end{align}
Let $D_q^*(y)=\mathbb E_q[x\mid y]$ and $D_p^*(y)=\mathbb E_p[x\mid y]$ be the population-optimal clean-latent denoisers under the same conditioning and corruption.
Applying \cref{eq:mf_app_denoising_score} to both distributions cancels the common $-y/\sigma_t^2$ term:
\begin{equation}
 s_p(y)-s_q(y)
 =\frac{\alpha_t}{\sigma_t^2}\bigl(D_p^*(y)-D_q^*(y)\bigr).
 \label{eq:mf_app_score_difference}
\end{equation}
By \cref{eq:mf_score_ratio}, the left-hand side is $\nabla_y\log r(y)$.
Replacing the population denoisers by their network estimates gives the estimated field of \cref{eq:mf_score_field}.

\paragraph{Noisy-prefix conditionals in bidirectional joint queries.}
For the jointly corrupted prefix and continuation, let $p^{\mathrm{joint}}(u_t,y\mid c)$ and $q^{\mathrm{joint}}(u_t,y\mid c)$ denote the teacher and student densities, and let $p^{\mathrm{pre}},q^{\mathrm{pre}}$ be their prefix marginals.
The conditional densities satisfy
\begin{equation}
 \tilde p(y\mid u_t,c)
 =\frac{p^{\mathrm{joint}}(u_t,y\mid c)}{p^{\mathrm{pre}}(u_t\mid c)},
 \qquad
 \tilde q(y\mid u_t,c)
 =\frac{q^{\mathrm{joint}}(u_t,y\mid c)}{q^{\mathrm{pre}}(u_t\mid c)}.
 \label{eq:mf_app_joint_conditionals}
\end{equation}
At fixed $u_t$, the prefix marginals are constant with respect to $y$, so
\begin{align}
 \nabla_y\log p^{\mathrm{joint}}(u_t,y\mid c)
 &=\nabla_y\log\tilde p(y\mid u_t,c), \notag\\
 \Delta^*(y)
 &=\nabla_y\log\frac{\tilde p(y\mid u_t,c)}{\tilde q(y\mid u_t,c)}
  =\nabla_y\log\tilde r(y\mid u_t,c).
 \label{eq:mf_app_conditional_score}
\end{align}
Thus the current-chunk component of a bidirectional joint score supplies a conditional score at the fixed noisy prefix, so a separate causal teacher is not needed for this identity.
The shared prefix also ensures that prefix-marginal density factors cancel when comparing continuations within a group, which is why the continuations of one state are directly comparable (\cref{sec:method_onpolicy}).
This result does not identify conditioning on $u_t$ with conditioning on the exact clean $u$: $\tilde r$ is a noisy-prefix proxy for the clean-prefix ratio $r$.
For the empty prefix, the same expressions reduce to prompt-conditioned continuation densities.

\paragraph{Step (ii): residual parameterization of the score difference.}
The implementation uses $D_q=D_p+R_\phi$, where $D_p$ is the frozen conditional teacher prediction and $R_\phi$ is a residual estimator with trainable parameters $\phi$.
Consequently, the estimated field of \cref{eq:mf_score_field} is
\begin{equation}
 \Delta(y)
 =\frac{\alpha_t}{\sigma_t^2}(D_p-D_q)
 =-\frac{\alpha_t}{\sigma_t^2}R_\phi(y).
 \label{eq:mf_residual_score}
\end{equation}
The cancellation is algebraic at every queried point, including integration midpoints, and is specific to the conditional teacher used as the residual baseline.
Whether this estimated field matches $\Delta^*$ depends on denoiser calibration, including at interpolated inputs.

\subsection{Estimation, Steps (iii) and (iv): Line Integration and Normalized Weights}
\label{sec:app_ratio}

This subsection supports steps (iii) and (iv) of the estimator in \cref{sec:method_ratio}.
It derives the line integral of \cref{eq:mf_line_integral} and its quadrature, explains why the corruption noise is shared, and derives the normalized weights of \cref{eq:mf_weights} together with their limits of validity.

\paragraph{Step (iii): recovering relative log ratios.}
All rollouts in a group share $(u_t,t,c)$.
Choose the medoid anchor $a=\bar y_{k^*}$ with $k^*=\arg\min_j\sum_k\|\bar y_j-\bar y_k\|_2$, and use the path $\zeta_k(\xi)=a+\xi(\bar y_k-a)$, $\xi\in[0,1]$.
For the exact field $\Delta^*=\nabla_y\log\tilde r$, the chain rule gives
\begin{equation}
 \frac{\mathrm d}{\mathrm d\xi}\log\tilde r(\zeta_k(\xi))
 =\Delta^*(\zeta_k(\xi))^\top(\bar y_k-a).
 \label{eq:mf_app_path_derivative}
\end{equation}
Integrating from $0$ to $1$ yields
\begin{equation}
 \int_0^1\Delta^*(\zeta_k(\xi))^\top(\bar y_k-a)\,\mathrm d\xi
 =\log\tilde r(\bar y_k)-\log\tilde r(a).
 \label{eq:mf_app_path_result}
\end{equation}
The left-hand side is the line integral of \cref{eq:mf_line_integral} along the straight path, and this is the sense in which its second relation is an equality for the exact score difference.
The common anchor leaves only an additive constant shared by the group, so absolute density normalizers are unnecessary for relative weighting.
This is integration in latent space, not integration over the diffusion time coordinate.

\paragraph{Three-point quadrature.}
Replacing the exact field by $\Delta$ and applying Simpson's rule to the scalar path integrand gives
\begin{equation}
 \ell_k=\frac16\left\langle
   \Delta(a)+4\Delta\!\left(\frac{a+\bar y_k}{2}\right)
   +\Delta(\bar y_k),\;\bar y_k-a
 \right\rangle.
 \label{eq:mf_simpson}
\end{equation}
This is the first relation of \cref{eq:mf_line_integral}.
The anchor rollout has zero displacement and hence $\ell_{k^*}=0$.
The coefficients integrate a cubic scalar integrand exactly, but a neural score field need not vary cubically along the path.
Score-estimation error and quadrature error are distinct: even an exact score field can require finer quadrature when it changes sharply between modes.
A learned residual field also need not be conservative, in which case different paths or anchors can yield different estimates rather than one globally consistent log-density ratio.

\paragraph{Why the corruption noise is shared.}
Write $a=\alpha_t\bar x_{k^*}+\sigma_t\eta$ and $\bar y_k=\alpha_t\bar x_k+\sigma_t\eta$.
Then every path point is
\begin{equation}
 \zeta_k(\xi)
 =\alpha_t\bigl[(1-\xi)\bar x_{k^*}+\xi\bar x_k\bigr]
   +\sigma_t\eta,
 \qquad
 \bar y_k-a=\alpha_t(\bar x_k-\bar x_{k^*}).
 \label{eq:mf_app_shared_noise_path}
\end{equation}
The Gaussian corruption variance is therefore $\sigma_t^2 I_d$ throughout the path.
For fixed endpoint indices with independent noises $\eta_{k^*}$ and $\eta_k$, interpolation would instead give
\begin{equation}
 \operatorname{Cov}\!\left[\sigma_t\bigl((1-\xi)\eta_{k^*}
                                  +\xi\eta_k\bigr)\right]
 =\sigma_t^2\bigl((1-\xi)^2+\xi^2\bigr)I_d,
 \label{eq:mf_app_independent_noise_path}
\end{equation}
which equals $\sigma_t^2 I_d/2$ at the midpoint despite querying the networks at noise level $t$.
Shared corruption removes this variance mismatch, although the interpolated clean content can still be outside the rollout distribution.
At $\alpha_t=0$, the shared-noise particles coincide, all path displacements vanish, and the estimator carries no continuation-specific mass information.

\paragraph{Step (iv): anchor cancellation and ideal importance weighting.}
To isolate normalization, first consider exact ratios $r=p/q$ and ideal IID reference particles from that same proposal $q$.
With $\ell_k^*=\log r(\bar y_k)-\log r(a)$, exponentiation and normalization give
\begin{equation}
 \frac{\exp(\ell_k^*)}{\sum_j\exp(\ell_j^*)}
 =\frac{r(\bar y_k)}{\sum_j r(\bar y_j)}.
 \label{eq:mf_app_anchor_cancellation}
\end{equation}
The anchor term has cancelled, and multiplying by $K$ gives the ratio of rollout $k$ divided by the group average, as stated after \cref{eq:mf_weights}.
For bounded test functions $\psi$, the corresponding self-normalized estimate obeys
\begin{equation}
 \frac{\sum_k r(\bar y_k)\psi(\bar y_k)}{\sum_k r(\bar y_k)}
 \;\longrightarrow\;
 \frac{\mathbb E_q[r\psi]}{\mathbb E_q[r]}
 =\mathbb E_p[\psi]
 \quad\text{as }K\to\infty.
 \label{eq:mf_app_ideal_snis}
\end{equation}
This is an IID population benchmark, not an exact statement about the practical shared-noise groups.
Even in the IID case, the finite-$K$ ratio of sample sums is generally biased.

\paragraph{Stabilized mixture weights.}
The logits in \cref{eq:mf_weights} are computed as
\begin{equation}
 \bar\ell=\frac1K\sum_j\ell_j,
 \qquad v_k=\operatorname{clip}\!\left(
     \frac{\ell_k-\bar\ell}{\tau},-\kappa,\kappa\right),
 \label{eq:mf_ratio_logits}
\end{equation}
where $\bar\ell$ is the group mean, $\tau>0$ is the temperature, and $\kappa>0$ bounds the logits \emph{after} temperature scaling.
Centering removes the common anchor offset before clipping, and the subsequent softmax yields the relative estimates $\widehat r_k$.
The detached, group-shared coefficient $\lambda_{\mathrm{eff}}\in[0,\lambda_{\max}]$, $\lambda_{\max}\le1$, incorporates the training ramp and reliability gates (\cref{sec:impl_mass}).
Without clipping, temperature scaling replaces $r$ by $r^{1/\tau}$.
In the ideal IID limit, the reweighted density is proportional to $p^{1/\tau}q^{1-1/\tau}$ rather than $p$ unless $\tau=1$.
Clipping further changes this weighting and bounds its concentration.
By construction,
\begin{equation}
 \sum_{k=0}^{K-1}\widehat r_k=K,
 \qquad \sum_{k=0}^{K-1} h_k=K,
 \qquad \sum_{k=0}^{K-1}\pi_k=1,
 \qquad \pi_k=\frac{h_k}{K},
 \label{eq:mf_app_weight_normalization}
\end{equation}
so the weights average to one over the group.
For $0\le\lambda_{\mathrm{eff}}\le1$, positivity of the normalized ratios also gives
\begin{equation}
 1-\lambda_{\mathrm{eff}}\le h_k
 \le 1-\lambda_{\mathrm{eff}}+\lambda_{\mathrm{eff}}K.
 \label{eq:mf_app_weight_bounds}
\end{equation}
The reference measure $\nu$ of \cref{eq:mf_empirical_target} can be written explicitly as
\begin{align}
 \nu&=(1-\lambda_{\mathrm{eff}})\mu_{\mathrm{ref}}
         +\lambda_{\mathrm{eff}}\widehat\nu_p, \notag\\
 \mu_{\mathrm{ref}}&=\frac1K\sum_k\delta_{\bar y_k},
 &\widehat\nu_p&=\frac1K\sum_k\widehat r_k\delta_{\bar y_k}.
 \label{eq:mf_app_empirical_mixture}
\end{align}
Here $\mu_{\mathrm{ref}}$ is the uniform detached reference measure and $\widehat\nu_p$ is the estimated teacher-reweighted measure, not a set of independently generated teacher samples.
This is the finite-group counterpart of the mixture $q\,h^*=(1-\lambda)q+\lambda p$ in \cref{eq:mf_ideal_mass}.

\paragraph{Shared-noise dependence and conditioning mismatch.}
The actual particles share one corruption draw $\eta$.
Conditioned on this draw and the clean prefix, increasing $K$ averages over generator randomness, not over the marginal noisy proposal $q$.
Even if the exact clean-prefix ratio $r$ were available, the conditional large-group limit would be
\begin{equation}
 \frac{\mathbb E_\epsilon[r(y_\theta)\psi(y_\theta)\mid\eta,u,c]}
      {\mathbb E_\epsilon[r(y_\theta)\mid\eta,u,c]},
 \qquad y_\theta=\alpha_tG_\theta(\epsilon;u,c)+\sigma_t\eta,
 \label{eq:mf_app_crn_limit}
\end{equation}
which is not generally $\mathbb E_{p}[\psi]$.
Averaging these normalized ratios over $\eta$ does not generally remove the difference.
In addition, the joint-score queries estimate a noisy-prefix ratio $\tilde r$, whereas a group is generated from one fixed clean prefix.
These facts, together with clipping, tempering, and learned scores, are why we interpret the practical weights as a finite-group coverage surrogate rather than exact forward-KL importance weights.

\subsection{Mass-Weighted Score Distillation}
\label{sec:app_local}

This subsection supports the first objective of \cref{sec:method_objectives}.
It derives the decoupled correction of \cref{eq:mf_local_gradient}, relates it to the ideally weighted gradient of \cref{sec:app_kl}, and shows that the loss of \cref{eq:mf_local_loss} injects this correction into the generator.

\paragraph{Separating conditional transport from guidance.}
At a common noisy input and conditioning state, the unguided and guided teacher predictions satisfy
\begin{align}
 D_{\mathrm{cfg}}
 &=D_{\mathrm{uncond}}+w_{\mathrm{cfg}}(D_p-D_{\mathrm{uncond}}), \notag\\
 D_q-D_{\mathrm{cfg}}
 &=(D_q-D_p)+(D_p-D_{\mathrm{cfg}}).
 \label{eq:mf_app_cfg_decomposition}
\end{align}
The first component compares student and conditional teacher, and the second supplies the additional CFG correction.
Because the mass estimates target the conditional teacher, applying $h_k$ to the entire guided correction would also multiply the guidance component by $h_k$.
The difference between that update and the chosen decomposition is
\begin{equation}
 h_k(D_{q,k}-D_{\mathrm{cfg},k})
 -\bigl[h_k(D_{q,k}-D_{p,k})+(D_{p,k}-D_{\mathrm{cfg},k})\bigr]
 =(h_k-1)(D_{p,k}-D_{\mathrm{cfg},k}).
 \label{eq:mf_app_guidance_modulation}
\end{equation}
\Cref{eq:mf_local_gradient} removes this additional modulation and reduces exactly to the unweighted guided correction when $h_k=1$.
It is a conditional mass correction plus uniform guidance, not the mixed-KL gradient to a single CFG-guided target distribution.

\paragraph{Denoiser units and normalization.}
Combining the Jacobian factor from \cref{eq:mf_rkl_gradient} with the Gaussian identity gives the ideal conditional contribution
\begin{equation}
 \omega(t)\alpha_tJ_\theta^\top h^*(s_q-s_p)
 =\omega(t)\frac{\alpha_t^2}{\sigma_t^2}
       J_\theta^\top h^*(D_q^*-D_p^*).
 \label{eq:mf_app_score_to_x0_gradient}
\end{equation}
The implementation instead uses network predictions and the per-rollout normalization
\begin{equation}
 n_k=\operatorname{sg}\!\left[
   \max\!\left(\frac1d\|\bar x_k-D_{\mathrm{cfg},k}\|_1,\epsilon_n\right)
 \right],\qquad \epsilon_n>0.
 \label{eq:mf_app_dmd_normalizer}
\end{equation}
Here $\epsilon_n$ is a numerical floor, not a Gaussian noise variable.
This sample-dependent normalization is not merely a noise-level weight $\omega(t)$, so the practical normalized update is not asserted to optimize the exact KL mixture in \cref{eq:mf_app_mixed_gradient}.

\paragraph{Injecting a prescribed correction.}
We regenerate $x_k=G_\theta(\epsilon_k;u,c)$ with the same sampler noise and unchanged generator parameters as the detached reference $\bar x_k$.
The forward values agree, while $x_k$ retains the generator graph.
The local surrogate in \cref{eq:mf_local_loss} is
\begin{equation}
 \mathcal L_{\mathrm{local}}
 =\frac{1}{2Kd}\sum_k
   \left\|x_k-\operatorname{sg}(x_k-g_k)\right\|_2^2.
 \label{eq:mf_local_surrogate}
\end{equation}
Here $x_k^{\mathrm{tgt}}=\operatorname{sg}(x_k-g_k)$ is a detached gradient-injection target, not a teacher-generated clean video.
During differentiation, the scores, $h_k$, $n_k$, and this target are fixed.
The derivative of \cref{eq:mf_local_surrogate} is therefore
\begin{align}
 \nabla_{x_k}\mathcal L_{\mathrm{local}}
 &=\frac{1}{Kd}(x_k-x_k^{\mathrm{tgt}})
   =\frac{g_k}{Kd}, \notag\\
 \nabla_\theta\mathcal L_{\mathrm{local}}
 &=\frac{1}{Kd}\sum_k J_{\theta,k}^\top g_k,
 \label{eq:mf_app_local_parameter_gradient}
\end{align}
where $J_{\theta,k}=\partial x_k/\partial\theta$ is the Jacobian of the replayed generator graph.
The factor $1/(Kd)$ comes from averaging over rollouts and latent coordinates.
Stopping the target derivative does not remove the correction: it is the derivative of the outer squared distance that supplies $g_k$.
No generator-update gradient passes through the teacher, fake-score estimator, or mass weights.

\subsection{Mass-Targeted Rollout-Set Matching}
\label{sec:app_set}

This subsection supports the second objective of \cref{sec:method_objectives}.
It shows why the local correction can vanish while the weights remain informative, expands the set-level loss of \cref{eq:mf_total_loss}, and derives the gradient of \cref{eq:mf_set_replay_gradient}.

\paragraph{Mode probabilities need not determine within-mode scores.}
Consider teacher and student mixtures with common normalized component densities $f_m(y)$ but different positive mode probabilities $\rho_m^{(p)},\rho_m^{(q)}$:
\begin{equation}
 \tilde p(y)=\sum_m\rho_m^{(p)}f_m(y),\qquad
 \tilde q(y)=\sum_m\rho_m^{(q)}f_m(y),\qquad
 \sum_m\rho_m^{(p)}=\sum_m\rho_m^{(q)}=1.
 \label{eq:mf_app_mode_mixtures}
\end{equation}
Where a well-separated component $m$ dominates both densities,
\begin{align}
 \nabla_y\log\tilde p(y)
 &\approx\nabla_y[\log\rho_m^{(p)}+\log f_m(y)]
  =\nabla_y\log f_m(y), \notag\\
 \nabla_y\log\tilde q(y)&\approx\nabla_y\log f_m(y), \notag\\
 \log\tilde r(y)&\approx\log\rho_m^{(p)}-\log\rho_m^{(q)}.
 \label{eq:mf_app_weak_local_scores}
\end{align}
The within-mode corrections can therefore be small while the relative log-ratio offsets differ across modes.
For the positive, smoothed densities considered here, these offsets are connected by score-field variation between modes, which the line integral can recover when that field is accurately estimated and integrated.
If the exact field were zero everywhere along the connecting path, the offsets could not differ.
This distinction concerns weak endpoint corrections, not an absence of information throughout the field, and does not guarantee the accuracy of three-point quadrature.

\paragraph{Kernel discrepancy between the empirical measures.}
Let $\mathcal H_\beta$ be the reproducing-kernel Hilbert space associated with the Gaussian kernel
\begin{equation}
 \mathcal K_\beta(y,y')
 =\exp\!\left(-\frac{\|y-y'\|_2^2}{2d\beta}\right),
 \qquad \beta>0.
 \label{eq:mf_kernel}
\end{equation}
Its feature map $\Phi_\beta$ satisfies $\langle\Phi_\beta(y),\Phi_\beta(y')\rangle_{\mathcal H_\beta}=\mathcal K_\beta(y,y')$.
Using the empirical source and reference in \cref{eq:mf_empirical_target}, the squared MMD~\citep{gretton_2012_jmlr_mmd,li_2015_icml_gmmn} is
\begin{align}
 \mathcal L_{\mathrm{set}}
 &=\left\|\frac1K\sum_i\Phi_\beta(y_i)
             -\sum_j\pi_j\Phi_\beta(\bar y_j)\right\|_{\mathcal H_\beta}^2 \notag\\
 &=\frac{1}{K^2}\sum_{i,j}\mathcal K_\beta(y_i,y_j)
    -\frac{2}{K}\sum_{i,j}\pi_j\mathcal K_\beta(y_i,\bar y_j)
    +\sum_{i,j}\pi_i\pi_j\mathcal K_\beta(\bar y_i,\bar y_j).
 \label{eq:mf_app_mmd_expansion}
\end{align}
Every pair, including the diagonal, is retained.
The scale $\beta$ is the detached median of off-diagonal mean squared distances in the pooled source--reference set, subject to the lower bound $\beta_{\min}>0$.
It is held fixed during differentiation, together with the reference locations and weights.
Because the reference reuses the same rollouts, this is a nonnegative empirical matching objective, not an unbiased statistic from two independent sample sets.

\paragraph{Differentiating the source interactions.}
The derivative with respect to $y_i$ receives contributions from both appearances of $y_i$ in the source--source sum.
For the symmetric kernel,
\begin{align}
 \nabla_{y_i}\left[\frac1{K^2}\sum_{j,k}\mathcal K_\beta(y_j,y_k)\right]
 &=\frac1{K^2}\sum_j\left[
     \nabla_1\mathcal K_\beta(y_i,y_j)
     +\nabla_2\mathcal K_\beta(y_j,y_i)\right] \notag\\
 &=\frac2{K^2}\sum_j\nabla_1\mathcal K_\beta(y_i,y_j), \notag\\
 \nabla_{y_i}\left[-\frac2K\sum_{k,j}\pi_j\mathcal K_\beta(y_k,\bar y_j)\right]
 &=-\frac2K\sum_j\pi_j\nabla_1\mathcal K_\beta(y_i,\bar y_j).
 \label{eq:mf_app_mmd_terms}
\end{align}
Here $i,j,k$ index rollouts, and $\nabla_1,\nabla_2$ differentiate the first and second kernel arguments.
The reference--reference term has zero source derivative.
Combining the two contributions gives
\begin{equation}
 \nabla_{y_i}\mathcal L_{\mathrm{set}}
 =\frac2K\left[
     \frac1K\sum_j\nabla_1\mathcal K_\beta(y_i,y_j)
     -\sum_j\pi_j\nabla_1\mathcal K_\beta(y_i,\bar y_j)
   \right],
 \label{eq:mf_set_gradient}
\end{equation}
with
\begin{equation}
 \nabla_1\mathcal K_\beta(y,y')
 =-\frac{y-y'}{d\beta}\mathcal K_\beta(y,y').
 \label{eq:mf_app_kernel_gradient}
\end{equation}
The source--source term is essential.
Retaining only source--reference attraction would define a different objective.

\paragraph{Identical support does not imply a zero update.}
At the start of an update, $y_i=\bar y_i$ but the source weights are $1/K$ and the reference weights are $\pi_j=h_j/K$.
Consequently, the gradient stated in \cref{eq:mf_set_replay_gradient} is
\begin{equation}
 \left.\nabla_{y_i}\mathcal L_{\mathrm{set}}\right|_{y=\bar y}
 =\frac2{K^2}\sum_j(1-h_j)
                  \nabla_1\mathcal K_\beta(\bar y_i,\bar y_j).
 \label{eq:mf_set_mass_gradient}
\end{equation}
Nonuniform mass weights can therefore induce a geometric correction even when local score differences at the rollouts are weak.
When all $h_j=1$, the measures coincide, the MMD loss is zero, and every source gradient vanishes.
Nonuniform weights do not guarantee a useful nonzero update: coincident particles or negligible kernel interactions can still eliminate the signal.
The objective compares distributions over the sampled support.
It neither supplies absent teacher continuations nor proves recovery of all missing modes.

\subsection{Combined Generator Gradient and Fake-Score Training}
\label{sec:app_training}
\label{sec:method_training}

This subsection combines the two objectives into the total loss of \cref{eq:mf_total_loss} and specifies how the fake denoiser of step (ii) is trained.
The collection and replay procedure is specified in \cref{sec:impl_updates}.
Setting $\gamma=0$ retains only local mass-weighted distillation.
Also setting $\lambda_{\mathrm{eff}}=0$ recovers unweighted conditional-chunk DMD, not the holistic Self Forcing data path.

\paragraph{The generator chain rule.}
The differentiable source satisfies $y_k=\alpha_tG_\theta(\epsilon_k;u,c)+\sigma_t\eta$, with the same fixed corruption as its reference.
Thus $\partial y_k/\partial\theta=\alpha_tJ_{\theta,k}$, and
\begin{equation}
 \nabla_\theta\mathcal L_{\mathrm{set}}
 =\sum_k\alpha_tJ_{\theta,k}^\top\nabla_{y_k}\mathcal L_{\mathrm{set}}.
 \label{eq:mf_app_set_parameter_gradient}
\end{equation}
Combining this expression with \cref{eq:mf_app_local_parameter_gradient} yields
\begin{equation}
 \nabla_\theta\mathcal L_{\mathrm{MF}}
 =\sum_kJ_{\theta,k}^{\top}\left[
    \frac{g_k}{Kd}
    +\gamma\alpha_t\nabla_{y_k}\mathcal L_{\mathrm{set}}
   \right].
 \label{eq:mf_total_gradient}
\end{equation}
This is the gradient of the per-update surrogate with reference locations, mass weights, normalization, and kernel scale fixed.
The references are collected and replayed with identical randomness before any generator optimizer step, and the set derivative includes both source appearances in every pairwise interaction.
With a full differentiable rollout, $J_{\theta,k}$ is the full conditional generator Jacobian.
Under Self Forcing's memory-efficient truncation, it is the Jacobian of the retained computation graph: gradients do not traverse the cached prefix or detached intermediate denoising transitions~\citep{huang_2026_nips_self_forcing}.
This implemented derivative is therefore not the full gradient through all earlier autoregressive states in \cref{eq:mf_app_ar_kl}.

\paragraph{Fake-score objective.}
The fake denoiser tracks the evolving student distribution without applying the generator's mass weights to its samples.
For a detached student joint clip $Y_0=[u,\bar x]$, let $t$ and $\eta_Y$ denote its sampled noise level and standard Gaussian corruption, giving $Y_t=(1-\sigma_t)Y_0+\sigma_t\eta_Y$.
The paired implementation caches these denoising examples for critic replay, as described in \cref{sec:impl_updates}.
The superscript $\mathrm{joint}$ denotes full-clip predictions, with $D_q^{\mathrm{joint}}=D_p^{\mathrm{joint}}+R_\phi^{\mathrm{joint}}$.
The flow-matching objective is
\begin{equation}
 \mathcal L_q
 =\mathbb E\left[
   \left\|\frac{Y_t-D_q^{\mathrm{joint}}(Y_t,t,c)}{\sigma_t}
        -(\eta_Y-Y_0)\right\|_{\mathrm{av}}^2\right],
 \label{eq:mf_critic_loss}
\end{equation}
where the expectation covers prompts, on-policy joint clips, noise levels, and corruption, and $\|\cdot\|_{\mathrm{av}}^2$ averages over joint-clip latent coordinates.
Only $\phi$ is updated.
The teacher predictions and generated clips remain detached.

\paragraph{Flow matching is denoising regression.}
For the same noisy joint clip,
\begin{equation}
 Y_t=(1-\sigma_t)Y_0+\sigma_t\eta_Y,
 \qquad
 \frac{Y_t-D_q^{\mathrm{joint}}}{\sigma_t}
       -(\eta_Y-Y_0)
 =\frac{Y_0-D_q^{\mathrm{joint}}}{\sigma_t}.
 \label{eq:mf_app_flow_error}
\end{equation}
The first fraction on the left is the predicted rectified-flow velocity expressed through the clean-latent prediction, and the target velocity is $\eta_Y-Y_0$.
Squaring \cref{eq:mf_app_flow_error} shows that the critic objective in \cref{eq:mf_critic_loss} equals
\begin{equation}
 \mathcal L_q
 =\mathbb E\!\left[\frac1{\sigma_t^2}
        \|D_q^{\mathrm{joint}}(Y_t,t,c)-Y_0\|_{\mathrm{av}}^2\right].
 \label{eq:mf_app_critic_x0}
\end{equation}
The noise-dependent factor is retained, so this is not unweighted clean-latent MSE across timesteps.

\paragraph{What the residual estimator learns.}
Substituting $D_q^{\mathrm{joint}}=D_p^{\mathrm{joint}}+R_\phi^{\mathrm{joint}}$ gives
\begin{equation}
 \mathcal L_q
 =\mathbb E\!\left[\frac1{\sigma_t^2}
       \|R_\phi^{\mathrm{joint}}(Y_t,t,c)
                -(Y_0-D_p^{\mathrm{joint}}(Y_t,t,c))\|_{\mathrm{av}}^2\right].
 \label{eq:mf_app_residual_regression}
\end{equation}
At fixed $(Y_t,t,c)$, the teacher prediction is a deterministic frozen offset and $1/\sigma_t^2$ is positive and constant with respect to the clean target.
The conditional squared-error derivative is proportional to the residual prediction minus the conditional mean of $Y_0-D_p^{\mathrm{joint}}$.
Setting this derivative to zero under the student joint-clip law gives
\begin{align}
 R^{\mathrm{joint},*}(Y_t,t,c)
 &=\mathbb E_q[Y_0\mid Y_t,t,c]-D_p^{\mathrm{joint}}(Y_t,t,c), \notag\\
 D_p^{\mathrm{joint}}(Y_t,t,c)+R^{\mathrm{joint},*}(Y_t,t,c)
 &=\mathbb E_q[Y_0\mid Y_t,t,c].
 \label{eq:mf_app_optimal_residual}
\end{align}
This is a function-space optimum, not a guarantee that a finite residual network attains it.
It explains why generated student clips suffice to train the residual: the teacher information enters through the known denoising offset, while the regression target distribution remains the student's.
The critic loss is not a forward-KL generator objective, and its samples are not reweighted by $h_k$.
Only $\phi$ is updated in this step.

\section{Implementation Details}
\label{sec:implementation_details}

This section specifies the models, sampling procedure, mass-activation rules, and optimizer updates used for Mass Forcing, and reports its training cost.
\Cref{tab:hyperparameters} lists all numerical settings, which the text refers to by symbol.
All step counts refer to generator optimizer updates unless otherwise stated.

\begin{table}[!htbp]
\caption{Hyperparameters of Mass Forcing. Step counts refer to generator updates.}
\label{tab:hyperparameters}
\centering
\footnotesize
\hypersetup{hidelinks}
\setlength{\tabcolsep}{1.9pt}
\setlength{\heavyrulewidth}{0.75pt}
\setlength{\lightrulewidth}{0.35pt}
\renewcommand{\arraystretch}{1.12}
\newcommand{\detailMissing}{\makebox[2.8em]{\textcolor{black!45}{--}}}
\newcommand{\appTableNote}[1]{\par\vspace{2pt}\parbox{\wd\appTableBox}{\raggedright\scriptsize #1\par}}
\newcommand{\detailHeader}[1]{{\scriptsize\begin{tabular}[c]{@{}>{\bfseries}c@{}}#1\end{tabular}}}
 \setlength{\tabcolsep}{8pt}
\begin{tabular}{@{}lll@{}}
\toprule
\textbf{Symbol} & \textbf{Description} & \textbf{Value} \\
\midrule
\multicolumn{3}{@{}l}{\textit{Rollouts and score queries}} \\
$B$ & Chunks per training video & 7 \\
$F$ & Latent frames per chunk & 3 \\
$K$ & Rollouts per shared prefix & 4 \\
-- & Video latent shape (time, channel, height, width) & $21\times16\times60\times104$ \\
-- & Additional noise on rollout history & 0 \\
-- & Student denoising timesteps (four-step sampler) & $\{1000,750,500,250\}$ \\
$T$ & Number of diffusion timesteps & 1000 \\
$s_{\mathrm{shift}}$ & Timestep shift for score queries & 5 \\
$[t_{\min},t_{\max}]$ & Score-query timestep range & $[20,980]$ \\
-- & Score-query microbatch size & 2 \\
\midrule
\multicolumn{3}{@{}l}{\textit{Mass estimation and activation}} \\
$\tau$ & Log-ratio temperature & 2 \\
$\kappa$ & Logit clipping bound & 2 \\
$\lambda_{\max}$ & Maximum mixture coefficient & 0.25 \\
$N_{\mathrm{ramp}}$ & Ramp length of $\lambda_{\mathrm{eff}}$ & 25 \\
$[E_{\mathrm{lo}},E_{\mathrm{hi}}]$ & ESS gate thresholds (fraction of $K$) & $[0.35,0.60]$ \\
$[\sigma_{\mathrm{lo}},\sigma_{\mathrm{hi}}]$ & Noise-level range of active cells & $[0.20,0.65]$ \\
\midrule
\multicolumn{3}{@{}l}{\textit{Objectives}} \\
$w_{\mathrm{cfg}}$ & Classifier-free guidance scale & 3.0 \\
$\gamma$ & Set-level (MMD) loss weight & 0.05 \\
$\epsilon_n$ & Floor of the DMD normalizer $n_k$ & $10^{-8}$ \\
$\beta_{\min}$ & Floor of the MMD bandwidth $\beta$ & $10^{-6}$ \\
\midrule
\multicolumn{3}{@{}l}{\textit{Optimization}} \\
$N_{\mathrm{train}}$ & Generator updates & 1000 \\
$N_{\mathrm{critic}}$ & Critic updates per generator update & 5 \\
$N_{\mathcal B}$ & Critic replay-buffer capacity (entries) & 8 \\
$N_{\mathrm{ckpt}}$ & Checkpoint interval & 100 \\
-- & Optimizer (generator and critic) & AdamW \\
-- & AdamW momentum coefficients & $(0,\,0.999)$ \\
-- & Weight decay & 0.01 \\
-- & Learning rate: generator / critic & $2\times10^{-6}$ / $4\times10^{-7}$ \\
-- & Batch size: per rank / global & 2 / 16 \\
-- & Gradient accumulation steps & 1 \\
-- & Generator EMA decay & 0.99 \\
-- & Generator EMA start update & 40 \\
-- & Generator EMA update interval & 10 \\
-- & Seed: training / inference & 3407 / 0 \\
\bottomrule
\end{tabular}
\end{table}
 
\subsection{Models and On-Policy Rollout Sampling}
\label{sec:impl_sampling}

We use a causal Wan2.1-T2V-1.3B generator initialized from the Self Forcing ODE checkpoint, a frozen bidirectional Wan2.1-T2V-14B teacher, and a bidirectional 1.3B residual estimator~\citep{wan_2025_wan,huang_2026_nips_self_forcing}.
The fake denoiser is parameterized as $D_q=D_p+R_\phi$ and initialized from the calibrated residual checkpoint before DMD training.
The teacher and text encoder remain frozen.

\paragraph{Prefix and continuation sampling.}
Each training video consists of $B$ chunks of $F$ latent frames, including the first chunk, with no separate first-frame block.
For each batch, we sample $b$ uniformly from $\{1,\ldots,B\}$ and synchronize it across workers.
We autoregressively generate the first $b-1$ chunks with the current student and detach the resulting prefix and KV cache.
No additional context noise is applied during rollout.
For $b=1$, the prefix is empty.
From this shared prefix, we sample $K$ alternative realizations of chunk $b$ with independently sampled IID Gaussian noise for initialization and inter-step re-noising, recording the randomness required for deterministic replay.
The score networks receive joint clips of $L=Fb$ latent frames, but only the current chunk is optimized in the generator update.
The student retains the four-step Self Forcing sampler with denoising-timestep warping enabled.

\paragraph{Score-query timesteps.}
DMD and critic updates retain the original full-range score-query schedule rather than sampling only inside the ratio band.
The inherited sampler draws $t_{\mathrm{raw}}$ uniformly from $\{0,\ldots,T-1\}$, applies the timestep shift $t=s_{\mathrm{shift}}t_{\mathrm{raw}}/\bigl(1+(s_{\mathrm{shift}}-1)t_{\mathrm{raw}}/T\bigr)$, and clamps $t$ to $[t_{\min},t_{\max}]$.
The corruption level $\sigma_t$ is obtained from the scheduler at the resulting timestep.
The ratio queries and set-level loss reuse the DMD timestep and endpoint corruption.
Critic training uses the full-schedule denoising examples stored in the replay buffer, as detailed in \cref{sec:impl_updates}.

\subsection{Mass Estimation and Activation}
\label{sec:impl_mass}

The ratio estimator uses the pure conditional teacher, shared prefix/current corruption across the $K$ rollouts, a medoid anchor, and three-point Simpson quadrature.
We center the log-ratio estimates before temperature scaling and clipping, using the temperature $\tau$ and logit bound $\kappa$ in \cref{eq:mf_ratio_logits}.
CFG, with scale $w_{\mathrm{cfg}}$, is used only in the generator correction and does not enter ratio estimation.
Endpoint predictions are reused for both the ratio and local DMD correction.
Under the residual parameterization, additional integration midpoints require only residual-network evaluations because the conditional teacher cancels in the score difference.

\paragraph{Effective mixture coefficient.}
We compute effective sample size from the normalized ratios before mixing them with the uniform weights:
\begin{equation}
 E=\frac{(\sum_k\widehat r_k)^2}{K\sum_k\widehat r_k^2},
 \qquad
 \chi_{\mathrm{ESS}}=\operatorname{clip}\!\left(
       \frac{E-E_{\mathrm{lo}}}{E_{\mathrm{hi}}-E_{\mathrm{lo}}},0,1\right).
 \label{eq:impl_ess_gate}
\end{equation}
Here $E=\mathrm{ESS}/K$, so the ESS attenuation $\chi_{\mathrm{ESS}}$ is zero for $E\le E_{\mathrm{lo}}$ and one for $E\ge E_{\mathrm{hi}}$.
Let $N_G$ denote the generator-update index, $\chi_{\mathrm{cal}}\in\{0,1\}$ indicate a valid fake-calibration certificate, and $\chi_{\mathrm{cell}}(\sigma_t,L,b)\in\{0,1\}$ select prevalidated noise-level and clip-length cells within $\sigma_t\in[\sigma_{\mathrm{lo}},\sigma_{\mathrm{hi}}]$.
The default schedule is
\begin{equation}
 \lambda_{\mathrm{eff}}
 =\lambda_{\max}\min\!\left(\frac{N_G}{N_{\mathrm{ramp}}},1\right)
    \chi_{\mathrm{cal}}\,
    \chi_{\mathrm{cell}}(\sigma_t,L,b)\,
    \chi_{\mathrm{ESS}}.
 \label{eq:impl_lambda_schedule}
\end{equation}
Thus the nominal ratio contribution ramps linearly from zero to $\lambda_{\max}$ over the first $N_{\mathrm{ramp}}$ generator updates.
Critic-only initialization steps do not advance this counter.
The particle-spread gate is disabled in this configuration.
All gate outputs and weights are detached, and normalization is performed separately for each prompt--prefix group.

\paragraph{Calibration certificate.}
The certificate $\chi_{\mathrm{cal}}$ is issued once, before DMD training.
We fit the residual $R_\phi$ to rollouts of the ODE-initialized student for 250 updates on 32 prompts.
The certificate is valid if two conditions hold on 8 prompts that are not used for fitting.
First, the clean-latent prediction error of $D_q$ does not regress: its median stays within $5\%$ of its value before fitting, and its spread across the $B=7$ clip lengths satisfies $(\max-\min)/\mathrm{median}\le 1$.
Second, the error along the interpolation paths used for integration is at most $4\times$ the error at the path endpoints.
The certificate is valid for the reported run.

\paragraph{Activation cells.}
We divide $[\sigma_{\mathrm{lo}},\sigma_{\mathrm{hi}}]$ into 6 noise-level bins and pair each bin with one of the $B=7$ chunk indices $b$ (equivalently, clip lengths $L=Fb$), which gives 42 cells.
A cell is active if it passes two tests on 32 probe prompts.
First, the teacher must use the conditioning in that cell. Its clean-latent prediction error must drop when shuffled conditioning is replaced by the true conditioning, and the lower end of the $95\%$ bootstrap interval of this drop must be above zero.
For $b=1$ we shuffle the prompt, and for $b>1$ we shuffle the noisy prefix.
Second, the fake denoiser must be accurate in that cell. The medians of its endpoint error and of its path error must both be within $4\times$ the corresponding medians over all cells.
37 of the 42 cells pass.
The certificate and the cell mask are fixed before DMD training and are not tuned on evaluation results.

\paragraph{Re-check during training.}
As a post-hoc diagnostic, we repeat the cell tests with the student and the fake denoiser at updates 500 and 1{,}000.
At update 500, 39 cells pass, namely the original 37 and two more.
At update 1{,}000, the same 37 cells pass.
These results are not used to change the cell mask during training.

Outside active cells, $\lambda_{\mathrm{eff}}=0$ and $h_k=1$.
The local correction then reduces to unweighted conditional DMD, and the mixed-reference MMD loss and gradient vanish under identical replay.
These activation rules do not restrict the full noise schedule used for DMD or critic learning.

\subsection{Replay, Optimization, and Evaluation}
\label{sec:impl_updates}

\paragraph{Generator update.}
We collect detached reference rollouts, compute their mass weights and score corrections, and then replay the same rollouts with identical randomness before changing the generator parameters.
The local correction uses the mean absolute clean-latent error normalizer in \cref{eq:mf_app_dmd_normalizer} with numerical floor $\epsilon_n$.
The MMD source is uniformly weighted, while its fixed reference uses $\pi_k=\operatorname{sg}(h_k/K)$.
We use the complete empirical RBF-MMD, including source--source interactions and diagonal terms, with mean squared latent distances and the detached median bandwidth in \cref{sec:app_set}, floored at $\beta_{\min}$.
Gradient accumulation preserves the $1/(Kd)$ local-loss reduction and the full MMD source derivative, including both occurrences of each sample in the source--source term.

\paragraph{Critic update and optimizer settings.}
Each generator optimizer step is followed by $N_{\mathrm{critic}}$ critic updates using detached joint student clips from a replay buffer $\mathcal B$ holding up to $N_{\mathcal B}$ entries.
Each replay entry retains the clean and noisy joint clips, corruption noise, timestep, and conditional teacher prediction, so critic replay does not require a new generator rollout or teacher forward pass.
The critic uses the flow-matching objective in \cref{eq:mf_critic_loss} without ratio reweighting.
Both networks are optimized with AdamW.
Learning rates, momentum coefficients, weight decay, batch sizes, and the training seed are listed in \cref{tab:hyperparameters}.
We retain the baseline's mixed-precision training and gradient checkpointing, and update only the retained generator replay graph rather than backpropagating through the detached prefix.

\paragraph{Training duration and inference.}
Training runs for $N_{\mathrm{train}}$ generator updates, and hence $N_{\mathrm{critic}}N_{\mathrm{train}}$ critic updates, saving a checkpoint every $N_{\mathrm{ckpt}}$ generator updates.
We maintain an exponential moving average (EMA) of the generator weights and use it for evaluation.
MovieGen-100 validation generates full $BF$-frame latent videos with a fixed inference seed.
At inference, only the four-step causal generator is deployed, with IID Gaussian sampling and no teacher, fake denoiser, ratio estimation, or MMD computation.

\clearpage
\subsection{Training Algorithm}
\label{sec:impl_algorithm}

\Cref{alg:mass_forcing_update} summarizes one training iteration, which is repeated for $N_{\mathrm{train}}$ generator updates.
It keeps only the steps that define the method.
The gates that set $\lambda_{\mathrm{eff}}$ are specified in \cref{sec:impl_mass}, the replay buffer, the EMA, and the optimizers in \cref{sec:impl_updates}, and the kernel bandwidth $\beta$ in \cref{sec:app_set}.
All hyperparameters are listed in \cref{tab:hyperparameters}.

\begin{algorithm}[H]
\caption{One training iteration of Mass Forcing}
\label{alg:mass_forcing_update}
\small
\begin{algorithmic}[1]
\Require generator $G_\theta$, frozen teacher $D_p$, fake denoiser $D_q$
\Statex \textbf{On-policy rollouts} (no gradient)
\State Sample a prompt $c$ and a chunk index $b$; generate the prefix $u$ with $G_\theta$ and detach it
\State Draw $K$ continuations $\bar x_k=G_\theta(\epsilon_k;u,c)$ and add one shared noise draw, $\bar y_k=\alpha_t\bar x_k+\sigma_t\eta$
\Statex \textbf{Teacher-relative mass} (\cref{sec:method_ratio})
\State Query both denoisers at the noisy prefix and set $\Delta\gets\alpha_t(D_p-D_q)/\sigma_t^2$
\State $\ell_k\gets$ three-point Simpson integral of $\Delta$ from the medoid anchor $a$ to $\bar y_k$ \Comment{\cref{eq:mf_simpson}}
\State $v_k\gets\operatorname{clip}\bigl((\ell_k-\bar\ell)/\tau,-\kappa,\kappa\bigr)$, \quad $\widehat r_k\gets K\exp(v_k)/\sum_j\exp(v_j)$
\State $h_k\gets1-\lambda_{\mathrm{eff}}+\lambda_{\mathrm{eff}}\,\widehat r_k$ \Comment{$\lambda_{\mathrm{eff}}=0$ outside active cells}
\Statex \textbf{Generator update} (weights and references fixed)
\State $g_k\gets\bigl[h_k(D_{q,k}-D_{p,k})+(D_{p,k}-D_{\mathrm{cfg},k})\bigr]/n_k$
\State Regenerate $x_k=G_\theta(\epsilon_k;u,c)$ with gradients and set $y_k=\alpha_tx_k+\sigma_t\eta$
\State $\mathcal L_{\mathrm{local}}\gets\frac1{2Kd}\sum_k\|x_k-\operatorname{sg}(x_k-g_k)\|_2^2$
\State $\mathcal L_{\mathrm{set}}\gets\operatorname{MMD}_{\mathcal K_\beta}^2\bigl(\frac1K\sum_k\delta_{y_k},\,\sum_k\frac{h_k}{K}\delta_{\bar y_k}\bigr)$
\State Update $\theta$ with $\nabla_\theta(\mathcal L_{\mathrm{local}}+\gamma\mathcal L_{\mathrm{set}})$
\Statex \textbf{Fake-denoiser updates}
\State Repeat $N_{\mathrm{critic}}$ times: update $D_q$ with the denoising loss $\mathcal L_q$ on stored, unweighted student clips \Comment{\cref{eq:mf_critic_loss}}
\end{algorithmic}
\end{algorithm}

\subsection{Training Cost}
\label{sec:appendix_cost}

Mass Forcing adds computation during training only.
Each generator update costs 2.6 times the floating-point operations of a Self Forcing update.
Almost all of this comes from the conditional-chunk data path, which samples $K=4$ continuations from a shared prefix in place of one holistic rollout.

The mass estimation and the set-level term together account for about 1 percent over the uniform conditional-chunk baseline of \cref{sec:appendix_matched_baseline}.
The line integral needs one additional forward pass of the 1.3B residual network at the Simpson midpoint, and no additional query to the 14B teacher.
The set-level term reuses rollouts that have already been generated.

A full run of 1{,}000 generator updates takes about 300 GPU-hours.
Inference is unchanged, so the throughput and latency in \cref{tab:main-table} are those of the four-step generator alone.

\section{Detailed Experimental Results}
\label{sec:appendix_dimensions}

\Crefrange{tab:moviegen-quality-dimensions}{tab:vbench-semantic-dimensions} report dimension-level results using the VBench evaluation taxonomy~\citep{huang_2024_cvpr_vbench}.
For the main comparisons, the methods, ordering, and model groups follow \cref{tab:main-table}.
Dimension-level scores are raw evaluator outputs multiplied by 100, whereas aggregate scores use the normalized scale.
Higher values indicate better performance, and dashes mark unavailable results.
\Cref{sec:appendix_quality_trajectory,sec:appendix_matched_baseline} then report visual quality during training and the matched unweighted baseline.

\subsection{MovieGen}
\label{sec:appendix_moviegen}

MovieGen provides text prompts but not the metadata required by VBench's semantic evaluators.
We construct this metadata using Qwen3-VL-32B-Instruct.
The annotation prompt and validation procedure are provided in \cref{sec:moviegen_metadata_prompt}.
No prompts in the evaluated MovieGen set were identified as satisfying VBench's criteria for Spatial Relationship, so we exclude this dimension rather than assigning it a zero score.
For the remaining eight dimensions, we retain VBench's standard per-dimension normalization and relative weights, then renormalize these weights to sum to one.
Since VBench assigns equal weight to each semantic dimension, the resulting Semantic score is the mean of the eight normalized scores.
All compared methods and ablation variants use the same metadata, dimension set, and aggregation procedure, ensuring a consistent within-benchmark comparison.
Quality aggregation is unchanged, and the Total score remains $0.8$ Quality plus $0.2$ Semantic.

\begin{table}[!htbp]
\caption{Detailed quality evaluation on MovieGen prompts. Scores are shown as percentages ($\uparrow$).}
\label{tab:moviegen-quality-dimensions}
\centering
\footnotesize
\hypersetup{hidelinks}
\setlength{\tabcolsep}{1.9pt}
\setlength{\heavyrulewidth}{0.75pt}
\setlength{\lightrulewidth}{0.35pt}
\renewcommand{\arraystretch}{1.12}
\newcommand{\detailMissing}{\makebox[2.8em]{\textcolor{black!45}{--}}}
\newcommand{\appTableNote}[1]{\par\vspace{2pt}\parbox{\wd\appTableBox}{\raggedright\scriptsize #1\par}}
\newcommand{\detailHeader}[1]{{\scriptsize\begin{tabular}[c]{@{}>{\bfseries}c@{}}#1\end{tabular}}}
 \setlength{\tabcolsep}{5pt}
\begin{tabular}{@{}l*{7}{c}@{\hspace{7pt}}c@{}}
\toprule
\textbf{Model} & \detailHeader{Subject\\Consistency} & \detailHeader{Background\\Consistency} & \detailHeader{Temporal\\Flickering} & \detailHeader{Motion\\Smoothness} & \detailHeader{Dynamic\\Degree} & \detailHeader{Aesthetic\\Quality} & \detailHeader{Imaging\\Quality} & \detailHeader{Quality\\Score} \\
\midrule
\rowcolor{black!4}
\multicolumn{9}{l}{\color{black!75}\textit{Bidirectional Video Diffusion Models}} \\
\addlinespace[1pt]
LTX-Video & 97.30 & 96.67 & 98.47 & 99.35 & 31.90 & 62.75 & 70.36 & 82.44 \\
Wan2.1 & 94.29 & 96.12 & 97.16 & 98.31 & 59.02 & 65.15 & 67.61 & 82.73 \\
AVD & 94.20 & 94.13 & 95.83 & 98.32 & 61.11 & 62.91 & 67.41 & 81.54 \\
\addlinespace[1pt]
\rowcolor{black!4}
\multicolumn{9}{l}{\color{black!75}\textit{Autoregressive Video Diffusion Models}} \\
\addlinespace[1pt]
NOVA & 92.24 & 93.99 & 98.26 & 99.01 & 50.45 & 56.05 & 52.57 & 78.36 \\
Pyramid Flow & 94.20 & 95.33 & 98.56 & 99.30 & 34.40 & 64.09 & 64.24 & 81.07 \\
SkyReels-V2 & 96.23 & 96.94 & 97.71 & 98.77 & 50.25 & 64.86 & 64.43 & 82.51 \\
MAGI-1 & 97.80 & 97.66 & 99.01 & 99.45 & 22.13 & 65.19 & 64.93 & 81.81 \\
\addlinespace[1pt]
\rowcolor{black!4}
\multicolumn{9}{l}{\color{black!75}\textit{Distilled Autoregressive Video Models}} \\
\addlinespace[1pt]
CausVid & 94.74 & 94.29 & 96.88 & 98.06 & 57.33 & 62.79 & 67.81 & 81.72 \\
Self Forcing & 94.17 & 94.85 & 96.75 & 98.32 & 63.41 & 63.88 & 69.84 & 82.77 \\
Causal Forcing & 94.37 & 94.92 & 95.52 & 97.64 & 77.57 & 64.42 & 70.87 & 83.28 \\
Causal-rCM & 95.93 & 95.88 & 97.55 & 98.64 & 48.26 & 65.35 & 69.50 & 82.80 \\
\midrule
\textbf{Mass Forcing (Ours)} & 95.10 & 95.02 & 95.94 & 98.41 & 83.35 & 65.92 & 69.87 & 84.53 \\
\bottomrule
\end{tabular}
\end{table}
\begin{table}[!htbp]
\caption{Detailed semantic evaluation on MovieGen prompts. Scores are shown as percentages ($\uparrow$).}
\label{tab:moviegen-semantic-dimensions}
\centering
\footnotesize
\hypersetup{hidelinks}
\setlength{\tabcolsep}{1.9pt}
\setlength{\heavyrulewidth}{0.75pt}
\setlength{\lightrulewidth}{0.35pt}
\renewcommand{\arraystretch}{1.12}
\newcommand{\detailMissing}{\makebox[2.8em]{\textcolor{black!45}{--}}}
\newcommand{\appTableNote}[1]{\par\vspace{2pt}\parbox{\wd\appTableBox}{\raggedright\scriptsize #1\par}}
\newcommand{\detailHeader}[1]{{\scriptsize\begin{tabular}[c]{@{}>{\bfseries}c@{}}#1\end{tabular}}}
 \setlength{\tabcolsep}{4pt}
\begin{tabular}{@{}l*{8}{c}@{\hspace{7pt}}c@{}}
\toprule
\textbf{Model} & \detailHeader{Object\\Class} & \detailHeader{Multiple\\Objects} & \detailHeader{Human\\Action} & \detailHeader{Color} & \detailHeader{Scene} & \detailHeader{Appearance\\Style} & \detailHeader{Temporal\\Style} & \detailHeader{Overall\\Consistency} & \detailHeader{Semantic\\Score} \\
\midrule
\rowcolor{black!4}
\multicolumn{10}{l}{\color{black!75}\textit{Bidirectional Video Diffusion Models}} \\
\addlinespace[1pt]
LTX-Video & 83.75 & 55.77 & 76.92 & 84.68 & 60.66 & 21.58 & 23.16 & 23.15 & 72.21 \\
Wan2.1 & 84.20 & 51.92 & 100.0 & 93.43 & 58.48 & 22.13 & 22.44 & 26.02 & 76.41 \\
AVD & 84.43 & 56.25 & 100.0 & 90.10 & 47.37 & 22.07 & 23.14 & 24.86 & 74.69 \\
\addlinespace[1pt]
\rowcolor{black!4}
\multicolumn{10}{l}{\color{black!75}\textit{Autoregressive Video Diffusion Models}} \\
\addlinespace[1pt]
NOVA & 82.39 & 59.13 & 84.62 & 83.34 & 50.24 & 23.06 & 23.15 & 25.57 & 73.14 \\
Pyramid Flow & 89.03 & 54.81 & 100.0 & 87.54 & 60.24 & 22.64 & 20.67 & 25.20 & 76.24 \\
SkyReels-V2 & 82.61 & 54.33 & 92.31 & 91.67 & 52.96 & 21.88 & 20.96 & 24.47 & 73.33 \\
MAGI-1 & 87.50 & 65.87 & 100.0 & 87.79 & 58.09 & 23.03 & 22.34 & 25.41 & 77.95 \\
\addlinespace[1pt]
\rowcolor{black!4}
\multicolumn{10}{l}{\color{black!75}\textit{Distilled Autoregressive Video Models}} \\
\addlinespace[1pt]
CausVid & 85.57 & 75.96 & 92.31 & 87.69 & 51.27 & 22.00 & 20.77 & 24.40 & 75.62 \\
Self Forcing & 86.36 & 65.87 & 92.31 & 92.50 & 60.54 & 21.87 & 23.28 & 25.66 & 77.70 \\
Causal Forcing & 88.69 & 71.63 & 100.0 & 95.63 & 59.78 & 21.97 & 21.79 & 25.45 & 79.42 \\
Causal-rCM & 89.55 & 61.54 & 92.31 & 93.25 & 62.71 & 21.67 & 21.61 & 25.31 & 77.21 \\
\midrule
\textbf{Mass Forcing (Ours)} & 85.63 & 67.79 & 92.31 & 96.39 & 63.53 & 21.98 & 22.97 & 25.79 & 78.78 \\
\bottomrule
\end{tabular}
\end{table}
 
\clearpage
\subsubsection{Semantic-Metadata Annotation Prompt}
\label{sec:moviegen_metadata_prompt}

For each MovieGen record, the annotation model receives the expanded English prompt used for video generation as its sole semantic ground truth, without rewriting it at annotation time.
We reproduce the system instruction and per-record user message below.
The official vocabularies of 79 VBench object classes, 86 VBench scene labels, and 400 Kinetics-400 action labels are appended to the system instruction at runtime.
Responses are parsed as JSON and checked against these vocabularies and the prompt text.
Unsupported or ambiguous fields are discarded before conversion to VBench evaluator queries.

\paragraph{System instruction.}
\begin{Verbatim}[fontsize=\footnotesize,breaklines=true,breaksymbolleft={},breaksymbolright={},frame=single,framesep=4pt]
You are a strict VBench semantic-metadata annotator. You receive one already-expanded English prompt that will be used verbatim to generate a video. Treat that expanded prompt as the only ground truth. Extract only semantic targets explicitly supported by its text.

Return exactly one JSON object and no markdown. Required keys:
{
  "primary_semantic_dimension": one of ["object_class", "multiple_objects", "human_action", "color", "spatial_relationship", "scene", "appearance_style", "temporal_style", "overall_consistency"],
  "primary_feature": a concise English feature string directly supported by the prompt,
  "confidence": a number in [0, 1],
  "semantic_metadata": {
    "object_class": {"object": "single common detector class noun"} or null,
    "multiple_objects": {"object_a": "common detector class noun", "object_b": "different common detector class noun"} or null,
    "human_action": {"kinetics_400_label": "exact label from the provided allowlist"} or null,
    "color": {"object": "common detector class noun", "color": "one allowed color"} or null,
    "spatial_relationship": {"object_a": "common detector class noun", "object_b": "different common detector class noun", "relationship": "one allowed relationship"} or null,
    "scene": {"scene": "short literal setting noun phrase"} or null,
    "appearance_style": {"appearance_style": "explicit visual style phrase"} or null,
    "temporal_style": {"temporal_style": "explicit temporal style phrase"} or null
  },
  "rationale": one short sentence.
}

Rules:
1. Always choose one primary_semantic_dimension. Use overall_consistency when no narrower semantic dimension is reliable. overall_consistency is implicit and must not appear inside semantic_metadata.
2. Do not infer or embellish anything beyond the supplied expanded prompt. Use null whenever a field is absent, ambiguous, merely implied, or cannot satisfy the exact schema.
3. object_class must be one exact label from the supplied official VBench object-class allowlist and must denote a visibly present main subject. Map men/women/children to "person", planes/jets to "airplane", and other subclasses only when the mapping is unambiguous. Do not use a label absent from the allowlist.
4. multiple_objects requires two different labels from that same object-class allowlist, both explicitly visible. Do not use two instances of one class, scenery, an action, or an abstract concept.
5. color is valid only when the prompt explicitly binds exactly one of [black, blue, gray, green, grey, orange, pink, purple, red, silver, white, yellow] to an object-class allowlist label. Return that label separately.
6. spatial_relationship requires two different object-class allowlist labels and an explicit relation exactly equal to one of [on the left of, on the right of, on the top of, on the bottom of]. Do not translate inside/behind/beside/through/above/below/reflecting/over/under into one of these relations.
7. scene must be one exact label from the supplied official VBench scene allowlist and must be explicitly supported by the prompt. Do not return an inferred setting or a longer description.
8. appearance_style is valid only for an explicit rendering or visual style, such as photorealistic, watercolor, anime, oil painting, stop-motion, or cinematic. Mood, clothing, image quality, camera angle, lighting, and colorfulness alone are not appearance styles.
9. temporal_style is valid only for an explicit temporal presentation such as time-lapse, slow motion, stop motion, reverse motion, looping, or step-printing. Ordinary actions and camera motion are not temporal styles.
10. human_action is valid only when every content word of one complete Kinetics-400 label appears in order in the supplied prompt (articles may differ) and describes a visible person. Do not choose a merely similar label: ordinary walking is not "walking the dog", and ordinary drinking is not "tasting beer". Copy the exact lowercase allowlist label or return null.
11. Keep all values in English. Never include a metadata key not listed in the schema.
\end{Verbatim}

\paragraph{Per-record user message.}
The placeholder \texttt{\{EXTENDED\_PROMPT\}} is replaced verbatim with the corresponding English generation prompt.
\begin{Verbatim}[fontsize=\footnotesize,breaklines=true,breaksymbolleft={},breaksymbolright={},frame=single,framesep=4pt]
Expanded MovieGen generation prompt (the sole metadata ground truth):
{EXTENDED_PROMPT}

Extract VBench semantic metadata now and return exactly the required JSON object.
\end{Verbatim}

\subsection{VBench}
\label{sec:appendix_vbench}

\begin{table}[!htbp]
\caption{Detailed quality evaluation on VBench prompts. Scores are shown as percentages ($\uparrow$).}
\label{tab:vbench-quality-dimensions}
\centering
\footnotesize
\hypersetup{hidelinks}
\setlength{\tabcolsep}{1.9pt}
\setlength{\heavyrulewidth}{0.75pt}
\setlength{\lightrulewidth}{0.35pt}
\renewcommand{\arraystretch}{1.12}
\newcommand{\detailMissing}{\makebox[2.8em]{\textcolor{black!45}{--}}}
\newcommand{\appTableNote}[1]{\par\vspace{2pt}\parbox{\wd\appTableBox}{\raggedright\scriptsize #1\par}}
\newcommand{\detailHeader}[1]{{\scriptsize\begin{tabular}[c]{@{}>{\bfseries}c@{}}#1\end{tabular}}}
 \setlength{\tabcolsep}{5pt}
\begin{tabular}{@{}l*{7}{c}@{\hspace{7pt}}c@{}}
\toprule
\textbf{Model} & \detailHeader{Subject\\Consistency} & \detailHeader{Background\\Consistency} & \detailHeader{Temporal\\Flickering} & \detailHeader{Motion\\Smoothness} & \detailHeader{Dynamic\\Degree} & \detailHeader{Aesthetic\\Quality} & \detailHeader{Imaging\\Quality} & \detailHeader{Quality\\Score} \\
\midrule
\rowcolor{black!4}
\multicolumn{9}{l}{\color{black!75}\textit{Bidirectional Video Diffusion Models}} \\
\addlinespace[1pt]
LTX-Video & 96.46 & 97.17 & 99.45 & 99.33 & 43.06 & 65.58 & 68.34 & 83.78 \\
Wan2.1 & 96.34 & 98.03 & 99.29 & 98.43 & 62.50 & 68.79 & 67.02 & 85.17 \\
AVD & 96.21 & 94.77 & 96.52 & 98.40 & 73.61 & 65.03 & 66.54 & 83.51 \\
\addlinespace[1pt]
\rowcolor{black!4}
\multicolumn{9}{l}{\color{black!75}\textit{Autoregressive Video Diffusion Models}} \\
\addlinespace[1pt]
NOVA & 94.25 & 95.44 & 99.18 & 99.23 & 44.44 & 59.43 & 55.01 & 79.96 \\
Pyramid Flow & 93.78 & 96.66 & 99.30 & 99.24 & 59.72 & 65.30 & 63.08 & 83.51 \\
SkyReels-V2 & 97.35 & 98.24 & 99.48 & 98.94 & 63.89 & 66.07 & 63.29 & 84.86 \\
MAGI-1 & 97.88 & 97.91 & 99.79 & 99.50 & 16.67 & 64.62 & 62.75 & 81.38 \\
\addlinespace[1pt]
\rowcolor{black!4}
\multicolumn{9}{l}{\color{black!75}\textit{Distilled Autoregressive Video Models}} \\
\addlinespace[1pt]
CausVid & 95.87 & 95.67 & 99.26 & 98.02 & 65.28 & 64.84 & 66.64 & 83.92 \\
Self Forcing & 96.15 & 95.89 & 98.98 & 98.62 & 61.11 & 65.90 & 69.53 & 84.51 \\
Causal Forcing & 95.01 & 95.46 & 97.95 & 97.59 & 86.11 & 66.49 & 69.28 & 85.22 \\
Causal-rCM & 95.95 & 96.55 & 99.24 & 98.36 & 69.44 & 66.24 & 68.38 & 85.10 \\
\midrule
\textbf{Mass Forcing (Ours)} & 97.56 & 96.17 & 97.18 & 98.97 & 88.89 & 67.08 & 66.93 & 86.17 \\
\bottomrule
\end{tabular}
\end{table}
\begin{table}[!htbp]
\caption{Detailed semantic evaluation on VBench prompts. Scores are shown as percentages ($\uparrow$).}
\label{tab:vbench-semantic-dimensions}
\centering
\footnotesize
\hypersetup{hidelinks}
\setlength{\tabcolsep}{1.9pt}
\setlength{\heavyrulewidth}{0.75pt}
\setlength{\lightrulewidth}{0.35pt}
\renewcommand{\arraystretch}{1.12}
\newcommand{\detailMissing}{\makebox[2.8em]{\textcolor{black!45}{--}}}
\newcommand{\appTableNote}[1]{\par\vspace{2pt}\parbox{\wd\appTableBox}{\raggedright\scriptsize #1\par}}
\newcommand{\detailHeader}[1]{{\scriptsize\begin{tabular}[c]{@{}>{\bfseries}c@{}}#1\end{tabular}}}
 \setlength{\tabcolsep}{3pt}
\begin{tabular}{@{}l*{9}{c}@{\hspace{7pt}}c@{}}
\toprule
\textbf{Model} & \detailHeader{Object\\Class} & \detailHeader{Multiple\\Objects} & \detailHeader{Human\\Action} & \detailHeader{Color} & \detailHeader{Spatial\\Relationship} & \detailHeader{Scene} & \detailHeader{Appearance\\Style} & \detailHeader{Temporal\\Style} & \detailHeader{Overall\\Consistency} & \detailHeader{Semantic\\Score} \\
\midrule
\rowcolor{black!4}
\multicolumn{11}{l}{\color{black!75}\textit{Bidirectional Video Diffusion Models}} \\
\addlinespace[1pt]
LTX-Video & 93.28 & 66.01 & 94.00 & 75.06 & 62.46 & 48.62 & 20.73 & 24.87 & 25.47 & 73.42 \\
Wan2.1 & 92.88 & 79.80 & 95.00 & 83.49 & 71.82 & 49.78 & 22.40 & 26.85 & 27.53 & 79.04 \\
AVD & 90.27 & 70.88 & 96.00 & 81.53 & 66.49 & 50.29 & 21.97 & 26.23 & 26.06 & 76.32 \\
\addlinespace[1pt]
\rowcolor{black!4}
\multicolumn{11}{l}{\color{black!75}\textit{Autoregressive Video Diffusion Models}} \\
\addlinespace[1pt]
NOVA & 88.05 & 70.88 & 95.00 & 77.00 & 65.94 & 48.76 & 23.44 & 26.47 & 26.94 & 76.11 \\
Pyramid Flow & 88.77 & 65.40 & 94.00 & 82.48 & 67.56 & 51.96 & 22.29 & 26.29 & 27.05 & 76.21 \\
SkyReels-V2 & 87.58 & 59.22 & 95.00 & 87.71 & 54.76 & 54.65 & 21.31 & 25.54 & 26.30 & 74.19 \\
MAGI-1 & 90.66 & 61.20 & 96.00 & 87.01 & 55.50 & 59.74 & 23.43 & 25.54 & 26.99 & 76.59 \\
\addlinespace[1pt]
\rowcolor{black!4}
\multicolumn{11}{l}{\color{black!75}\textit{Distilled Autoregressive Video Models}} \\
\addlinespace[1pt]
CausVid & 88.61 & 83.23 & 97.00 & 86.86 & 60.33 & 49.71 & 21.14 & 25.62 & 26.64 & 77.11 \\
Self Forcing & 93.04 & 80.03 & 97.00 & 84.21 & 68.68 & 55.23 & 21.72 & 26.54 & 27.51 & 79.40 \\
Causal Forcing & 92.64 & 85.82 & 95.00 & 86.81 & 74.61 & 50.15 & 21.55 & 26.57 & 27.01 & 79.83 \\
Causal-rCM & 93.91 & 83.92 & 94.00 & 82.85 & 74.83 & 47.24 & 21.67 & 26.40 & 27.03 & 78.84 \\
\midrule
\textbf{Mass Forcing (Ours)} & 93.20 & 83.08 & 99.00 & 88.68 & 71.42 & 53.27 & 21.62 & 25.94 & 27.30 & 80.23 \\
\bottomrule
\end{tabular}
\end{table}
 
\clearpage
\subsection{Ablation Studies}
\label{sec:appendix_ablation}

\Crefrange{tab:ablation-moviegen-quality-dimensions}{tab:ablation-vbench-semantic-dimensions} provide the dimension-level results for \cref{tab:ablation}, together with the Quality, Semantic, and Total scores.
The Ours rows repeat the reference results from the main comparison tables above.
The w/o CFG decoupling variant omits both CFG decoupling and MMD, whereas w/o MMD retains CFG decoupling but removes set-level supervision.
These variants form a progression from coupled local weighting, through decoupled local weighting, to the full local-and-set objective.
The gap between w/o CFG decoupling and Ours therefore includes both changes.
The MovieGen and VBench evaluations contain 1,003 and 946 videos per ablation variant, respectively.
MovieGen semantic results omit Spatial Relationship and aggregate the eight available dimensions, whereas VBench uses all nine semantic dimensions.

\subsubsection{MovieGen}
\label{sec:appendix_ablation_moviegen}

\begin{table}[!htbp]
\caption{Detailed quality ablation results on MovieGen prompts. Scores are shown as percentages ($\uparrow$).}
\label{tab:ablation-moviegen-quality-dimensions}
\centering
\footnotesize
\hypersetup{hidelinks}
\setlength{\tabcolsep}{1.9pt}
\setlength{\heavyrulewidth}{0.75pt}
\setlength{\lightrulewidth}{0.35pt}
\renewcommand{\arraystretch}{1.12}
\newcommand{\detailMissing}{\makebox[2.8em]{\textcolor{black!45}{--}}}
\newcommand{\appTableNote}[1]{\par\vspace{2pt}\parbox{\wd\appTableBox}{\raggedright\scriptsize #1\par}}
\newcommand{\detailHeader}[1]{{\scriptsize\begin{tabular}[c]{@{}>{\bfseries}c@{}}#1\end{tabular}}}
 \setlength{\tabcolsep}{5pt}
\begin{tabular}{@{}l*{7}{c}@{\hspace{7pt}}*{1}{c}@{}}
\toprule
\textbf{Variant} & \detailHeader{Subject\\Consistency} & \detailHeader{Background\\Consistency} & \detailHeader{Temporal\\Flickering} & \detailHeader{Motion\\Smoothness} & \detailHeader{Dynamic\\Degree} & \detailHeader{Aesthetic\\Quality} & \detailHeader{Imaging\\Quality} & \detailHeader{Quality\\Score} \\
\midrule
\textbf{Ours} & 95.10 & 95.02 & 95.94 & 98.41 & 83.35 & 65.92 & 69.87 & 84.53 \\
\midrule
w/o CFG decoupling & 95.90 & 95.53 & 96.82 & 98.53 & 61.81 & 65.91 & 69.05 & 83.42 \\
w/o MMD & 95.69 & 95.12 & 96.00 & 98.56 & 65.00 & 65.91 & 69.86 & 83.35 \\
Shuffled-ratio control & 95.72 & 95.18 & 96.35 & 98.72 & 59.62 & 66.16 & 69.88 & 83.22 \\
$K=2$ & 95.09 & 94.99 & 95.57 & 98.52 & 74.78 & 65.38 & 68.21 & 83.43 \\
Endpoint-only & 95.19 & 94.94 & 95.54 & 98.34 & 74.68 & 66.19 & 69.14 & 83.59 \\
$\lambda_{\max}=0.5$ & 95.24 & 94.94 & 96.34 & 98.45 & 64.31 & 65.73 & 69.78 & 83.22 \\
$\lambda_{\max}=1.0$ & 95.00 & 94.87 & 96.22 & 98.48 & 72.98 & 65.44 & 69.31 & 83.67 \\
\bottomrule
\end{tabular}
\end{table}
\begin{table}[!htbp]
\caption{Detailed semantic ablation results on MovieGen prompts. Scores are shown as percentages ($\uparrow$).}
\label{tab:ablation-moviegen-semantic-dimensions}
\centering
\footnotesize
\hypersetup{hidelinks}
\setlength{\tabcolsep}{1.9pt}
\setlength{\heavyrulewidth}{0.75pt}
\setlength{\lightrulewidth}{0.35pt}
\renewcommand{\arraystretch}{1.12}
\newcommand{\detailMissing}{\makebox[2.8em]{\textcolor{black!45}{--}}}
\newcommand{\appTableNote}[1]{\par\vspace{2pt}\parbox{\wd\appTableBox}{\raggedright\scriptsize #1\par}}
\newcommand{\detailHeader}[1]{{\scriptsize\begin{tabular}[c]{@{}>{\bfseries}c@{}}#1\end{tabular}}}
 \setlength{\tabcolsep}{3pt}
\begin{tabular}{@{}l*{8}{c}@{\hspace{7pt}}*{2}{c}@{}}
\toprule
\textbf{Variant} & \detailHeader{Object\\Class} & \detailHeader{Multiple\\Objects} & \detailHeader{Human\\Action} & \detailHeader{Color} & \detailHeader{Scene} & \detailHeader{Appearance\\Style} & \detailHeader{Temporal\\Style} & \detailHeader{Overall\\Consistency} & \detailHeader{Semantic\\Score} & \detailHeader{Total\\Score} \\
\midrule
\textbf{Ours} & 85.63 & 67.79 & 92.31 & 96.39 & 63.53 & 21.98 & 22.97 & 25.79 & 78.78 & 83.38 \\
\midrule
w/o CFG decoupling & 86.25 & 67.79 & 92.31 & 80.11 & 56.52 & 21.84 & 22.02 & 25.41 & 75.24 & 81.79 \\
w/o MMD & 86.76 & 72.60 & 92.31 & 96.59 & 58.09 & 21.90 & 22.81 & 25.67 & 78.59 & 82.40 \\
Shuffled-ratio control & 86.08 & 75.48 & 92.31 & 94.88 & 52.92 & 22.02 & 21.58 & 25.63 & 77.48 & 82.07 \\
$K=2$ & 86.82 & 75.96 & 100.00 & 86.53 & 53.53 & 22.16 & 22.32 & 25.45 & 77.90 & 82.32 \\
Endpoint-only & 86.65 & 62.50 & 92.31 & 93.18 & 61.59 & 22.01 & 23.19 & 25.76 & 77.63 & 82.40 \\
$\lambda_{\max}=0.5$ & 86.14 & 68.27 & 100.00 & 80.63 & 58.42 & 21.97 & 22.69 & 25.66 & 76.97 & 81.97 \\
$\lambda_{\max}=1.0$ & 84.60 & 66.83 & 100.00 & 91.13 & 61.26 & 21.94 & 22.88 & 25.60 & 78.38 & 82.61 \\
\bottomrule
\end{tabular}
\end{table}
 
\clearpage
\subsubsection{VBench}
\label{sec:appendix_ablation_vbench}

\begin{table}[!htbp]
\caption{Detailed quality ablation results on VBench prompts. Scores are shown as percentages ($\uparrow$).}
\label{tab:ablation-vbench-quality-dimensions}
\centering
\footnotesize
\hypersetup{hidelinks}
\setlength{\tabcolsep}{1.9pt}
\setlength{\heavyrulewidth}{0.75pt}
\setlength{\lightrulewidth}{0.35pt}
\renewcommand{\arraystretch}{1.12}
\newcommand{\detailMissing}{\makebox[2.8em]{\textcolor{black!45}{--}}}
\newcommand{\appTableNote}[1]{\par\vspace{2pt}\parbox{\wd\appTableBox}{\raggedright\scriptsize #1\par}}
\newcommand{\detailHeader}[1]{{\scriptsize\begin{tabular}[c]{@{}>{\bfseries}c@{}}#1\end{tabular}}}
 \setlength{\tabcolsep}{5pt}
\begin{tabular}{@{}l*{7}{c}@{\hspace{7pt}}*{1}{c}@{}}
\toprule
\textbf{Variant} & \detailHeader{Subject\\Consistency} & \detailHeader{Background\\Consistency} & \detailHeader{Temporal\\Flickering} & \detailHeader{Motion\\Smoothness} & \detailHeader{Dynamic\\Degree} & \detailHeader{Aesthetic\\Quality} & \detailHeader{Imaging\\Quality} & \detailHeader{Quality\\Score} \\
\midrule
\textbf{Ours} & 97.56 & 96.17 & 97.18 & 98.97 & 88.89 & 67.08 & 66.93 & 86.17 \\
\midrule
w/o CFG decoupling & 97.11 & 95.89 & 98.30 & 98.69 & 65.28 & 66.90 & 68.60 & 84.77 \\
w/o MMD & 97.13 & 95.21 & 96.96 & 98.66 & 77.78 & 67.44 & 69.21 & 85.19 \\
Shuffled-ratio control & 96.91 & 95.38 & 97.08 & 98.77 & 69.44 & 67.79 & 69.03 & 84.68 \\
$K=2$ & 96.61 & 95.16 & 96.71 & 98.71 & 87.50 & 66.13 & 67.49 & 85.29 \\
Endpoint-only & 96.48 & 95.16 & 96.56 & 98.45 & 86.11 & 67.56 & 68.53 & 85.34 \\
$\lambda_{\max}=0.5$ & 96.57 & 95.61 & 97.58 & 98.62 & 62.50 & 67.54 & 69.25 & 84.25 \\
$\lambda_{\max}=1.0$ & 96.47 & 95.12 & 97.45 & 98.67 & 79.17 & 66.77 & 68.81 & 85.21 \\
\bottomrule
\end{tabular}
\end{table}
\begin{table}[!htbp]
\caption{Detailed semantic ablation results on VBench prompts. Scores are shown as percentages ($\uparrow$).}
\label{tab:ablation-vbench-semantic-dimensions}
\centering
\footnotesize
\hypersetup{hidelinks}
\setlength{\tabcolsep}{1.9pt}
\setlength{\heavyrulewidth}{0.75pt}
\setlength{\lightrulewidth}{0.35pt}
\renewcommand{\arraystretch}{1.12}
\newcommand{\detailMissing}{\makebox[2.8em]{\textcolor{black!45}{--}}}
\newcommand{\appTableNote}[1]{\par\vspace{2pt}\parbox{\wd\appTableBox}{\raggedright\scriptsize #1\par}}
\newcommand{\detailHeader}[1]{{\scriptsize\begin{tabular}[c]{@{}>{\bfseries}c@{}}#1\end{tabular}}}
 \setlength{\tabcolsep}{3pt}
\begin{tabular}{@{}l*{9}{c}@{\hspace{7pt}}*{2}{c}@{}}
\toprule
\textbf{Variant} & \detailHeader{Object\\Class} & \detailHeader{Multiple\\Objects} & \detailHeader{Human\\Action} & \detailHeader{Color} & \detailHeader{Spatial\\Relationship} & \detailHeader{Scene} & \detailHeader{Appearance\\Style} & \detailHeader{Temporal\\Style} & \detailHeader{Overall\\Consistency} & \detailHeader{Semantic\\Score} & \detailHeader{Total\\Score} \\
\midrule
\textbf{Ours} & 93.20 & 83.08 & 99.00 & 88.68 & 71.42 & 53.27 & 21.62 & 25.94 & 27.30 & 80.23 & 84.98 \\
\midrule
w/o CFG decoupling & 97.63 & 83.08 & 96.00 & 83.98 & 69.86 & 56.03 & 21.51 & 26.48 & 27.20 & 80.16 & 83.85 \\
w/o MMD & 95.57 & 78.96 & 96.00 & 83.50 & 68.78 & 47.38 & 21.50 & 26.65 & 27.51 & 78.27 & 83.81 \\
Shuffled-ratio control & 94.46 & 82.24 & 97.00 & 83.48 & 71.65 & 53.20 & 21.89 & 26.56 & 27.64 & 79.89 & 83.72 \\
$K=2$ & 91.30 & 79.73 & 96.00 & 84.21 & 64.76 & 48.98 & 21.82 & 25.97 & 26.95 & 77.48 & 83.73 \\
Endpoint-only & 93.75 & 83.16 & 97.00 & 84.42 & 74.09 & 51.74 & 21.93 & 26.68 & 27.76 & 80.18 & 84.31 \\
$\lambda_{\max}=0.5$ & 93.51 & 82.91 & 96.00 & 84.42 & 73.50 & 53.20 & 21.87 & 26.56 & 27.48 & 80.00 & 83.40 \\
$\lambda_{\max}=1.0$ & 93.12 & 79.65 & 98.00 & 84.59 & 67.05 & 53.63 & 21.84 & 26.61 & 27.43 & 79.17 & 84.00 \\
\bottomrule
\end{tabular}
\end{table}
 
\subsection{Visual Quality During DMD Training}
\label{sec:appendix_quality_trajectory}

\begin{figure}[H]
    \centering
    \includegraphics[width=0.86\linewidth]{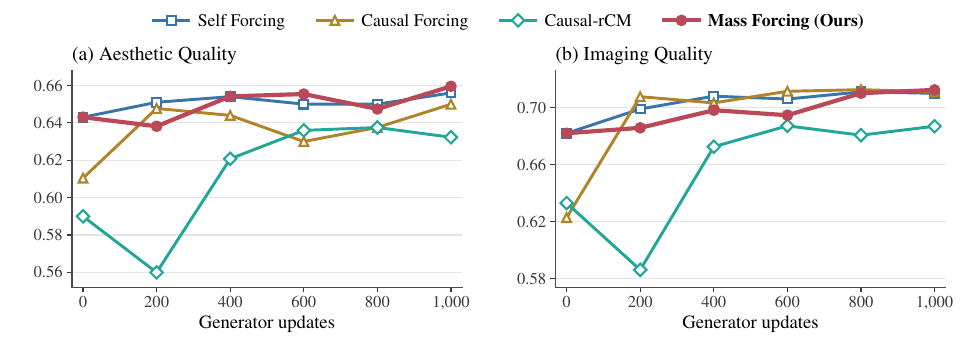}
    \caption{\textbf{Visual quality during DMD training.}
    (a) Aesthetic Quality and (b) Imaging Quality of Self Forcing, Causal Forcing, Causal-rCM, and Mass Forcing, on the original 0--1 scale.
    Markers are checkpoint scores, and lines connect adjacent checkpoints.
    Self Forcing and Mass Forcing share the same ODE initialization at update 0.}
    \label{fig:sf_quality_trajectory}
\end{figure}

The loss of temporal dynamics in the baseline trajectories need not coincide with declining visual quality.
\Cref{fig:sf_quality_trajectory} compares Aesthetic Quality and Imaging Quality for Self Forcing, Causal Forcing, Causal-rCM, and Mass Forcing at generator updates $0,200,400,600,800$, and $1{,}000$.
For Self Forcing, Aesthetic Quality increases from $0.643$ to $0.656$, while Imaging Quality increases from $0.682$ to $0.710$.
Causal Forcing and Causal-rCM also finish above their initial scores in both metrics, although Causal-rCM exhibits an early decline before recovering.
These improvements are not monotonic and coexist with the late-training decline in baseline Dynamic Degree shown in \cref{fig:teaser}.

At update $1{,}000$, Mass Forcing reaches $0.660$ Aesthetic Quality and $0.712$ Imaging Quality, the highest final-checkpoint scores among the four plotted methods, while retaining $83\%$ Dynamic Degree (\cref{fig:teaser}).
The baseline trajectories expose a checkpoint-selection trade-off between visual quality and dynamics, whereas Mass Forcing sustains high scores on both measures at the end of training.

\subsection{Matched Unweighted Baseline}
\label{sec:appendix_matched_baseline}

Mass Forcing replaces the holistic Self Forcing data path with conditional-chunk updates on rollouts that share a prefix.
To separate this change from the mass-aware objectives, we train a uniform conditional-chunk baseline.
It is identical to Mass Forcing in initialization, chunk sampling, shared prefix, $K=4$ rollouts, replay, critic, optimizer, seed, and the 1{,}000 generator updates.
The only differences are $\lambda_{\max}=0$, which gives $h_k=1$, and $\gamma=0$.

\Cref{tab:matched-baseline} compares it with Self Forcing, the w/o MMD variant, and the full method.
The last column reports DD during our own training runs, with the protocol of \cref{fig:teaser}.

The uniform baseline loses its dynamics like Self Forcing: its DD falls from $65\%$ at update 100 to $11\%$ at update 1{,}000.
It does not improve Total over Self Forcing, and it trails Mass Forcing by 1.83 points on MovieGen and 3.00 points on VBench.
The conditional-chunk data path alone therefore explains neither the quality gain nor the sustained dynamics.
Adding mass weighting halts the decline in DD and raises Total by 0.85 and 1.83 points.
Adding MMD then lifts DD to $83\%$ and raises Total by another 0.98 and 1.17 points.

\begin{table}[H]
\caption{Matched unweighted baseline. Total and Dynamic Degree on the full benchmarks, and Dynamic Degree during training.}
\label{tab:matched-baseline}
\centering
\footnotesize
\hypersetup{hidelinks}
\setlength{\tabcolsep}{1.9pt}
\setlength{\heavyrulewidth}{0.75pt}
\setlength{\lightrulewidth}{0.35pt}
\renewcommand{\arraystretch}{1.12}
\newcommand{\detailMissing}{\makebox[2.8em]{\textcolor{black!45}{--}}}
\newcommand{\appTableNote}[1]{\par\vspace{2pt}\parbox{\wd\appTableBox}{\raggedright\scriptsize #1\par}}
\newcommand{\detailHeader}[1]{{\scriptsize\begin{tabular}[c]{@{}>{\bfseries}c@{}}#1\end{tabular}}}
 \setlength{\tabcolsep}{6pt}
\sbox{\appTableBox}{\begin{tabular}{@{}lcc*{4}{c}c@{}}
\toprule
\multirow{2}{*}{\textbf{Method}} & \multirow{2}{*}{\detailHeader{Mass\\weighting}} & \multirow{2}{*}{\textbf{MMD}} &
\multicolumn{2}{c}{\textbf{MovieGen}} & \multicolumn{2}{c}{\textbf{VBench}} &
\multirow{2}{*}{\detailHeader{Dynamic Degree (\%) at update\\$100\rightarrow500\rightarrow1{,}000$}} \\
\cmidrule(lr){4-5}\cmidrule(lr){6-7}
& & & \textbf{Total} & \textbf{Dynamic Degree} & \textbf{Total} & \textbf{Dynamic Degree} & \\
\midrule
Self Forcing & -- & -- & 81.75 & 63.41 & 83.49 & 61.11 & $66\rightarrow36\rightarrow13$ \\
Uniform conditional-chunk & -- & -- & 81.55 & 62.49 & 81.98 & 60.32 & $65\rightarrow33\rightarrow11$ \\
w/o MMD & \checkmark & -- & 82.40 & 65.00 & 83.81 & 77.78 & $65\rightarrow69\rightarrow65$ \\
\textbf{Mass Forcing (Ours)} & \checkmark & \checkmark & \textbf{83.38} & \textbf{83.35} & \textbf{84.98} & \textbf{88.89} & $65\rightarrow77\rightarrow83$ \\
\bottomrule
\end{tabular}}%
\usebox{\appTableBox}
\appTableNote{Total and Dynamic Degree evaluate the checkpoint reported for each method, whereas the last column tracks Dynamic Degree across one training run.}
\end{table}

\clearpage
\section{Analysis of Mode Coverage and the Mass Estimator}
\label{sec:appendix_analysis}

This appendix examines the mechanism behind Mass Forcing.
\Cref{sec:appendix_coverage} measures mode coverage directly, by comparing videos generated from the same prompt with those of the teacher.
\Cref{sec:appendix_mixture,sec:appendix_stability} then turn to the mass estimator.
The weights of \cref{sec:method_ratio} are estimated from a learned score field along finite paths between a small group of rollouts, and we examine how accurate they are where ground truth is available, and how much they move when parts of the estimator are changed.

\subsection{Conditional Mode Coverage}
\label{sec:appendix_coverage}

Dynamic Degree measures temporal dynamics and not mode coverage.
To measure coverage directly, we compare videos generated from the same prompt with different seeds.

The 32 prompts are drawn at random.
For each prompt, every student generates 32 videos with seeds 0 to 31.
The 14B teacher generates 64 videos per prompt, with seeds 0 to 63, using the sampling settings recommended for Wan2.1: 50 steps of the UniPC solver, a guidance scale of 5.0, and a timestep shift of 5.0.
This sampler guidance scale is distinct from the guidance weight $w_{\mathrm{cfg}}=3.0$ of the distillation objective in \cref{tab:hyperparameters}, which scales a term of the training gradient rather than a sampler.
The teacher videos with seeds 0 to 31 form the reference set.
Those with seeds 32 to 63 give the teacher row of \cref{tab:conditional-coverage}, which measures the teacher against itself across disjoint seeds and is therefore an upper bound.
All videos are embedded with VideoMAE.
Diversity is the mean pairwise distance between the 32 videos of one prompt, averaged over prompts.
Precision and recall use $k$-nearest neighbors with $k=3$, taking the teacher videos for the same prompt as the reference set.
\Cref{tab:conditional-coverage} reports the averages over the 32 prompts.

Self Forcing loses diversity relative to its own initialization, from 0.687 to 0.556, which is the behavior that reverse-KL mode seeking predicts.
Mass Forcing starts from the same initialization and ends at 0.787.
It also raises precision against the teacher from 0.351 to 0.455, so the additional variety is not obtained by leaving the teacher's support.
Recall is near zero for every 1.3B student, including the initialization, and Mass Forcing is the only one above zero.
The 1.3B students therefore remain far from covering the teacher's conditional distribution.
We claim only that mass weighting preserves more coverage during training than the unweighted objective does.

\begin{table}[H]
\caption{Conditional mode coverage across seeds for the same prompt. Diversity is the mean pairwise distance within a prompt. Precision and recall are computed against teacher videos for the same prompt.}
\label{tab:conditional-coverage}
\centering
\footnotesize
\hypersetup{hidelinks}
\setlength{\tabcolsep}{1.9pt}
\setlength{\heavyrulewidth}{0.75pt}
\setlength{\lightrulewidth}{0.35pt}
\renewcommand{\arraystretch}{1.12}
\newcommand{\detailMissing}{\makebox[2.8em]{\textcolor{black!45}{--}}}
\newcommand{\appTableNote}[1]{\par\vspace{2pt}\parbox{\wd\appTableBox}{\raggedright\scriptsize #1\par}}
\newcommand{\detailHeader}[1]{{\scriptsize\begin{tabular}[c]{@{}>{\bfseries}c@{}}#1\end{tabular}}}
 \setlength{\tabcolsep}{6pt}
\sbox{\appTableBox}{\begin{tabular}{@{}l*{3}{c}@{}}
\toprule
\textbf{Model} & \textbf{Diversity}\,$\uparrow$ & \textbf{Precision}\,$\uparrow$ & \textbf{Recall}\,$\uparrow$ \\
\midrule
Teacher (14B), new seeds & 0.865 & 0.774 & 0.830 \\
\midrule
ODE initialization & 0.687 & 0.010 & 0.000 \\
Self Forcing & 0.556 & 0.351 & 0.000 \\
\textbf{Mass Forcing (Ours)} & \textbf{0.787} & \textbf{0.455} & \textbf{0.045} \\
\bottomrule
\end{tabular}}%
\usebox{\appTableBox}
\appTableNote{The teacher row uses seeds disjoint from the reference set. It is an upper bound, not a competing method.}
\end{table}

\subsection{Validation of the Mass Estimator on Gaussian Mixtures}
\label{sec:appendix_mixture}
\label{sec:appendix_estimator}

On video data the true density ratios are unknown.
We therefore validate the estimator on Gaussian mixtures, where they are available in closed form.

Teacher and student share four well-separated components.
The teacher weights them uniformly and the student weights them $[0.70, 0.20, 0.09, 0.01]$, so the student under-represents three of the four.
We draw 4{,}096 groups of $K=4$ samples from the student per configuration and run the estimator of \cref{sec:method_ratio} unchanged, with three-point Simpson quadrature, a medoid anchor, and the noise levels used in training.
The experiment is repeated in 2 and in 32 dimensions.
Two score oracles are compared.
Exact scores isolate the error of the path integration.
A fake score learned as in Mass Forcing adds the estimation error that is present during training, and its results are averaged over three seeds.

\Cref{tab:mixture-ordering} reports how often two samples of a group are ordered correctly.
Two samples drawn from the same mixture component have almost identical true ratios and no well-defined ordering, so such pairs are excluded from the accuracy.
With exact scores, 96 to 99 percent of pairs are ordered correctly at every noise level and in both dimensions.
With the learned fake score the accuracy is 77 to 94 percent, against a chance level of 50.
When the student equals the teacher, the estimated weights stay near uniform, with an L1 deviation of at most 0.012.

The estimator recovers the ordering of the weights and not their magnitude.
With exact scores, the L1 error of the operative weights is 0.36 to 0.40 at the lower noise levels, so the estimated ratios are not calibrated density ratios.
Mass Forcing uses the ordering only.
The weights are centered, tempered, clipped, and mixed with uniform weights before use, and at $\lambda_{\max}=0.25$ each weight lies in $[0.75, 1.75]$ (\cref{sec:app_ratio}).

\begin{table}[H]
\caption{Ordering accuracy of the estimated mass weights on Gaussian mixtures with known density ratios. Each entry is the fraction of sample pairs within a group of $K=4$ that are ordered correctly, in percent, against a chance level of 50.}
\label{tab:mixture-ordering}
\centering
\footnotesize
\hypersetup{hidelinks}
\setlength{\tabcolsep}{1.9pt}
\setlength{\heavyrulewidth}{0.75pt}
\setlength{\lightrulewidth}{0.35pt}
\renewcommand{\arraystretch}{1.12}
\newcommand{\detailMissing}{\makebox[2.8em]{\textcolor{black!45}{--}}}
\newcommand{\appTableNote}[1]{\par\vspace{2pt}\parbox{\wd\appTableBox}{\raggedright\scriptsize #1\par}}
\newcommand{\detailHeader}[1]{{\scriptsize\begin{tabular}[c]{@{}>{\bfseries}c@{}}#1\end{tabular}}}
 \setlength{\tabcolsep}{6pt}
\sbox{\appTableBox}{\begin{tabular}{@{}l*{4}{c}@{}}
\toprule
\multirow{2}{*}{\textbf{Noise level $\sigma$}} &
\multicolumn{2}{c}{\textbf{Exact scores}} &
\multicolumn{2}{c@{}}{\textbf{Learned fake score}} \\
\cmidrule(lr){2-3}\cmidrule(lr){4-5}
& $d=2$ & $d=32$ & $d=2$ & $d=32$ \\
\midrule
0.20 & 96.1 & 96.2 & 77.3 & 84.3 \\
0.425 & 97.2 & 97.3 & 82.6 & 89.5 \\
0.65 & 98.5 & 98.6 & 93.9 & 90.6 \\
\bottomrule
\end{tabular}}%
\usebox{\appTableBox}
\appTableNote{Exact scores isolate the integration error. The learned fake score adds the estimation error present in training.}
\end{table}
 
\subsection{Stability of the Estimated Weights}
\label{sec:appendix_stability}

We also measure how much the weights move when one part of the estimator is changed on the trained model.
The design covers 8 prompts, chunks 0, 3, and 6, two prefix seeds, and three noise levels, which gives 144 groups of $K=4$ rollouts.
Chunks 3 and 6 fall in active cells, and \cref{tab:estimator-stability} reports the 96 groups in those cells.
Each row changes one factor and compares the resulting weights with the default estimator on the same rollouts.

Refining the quadrature from three to nine points changes the ranking least.
Rotating the anchor and resampling the query noise change it slightly more.
In all three cases the median rank correlation is 1, and in at least 90 percent of groups the ranking of the four rollouts changes by at most one adjacent swap.
The L1 change in the weights stays at or below 0.030.

A path that passes through a second rollout, rather than going directly to the anchor, reproduces the rotated-anchor values to within $5\times10^{-5}$ nats.
This agreement follows numerically from the change of anchor and is not an independent check.

These measurements were taken after training and are diagnostic.
They were not used to select the quadrature, the anchor, or the gating thresholds, all of which are fixed before DMD training as described in \cref{sec:impl_mass}.

\begin{table}[H]
\caption{Stability of the estimated weights under one changed factor at a time, over 96 groups of $K=4$ rollouts in active cells. Each row is compared with the default estimator on the same rollouts.}
\label{tab:estimator-stability}
\centering
\footnotesize
\hypersetup{hidelinks}
\setlength{\tabcolsep}{1.9pt}
\setlength{\heavyrulewidth}{0.75pt}
\setlength{\lightrulewidth}{0.35pt}
\renewcommand{\arraystretch}{1.12}
\newcommand{\detailMissing}{\makebox[2.8em]{\textcolor{black!45}{--}}}
\newcommand{\appTableNote}[1]{\par\vspace{2pt}\parbox{\wd\appTableBox}{\raggedright\scriptsize #1\par}}
\newcommand{\detailHeader}[1]{{\scriptsize\begin{tabular}[c]{@{}>{\bfseries}c@{}}#1\end{tabular}}}
 \setlength{\tabcolsep}{6pt}
\sbox{\appTableBox}{\begin{tabular}{@{}l*{2}{c}@{}}
\toprule
\textbf{Changed factor} & \detailHeader{Spearman rank\\correlation $\uparrow$} & \detailHeader{L1 change in\\weights $h$ $\downarrow$} \\
\midrule
Simpson quadrature, 3 points $\rightarrow$ 9 points & 0.954 & 0.016 \\
Rotated anchor & 0.929 & 0.030 \\
Resampled query noise & 0.935 & 0.019 \\
\bottomrule
\end{tabular}}%
\usebox{\appTableBox}
\appTableNote{The median rank correlation is 1 in all three cases. At $K=4$, a correlation of 0.8 corresponds to a single adjacent swap.}
\end{table}

\section{Additional Evaluations}
\label{sec:appendix_additional_eval}

The main experiments use 5-second videos, automatic metrics, and a single training run.
This appendix extends them along each of these three axes: 60-second videos (\cref{sec:appendix_long_video}), a user study (\cref{sec:appendix_user_study}), and a second training seed (\cref{sec:appendix_runs}).

\subsection{Long-Video Evaluation}
\label{sec:appendix_long_video}

The main evaluation uses 5-second videos, which is the protocol of the methods we compare with.
To check whether the sustained dynamics survive at a longer horizon~\citep{feng_2026_iclr_narrlv}, we also evaluate 60-second videos on VBench-Long.
We use the full official VBench prompt set and generate one video per prompt with a fixed seed, which gives 944 videos per method.
VBench-Long scores a video by splitting it into clips of 2 or 10 seconds, depending on the dimensions that the prompt belongs to, and averaging the per-clip scores.

Self Forcing, Causal Forcing, and Mass Forcing are rolled out with Infinity-RoPE~\citep{yesiltepe_2026_cvpr_infinity_rope}, a training-free inference technique for Self-Forcing-style models, under one shared configuration.
\Cref{tab:vbench-long} reports the results, and Mass Forcing is highest in Quality, Semantic, and Total.

We also evaluate the last 5 seconds of each 60-second video, on a smaller pool of 20 prompts with 2 seeds per prompt.
Mass Forcing keeps 78 percent Dynamic Degree in that window, against 9 percent for Self Forcing.
On the same pool, its Imaging Quality declines from 69.9 to 67.8 across the 60 seconds.
Self Forcing, Causal Forcing, and Mass Forcing share the same rollout technique here, so the difference in the last window comes from training rather than from inference.

\Cref{fig:long_video_frames} shows frames from one 60-second rollout of each method, at matched timestamps and from the same prompt and seed.

\begin{table}[H]
\caption{VBench-Long evaluation of 60-second videos, one video per prompt on the full official VBench prompt set.}
\label{tab:vbench-long}
\centering
\footnotesize
\hypersetup{hidelinks}
\setlength{\tabcolsep}{1.9pt}
\setlength{\heavyrulewidth}{0.75pt}
\setlength{\lightrulewidth}{0.35pt}
\renewcommand{\arraystretch}{1.12}
\newcommand{\detailMissing}{\makebox[2.8em]{\textcolor{black!45}{--}}}
\newcommand{\appTableNote}[1]{\par\vspace{2pt}\parbox{\wd\appTableBox}{\raggedright\scriptsize #1\par}}
\newcommand{\detailHeader}[1]{{\scriptsize\begin{tabular}[c]{@{}>{\bfseries}c@{}}#1\end{tabular}}}
 \setlength{\tabcolsep}{6pt}
\sbox{\appTableBox}{\begin{tabular}{@{}l*{3}{c}@{}}
\toprule
\textbf{Metric} & \textbf{Self Forcing} & \textbf{Causal Forcing} & \textbf{Mass Forcing (Ours)} \\
\midrule
Quality & 79.31 & \underline{83.95} & \textbf{84.03} \\
Semantic & 69.05 & \underline{71.20} & \textbf{78.95} \\
Total & 77.26 & \underline{81.40} & \textbf{83.01} \\
\bottomrule
\end{tabular}}%
\usebox{\appTableBox}
\appTableNote{Infinity-RoPE is applied to all three methods under one shared configuration.}
\end{table}

\begingroup
\newlength{\longVidLabelWidth}
\newlength{\longVidFrameWidth}
\newlength{\longVidImageGap}
\setlength{\longVidLabelWidth}{32pt}
\setlength{\longVidImageGap}{1pt}
\setlength{\longVidFrameWidth}{\dimexpr(\linewidth-\longVidLabelWidth-3pt-5\longVidImageGap)/6\relax}
\newcommand{\longVidFrame}[2]{%
    \raisebox{-0.5\height}{\mfImage{assets/long-video/prompt_001_seed0/#1/#2}}}
\newcommand{\longVidMethod}[1]{%
    \parbox[c]{\longVidLabelWidth}{\centering#1}}

\begin{figure}[H]
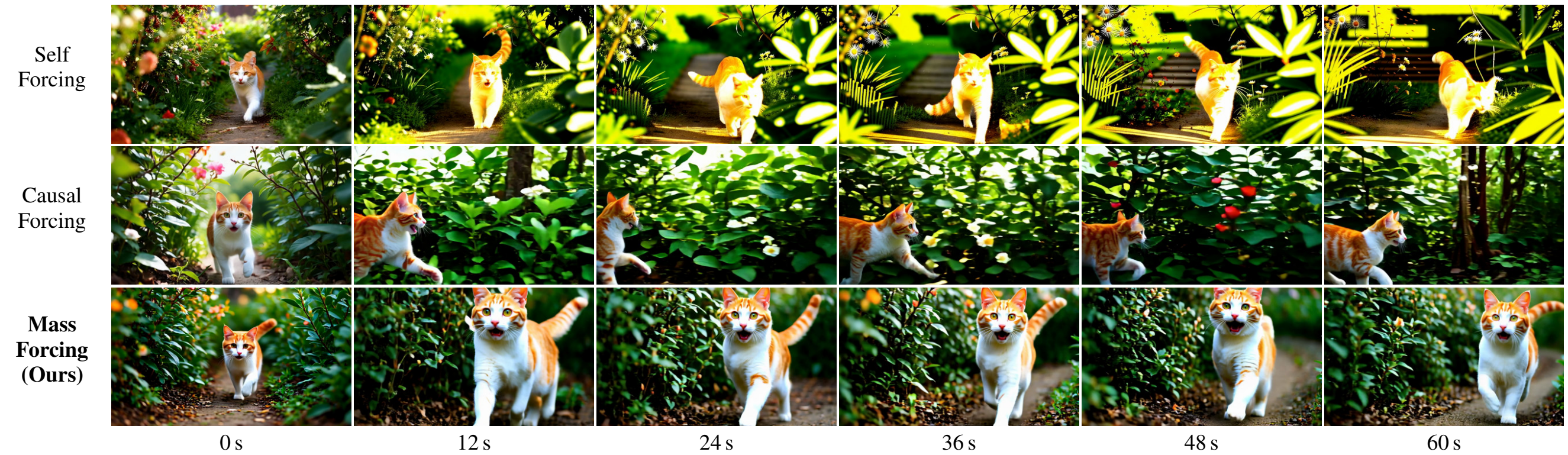

    \centering
    \begingroup
    \small
    \noindent\parbox{\linewidth}{\textbf{Prompt.} \textit{A cinematic and grainy photograph captures a white and orange tabby cat \textbf{joyfully darting through a dense garden}, as if chasing something. The cat's eyes are wide and filled with happiness as it jogs forward, scanning the branches, flowers, and leaves. The narrow path winds between the lush greenery, and the scene is captured from a ground-level angle, providing a low and intimate perspective. The image has \textbf{warm tones} and a subtle grainy texture, with scattered daylight filtering through the leaves and plants above, creating a warm contrast that highlights the cat's orange fur. The shot is clear and sharp, with a shallow depth of field that focuses solely on the cat's movements and expressions.}}\par
    \vspace{4pt}
    \scriptsize
    \setlength{\tabcolsep}{0pt}
    \setlength{\extrarowheight}{0pt}
    \renewcommand{\arraystretch}{1}
    \begin{tabular}{@{}c@{\hspace{3pt}}c@{\hspace{\longVidImageGap}}c@{\hspace{\longVidImageGap}}c@{\hspace{\longVidImageGap}}c@{\hspace{\longVidImageGap}}c@{\hspace{\longVidImageGap}}c@{}}
        \longVidMethod{Self\\Forcing} &
        \longVidFrame{self_forcing}{f1.jpg} &
        \longVidFrame{self_forcing}{f2.jpg} &
        \longVidFrame{self_forcing}{f3.jpg} &
        \longVidFrame{self_forcing}{f4.jpg} &
        \longVidFrame{self_forcing}{f5.jpg} &
        \longVidFrame{self_forcing}{f6.jpg} \\
        \noalign{\vskip\longVidImageGap}
        \longVidMethod{Causal\\Forcing} &
        \longVidFrame{causal_forcing}{f1.jpg} &
        \longVidFrame{causal_forcing}{f2.jpg} &
        \longVidFrame{causal_forcing}{f3.jpg} &
        \longVidFrame{causal_forcing}{f4.jpg} &
        \longVidFrame{causal_forcing}{f5.jpg} &
        \longVidFrame{causal_forcing}{f6.jpg} \\
        \noalign{\vskip\longVidImageGap}
        \longVidMethod{\textbf{Mass}\\\textbf{Forcing}\\\textbf{(Ours)}} &
        \longVidFrame{ours}{f1.jpg} &
        \longVidFrame{ours}{f2.jpg} &
        \longVidFrame{ours}{f3.jpg} &
        \longVidFrame{ours}{f4.jpg} &
        \longVidFrame{ours}{f5.jpg} &
        \longVidFrame{ours}{f6.jpg} \\
        \noalign{\vskip2pt}
        & 0\,s & 12\,s & 24\,s & 36\,s & 48\,s & 60\,s \\
    \end{tabular}\par
    \endgroup
\caption{\textbf{Sixty-second rollouts for one prompt.} All three methods use the same prompt and the same seed, and frames are taken at matched timestamps. Self Forcing drifts into increasingly saturated greens and yellows after about 24 seconds, the color oversaturation that motivates this work. Causal Forcing keeps its colors, but the cat drifts to the left edge of the frame and is partly cut off in the later frames. Mass Forcing keeps stable color and a consistently framed subject whose pose changes across the full minute. This is one example and is not evidence for the aggregate scores in \cref{tab:vbench-long}.}
\label{fig:long_video_frames}
\end{figure}
\endgroup

\subsection{User Study}
\label{sec:appendix_user_study}

Automatic metrics do not settle whether the additional motion is judged natural.
We therefore ran a blind user study with 116 participants.

Each participant saw ten sets of four videos generated from the same prompt.
Three of the four videos came from Self Forcing, Causal Forcing, and Mass Forcing, and the fourth came from an additional method that we do not compare with in this paper.
Eight sets used 5-second videos and two used 60-second videos, with prompts drawn from a pool of 100.
Within each set the order of the four videos was randomized, and the methods were not named.
Participants ranked the four videos on motion naturalness, dynamic degree, and imaging quality, ranking each criterion separately.

For simplicity, we remove the additional method from each ranking and report the rankings among the three methods compared here.
\Cref{tab:user-study} reports how often each method was ranked first.
Mass Forcing is ranked first most often on all three criteria, for both short and long videos.
A Friedman test, treating participants as blocks and using each participant's mean rank, rejects equal ranks at $p<0.001$ for every criterion.
Uncertainty is obtained by resampling participants 10{,}000 times.
Mass Forcing leads on motion naturalness and imaging quality as well as on dynamic degree, so its additional motion is not judged to come at the cost of naturalness or visual quality.

\begin{table}[H]
\caption{Blind user study with 116 participants. Each participant ranked ten sets of four videos on three criteria, with the order of the videos randomized within each set. The fourth video in each set came from an additional method, which we remove from each ranking. Entries are the percentage of sets in which a method was ranked first among the three methods shown, against a chance level of 33.3.}
\label{tab:user-study}
\centering
\footnotesize
\hypersetup{hidelinks}
\setlength{\tabcolsep}{1.9pt}
\setlength{\heavyrulewidth}{0.75pt}
\setlength{\lightrulewidth}{0.35pt}
\renewcommand{\arraystretch}{1.12}
\newcommand{\detailMissing}{\makebox[2.8em]{\textcolor{black!45}{--}}}
\newcommand{\appTableNote}[1]{\par\vspace{2pt}\parbox{\wd\appTableBox}{\raggedright\scriptsize #1\par}}
\newcommand{\detailHeader}[1]{{\scriptsize\begin{tabular}[c]{@{}>{\bfseries}c@{}}#1\end{tabular}}}
 \setlength{\tabcolsep}{6pt}
\begin{tabular}{@{}l*{3}{c}@{}}
\toprule
\textbf{Ranked first (\%)} & \textbf{Self Forcing} & \textbf{Causal Forcing} & \textbf{Mass Forcing (Ours)} \\
\midrule
\multicolumn{4}{@{}l}{\textit{Short videos}} \\
Motion naturalness & 2.2 & \underline{16.3} & \textbf{81.6} \\
Dynamic degree & 2.6 & \underline{30.2} & \textbf{67.2} \\
Imaging quality & 5.9 & \underline{11.1} & \textbf{83.0} \\
\addlinespace[2pt]
\multicolumn{4}{@{}l}{\textit{Long videos}} \\
Motion naturalness & 3.4 & \underline{28.4} & \textbf{68.1} \\
Dynamic degree & 3.0 & \underline{47.0} & \textbf{50.0} \\
Imaging quality & 9.9 & \underline{18.5} & \textbf{71.6} \\
\bottomrule
\end{tabular}
\end{table}

\subsection{Robustness to the Training Seed}
\label{sec:appendix_runs}
\label{sec:appendix_seeds}

\Cref{tab:main-table} reports a single training run.
To check that the result is not specific to that run, we trained Mass Forcing a second time, changing only the training seed from 3407 to 3408.

\Cref{tab:training-seeds} compares the two runs.
They differ by at most 0.18 in Total and by at most 0.99 in Dynamic Degree on either benchmark.
Both exceed every baseline in \cref{tab:main-table} in Total and in Dynamic Degree on both benchmarks.

The effect of mass weighting is larger than this run-to-run variation.
The four runs trained without it, namely Self Forcing, Causal Forcing, and Causal-rCM in \cref{fig:teaser} together with the uniform baseline of \cref{sec:appendix_matched_baseline}, all end between 11 and 32 percent Dynamic Degree.
The runs trained with it end between 65 and 83 percent.

\begin{table}[H]
\caption{Mass Forcing trained twice with different training seeds and otherwise identical settings.}
\label{tab:training-seeds}
\centering
\footnotesize
\hypersetup{hidelinks}
\setlength{\tabcolsep}{1.9pt}
\setlength{\heavyrulewidth}{0.75pt}
\setlength{\lightrulewidth}{0.35pt}
\renewcommand{\arraystretch}{1.12}
\newcommand{\detailMissing}{\makebox[2.8em]{\textcolor{black!45}{--}}}
\newcommand{\appTableNote}[1]{\par\vspace{2pt}\parbox{\wd\appTableBox}{\raggedright\scriptsize #1\par}}
\newcommand{\detailHeader}[1]{{\scriptsize\begin{tabular}[c]{@{}>{\bfseries}c@{}}#1\end{tabular}}}
 \setlength{\tabcolsep}{6pt}
\sbox{\appTableBox}{\begin{tabular}{@{}l*{8}{c}@{}}
\toprule
\multirow{2}{*}{\textbf{Training seed}} &
\multicolumn{4}{c}{\textbf{MovieGen}\,$\uparrow$} &
\multicolumn{4}{c@{}}{\textbf{VBench}\,$\uparrow$} \\
\cmidrule(lr){2-5}\cmidrule(lr){6-9}
& \textbf{Total} & \textbf{Quality} & \textbf{Semantic} & \textbf{Dynamic Degree} & \textbf{Total} & \textbf{Quality} & \textbf{Semantic} & \textbf{Dynamic Degree} \\
\midrule
3407 (reported in \cref{tab:main-table}) & 83.38 & 84.53 & 78.78 & 83.35 & 84.98 & 86.17 & 80.23 & 88.89 \\
3408 & 83.54 & 84.67 & 79.03 & 82.36 & 84.80 & 85.95 & 80.21 & 88.77 \\
\midrule
Difference & 0.16 & 0.14 & 0.25 & 0.99 & 0.18 & 0.22 & 0.02 & 0.12 \\
\bottomrule
\end{tabular}}%
\usebox{\appTableBox}
\appTableNote{Both runs exceed every baseline in \cref{tab:main-table} in Total and in Dynamic Degree on both benchmarks.}
\end{table}

\clearpage
\section{Additional Qualitative Results}
\label{sec:appendix_qualitative}

\Cref{sec:extended_qualitative} compares Mass Forcing with the baselines on further prompts, and \cref{sec:qualitative_ablation} compares the ablation variants on one prompt.

\subsection{Extended Qualitative Comparisons}
\label{sec:extended_qualitative}

\Cref{fig:extended_scenes,fig:extended_subjects} compare prompt alignment and temporal variation, with full prompts and key requirements highlighted in bold.
Rows show Self Forcing, Causal Forcing, and Mass Forcing.
Columns correspond to $0$, $1.25$, $2.5$, $3.75$, and $5$ seconds.

In \cref{fig:extended_scenes}, Mass Forcing shows more pronounced changes in the waves and foam, making the requested crashing surf more evident across time.
For the dewdrop, it renders a larger, more detailed miniature world of reflected foliage within the droplet and develops the close-up across frames, whereas the baselines retain smaller reflections and show more limited compositional change.

In \cref{fig:extended_subjects}, the dragon exhibits clearer changes in head pose and framing while retaining detailed silver-edged scales.
The listening man shows more expressive changes in his smile, gaze, and upper-body posture, better conveying the requested amused chuckle than the comparatively restrained baseline sequences.
The illustrated boy provides a stylized example with visible changes in expression and gaze, while retaining the clasped hands and lifted-foot pose specified in the prompt.

\begingroup
\newlength{\extendedQualLabelWidth}
\newlength{\extendedQualFrameWidth}
\newlength{\extendedQualImageGap}
\setlength{\extendedQualLabelWidth}{32pt}
\setlength{\extendedQualImageGap}{1pt}
\setlength{\extendedQualFrameWidth}{\dimexpr(0.8\linewidth-\extendedQualLabelWidth-3pt-4\extendedQualImageGap)/5\relax}
\newcommand{\extendedQualFrame}[3]{%
    \raisebox{-0.5\height}{\mfImage{assets/extended-qualitative/prompt_#1/#2/#3}}}
\newcommand{\extendedQualMethod}[1]{%
    \parbox[c]{\extendedQualLabelWidth}{\centering#1}}
\newcommand{\extendedQualCase}[2]{%
        \centering
        \begingroup
        \small
        \noindent\parbox{\linewidth}{\textbf{Prompt.} \textit{#2}}\par
        \vspace{4pt}
        \scriptsize
        \setlength{\tabcolsep}{0pt}
        \setlength{\extrarowheight}{0pt}
        \renewcommand{\arraystretch}{1}
        \begin{tabular}{@{}c@{\hspace{3pt}}c@{\hspace{\extendedQualImageGap}}c@{\hspace{\extendedQualImageGap}}c@{\hspace{\extendedQualImageGap}}c@{\hspace{\extendedQualImageGap}}c@{}}
            \extendedQualMethod{Self\\Forcing} &
            \extendedQualFrame{#1}{self_forcing}{frame_000_0s.jpg} &
            \extendedQualFrame{#1}{self_forcing}{frame_020_1p25s.jpg} &
            \extendedQualFrame{#1}{self_forcing}{frame_040_2p5s.jpg} &
            \extendedQualFrame{#1}{self_forcing}{frame_060_3p75s.jpg} &
            \extendedQualFrame{#1}{self_forcing}{frame_080_5s.jpg} \\
            \noalign{\vskip\extendedQualImageGap}
            \extendedQualMethod{Causal\\Forcing} &
            \extendedQualFrame{#1}{causal_forcing}{frame_000_0s.jpg} &
            \extendedQualFrame{#1}{causal_forcing}{frame_020_1p25s.jpg} &
            \extendedQualFrame{#1}{causal_forcing}{frame_040_2p5s.jpg} &
            \extendedQualFrame{#1}{causal_forcing}{frame_060_3p75s.jpg} &
            \extendedQualFrame{#1}{causal_forcing}{frame_080_5s.jpg} \\
            \noalign{\vskip\extendedQualImageGap}
            \extendedQualMethod{\textbf{Mass}\\\textbf{Forcing}\\\textbf{(Ours)}} &
            \extendedQualFrame{#1}{ours}{frame_000_0s.jpg} &
            \extendedQualFrame{#1}{ours}{frame_020_1p25s.jpg} &
            \extendedQualFrame{#1}{ours}{frame_040_2p5s.jpg} &
            \extendedQualFrame{#1}{ours}{frame_060_3p75s.jpg} &
            \extendedQualFrame{#1}{ours}{frame_080_5s.jpg} \\
            \noalign{\vskip2pt}
            & 0\,s & 1.25\,s & 2.5\,s & 3.75\,s & 5\,s \\
        \end{tabular}\par
        \endgroup}

\begin{figure}[H]
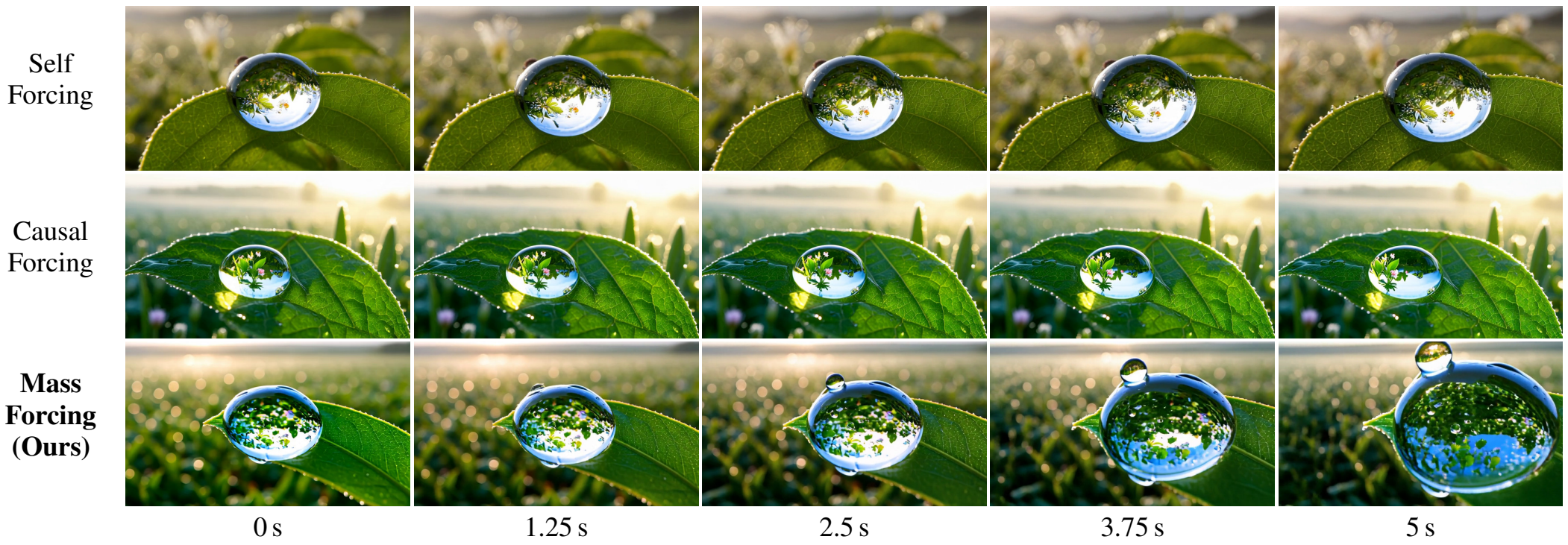

    \centering
\extendedQualCase{000823}{A dramatic landscape photograph capturing a steep cliffside with a stone pedestal jutting out, leading down to reveal the \textbf{powerful waves crashing against jagged rocks} far below. The cliffs are rugged and weathered, with green moss covering parts of the rock faces. The water is turbulent, with white foam rising from the impact of the waves against the rocks. The sky above is a mix of deep blues and grays, creating a moody and atmospheric scene. The photo has a natural and realistic texture, emphasizing the \textbf{dynamic movement of the waves} and the dramatic drop-off. A wide-angle shot from a low angle, looking up towards the cliff.}

\par\vspace{12pt}
\extendedQualCase{000827}{A close-up zoom-in on a morning dewdrop perched on a leaf, capturing the \textbf{intricate reflection of its surroundings within its translucent surface}. The dewdrop glistens under the morning sunlight, revealing \textbf{a miniature world of shimmering leaves, distant flowers, and faint shadows}. The leaf itself is crisp and green, with visible veins and edges. The background is a blurred landscape, hinting at a dewy meadow bathed in early morning light. The dewdrop sparkles with a soft, ethereal glow, emphasizing the beauty of nature's smallest wonders. A macro shot from a slightly tilted angle.}
\caption{\textbf{Scene dynamics and fine-grained prompt details.}
Top: waves breaking against a cliff. Bottom: a dewdrop containing a miniature reflected world.}
\label{fig:extended_scenes}
\end{figure}

\clearpage

\begin{figure}[H]
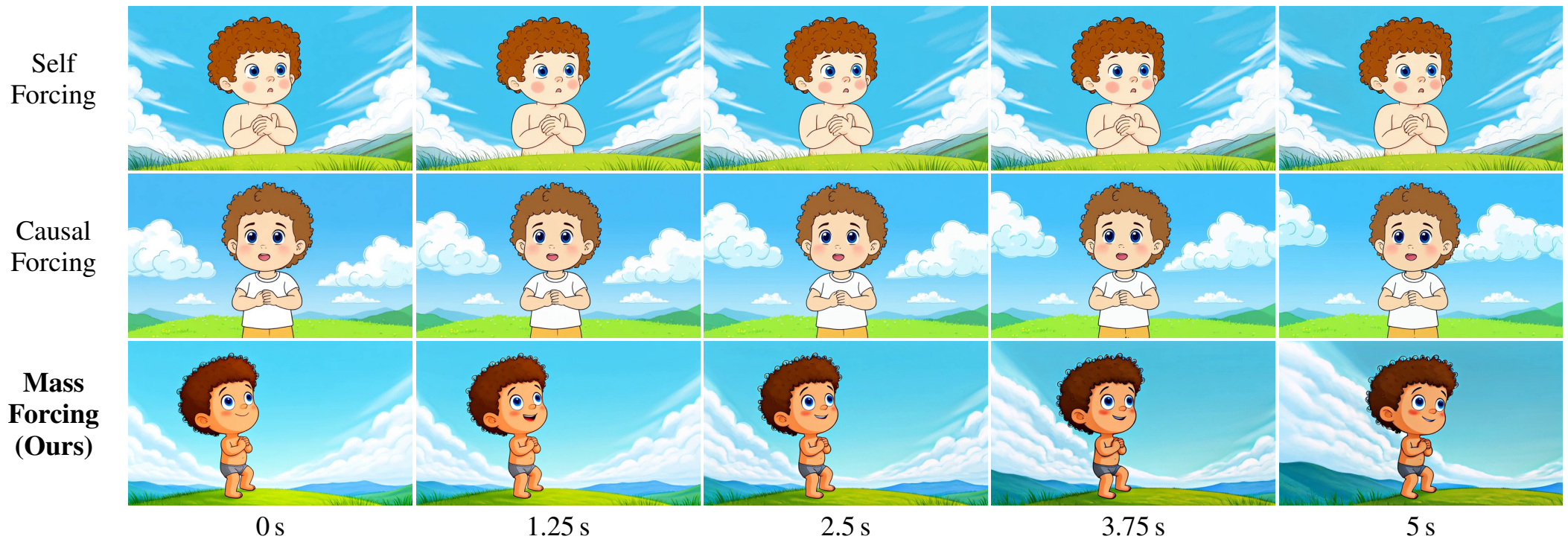

    \centering
\extendedQualCase{000391}{A close-up shot of a majestic white dragon with \textbf{pearlescent, silver-edged scales}, icy blue eyes, and elegant ivory horns. The dragon's face is detailed with subtle wrinkles and sharp, defined features, capturing a regal and serene expression. Its \textbf{breath forms a gentle mist}, adding to the ethereal quality. The scales are meticulously textured, reflecting light in a way that highlights their depth and shine. Set against a softly blurred background, the scene is bathed in a soft, ambient glow, emphasizing the dragon's majesty and otherworldly presence. The background hints at a misty forest, with blurred outlines of ancient trees and vines, creating a mystical atmosphere.}

\par\vspace{4pt}
\extendedQualCase{000338}{A middle-aged man with a \textbf{kind smile and amused chuckle} listens intently to a funny story being told. He has short graying hair and wears a casual blue shirt with a pair of khaki pants. His hands are loosely clasped together, and he \textbf{leans slightly forward} in his comfortable armchair. The background shows a cozy living room with warm lighting, books on a nearby shelf, and a small potted plant on the windowsill. The scene captures the moment when he is fully engaged, with a soft focus on his face and a slight tilt of his head. A close-up shot from a slightly elevated angle.}
\par\vspace{4pt}
\extendedQualCase{000868}{A hand-drawn simple line art illustration of a young boy with a \textbf{look of wonder and amazement} on his face, gazing up at the sky. He has curly brown hair and bright blue eyes that sparkle with curiosity. His small hands are clasped together in front of him, and he stands on a grassy hill, \textbf{one foot slightly lifted}. The background features a clear blue sky with fluffy clouds and distant mountains, creating a serene and peaceful atmosphere. A close-up shot from a slightly lower angle, capturing the child's innocent and awe-filled expression.}
\caption{\textbf{Character appearance and motion.}
Top: a white dragon with detailed scales. Middle: a man reacting to a humorous story. Bottom: an illustrated boy gazing at the sky.}
\label{fig:extended_subjects}
\end{figure}
\endgroup

\clearpage
\subsection{Qualitative Ablation}
\label{sec:qualitative_ablation}

\begin{figure}[H]
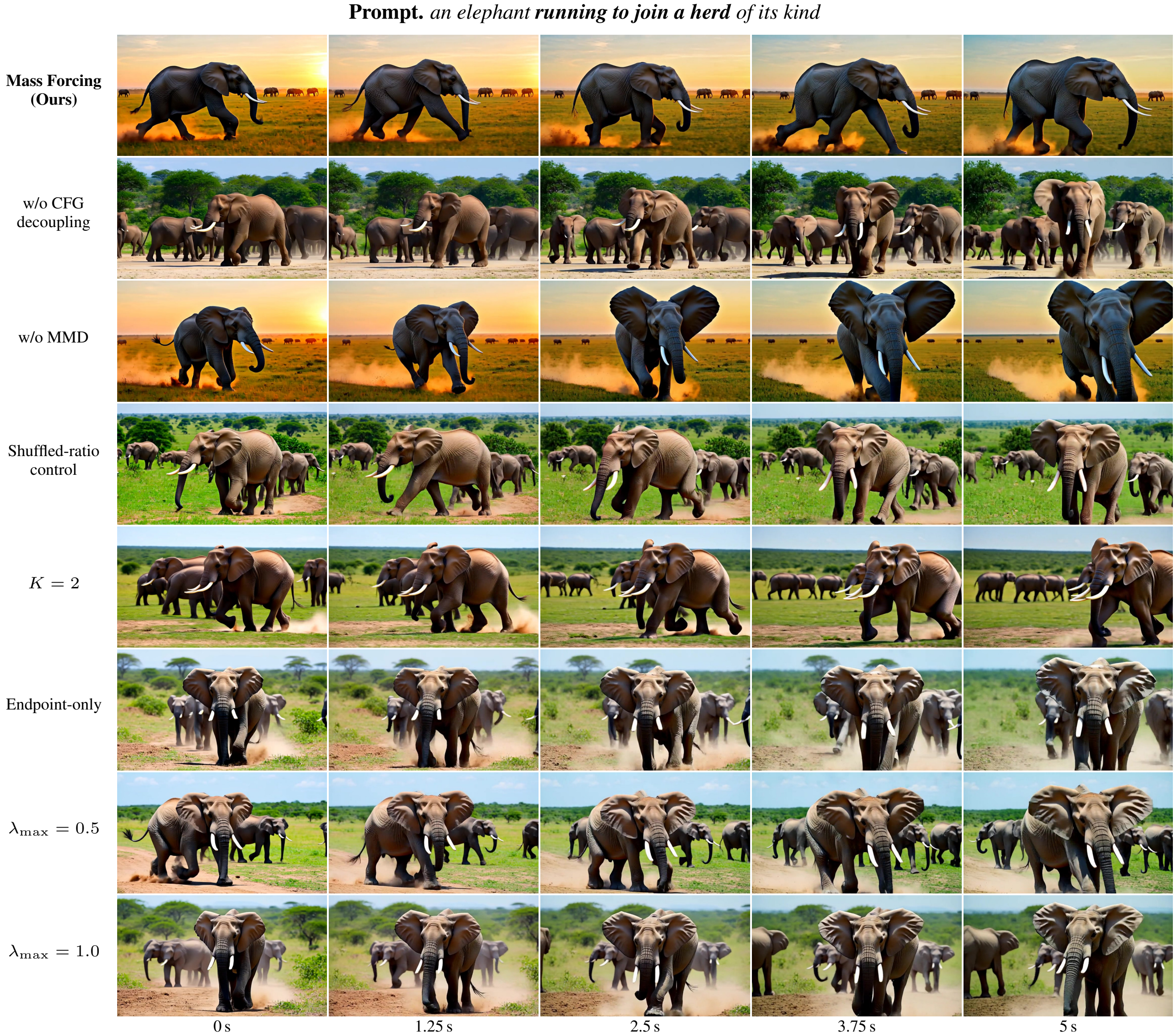

    \centering
    \begingroup
    \newlength{\ablationQualLabelWidth}
    \newlength{\ablationQualFrameWidth}
    \newlength{\ablationQualGap}
    \setlength{\ablationQualLabelWidth}{48pt}
    \setlength{\ablationQualGap}{1pt}
    \setlength{\ablationQualFrameWidth}{\dimexpr(\linewidth-\ablationQualLabelWidth-3pt-4\ablationQualGap)/5\relax}
    \newcommand{\ablationQualFrame}[2]{%
        \raisebox{-0.5\height}{\mfImage{assets/qualitative-ablation/prompt_000319/#1/#2}}}
    \newcommand{\ablationQualRow}[2]{%
        \parbox[c]{\ablationQualLabelWidth}{\centering#1} &
        \ablationQualFrame{#2}{frame_000_0s.jpg} &
        \ablationQualFrame{#2}{frame_020_1p25s.jpg} &
        \ablationQualFrame{#2}{frame_040_2p5s.jpg} &
        \ablationQualFrame{#2}{frame_060_3p75s.jpg} &
        \ablationQualFrame{#2}{frame_080_5s.jpg} \\
        \noalign{\vskip\ablationQualGap}}
    {\small\textbf{Prompt.} \textit{an elephant \textbf{running to join a herd} of its kind}\par}
    \vspace{4pt}
    \scriptsize
    \setlength{\tabcolsep}{0pt}
    \setlength{\extrarowheight}{0pt}
    \renewcommand{\arraystretch}{1}
    \begin{tabular}{@{}c@{\hspace{3pt}}c@{\hspace{\ablationQualGap}}c@{\hspace{\ablationQualGap}}c@{\hspace{\ablationQualGap}}c@{\hspace{\ablationQualGap}}c@{}}
        \ablationQualRow{\textbf{Mass Forcing}\\\textbf{(Ours)}}{ours}
        \ablationQualRow{w/o CFG\\decoupling}{wo_cfg_decoupling}
        \ablationQualRow{w/o MMD}{wo_mmd}
        \ablationQualRow{Shuffled-ratio\\control}{shuffled_ratio}
        \ablationQualRow{$K=2$}{k2}
        \ablationQualRow{Endpoint-only}{endpoint_only}
        \ablationQualRow{$\lambda_{\max}=0.5$}{lambda05}
        \ablationQualRow{$\lambda_{\max}=1.0$}{lambda1}
        & 0\,s & 1.25\,s & 2.5\,s & 3.75\,s & 5\,s \\
    \end{tabular}
    \endgroup
    \caption{\textbf{Qualitative comparison of the ablation variants.}
    Rows follow \cref{tab:ablation}, and columns show matched timestamps.
    Ours renders a pronounced running stride and evolving dust trail with the herd visible in the scene.
    The w/o CFG decoupling variant also omits MMD.}
    \label{fig:qualitative_ablation}
\end{figure}

\Cref{fig:qualitative_ablation} compares the eight variants in \cref{tab:ablation} on the same prompt, with five frames per video at matched timestamps.
The labels follow the quantitative ablations: w/o CFG decoupling also omits MMD, w/o MMD retains decoupling, and Ours includes both components.

Mass Forcing makes the requested running action visually explicit: the elephant exhibits extended, alternating leg poses, forward body motion, and a changing dust trail while the herd remains visible in the background.
The side-on view keeps this locomotion legible across the sequence, matching the prompt's emphasis on an elephant running in the context of its herd.
Without MMD, the elephant increasingly turns toward and approaches the camera, making its relation to the distant herd less prominent.
Without both components, the sequence depicts an elephant already among the herd with more restrained changes in pose.
The other variants also produce motion, but differ in framing, stride visibility, and depiction of the group.
This example complements the aggregate results by showing how the full method combines visible subject motion with the scene content specified by the prompt.

\end{document}